\documentclass{article}
\usepackage{authblk}

\usepackage[table,dvipsnames]{xcolor}
\usepackage{xparse}

\usepackage[margin=1in]{geometry}
\usepackage{subcaption}
\usepackage{enumitem}
\usepackage{float}
\usepackage{pifont}

\usepackage{graphicx}
\usepackage{tikz}
\usetikzlibrary{arrows.meta,calc,positioning,shapes.geometric}

\usepackage{amsmath,amssymb,amsthm}
\usepackage{booktabs}
\usepackage{tabularx}
\usepackage{array}
\usepackage{makecell}
\usepackage{multirow}
\usepackage{caption}
\usepackage{subcaption}
\usepackage{enumitem}
\usepackage{float}

\usepackage{hyperref}
\usepackage{url}
\usepackage{wrapfig}

\newcommand{\bfx}{\mathbf{x}}

\newcommand{\SC}{(S,C)}

\newcommand{\SCsharpTr}{\mathrm{trace(H)}}
\newcommand{\SCsharpTrFS}{\mathrm{trace(H)}_{\mathrm{fs}}}
\newcommand{\SCsharpMax}{\lambda_{\max}\mathrm{(H)}}
\newcommand{\SCsharpMaxFS}{\lambda_{\max}\mathrm{(H)}_{\mathrm{fs}}}

\newcommand{\SCcompLtwo}{\text{param norm}}
\newcommand{\SCcompSpec}{\text{spec norm}}
\newcommand{\SCcompPath}{\text{path norm}}
\newcommand{\SCcompFunc}{\text{func norm}}

\newlength{\SCmetricwidth}
\newcommand{\SCpair}[2]{%
  \ensuremath{\bigl(\makebox[\SCmetricwidth][l]{\ensuremath{#1}}\,,\,#2\bigr)}%
}

\newcommand{\SCtrLtwo}{\SCpair{\SCsharpTr}{\SCcompLtwo}}
\newcommand{\SCtrSpec}{\SCpair{\SCsharpTr}{\SCcompSpec}}
\newcommand{\SCtrPath}{\SCpair{\SCsharpTr}{\SCcompPath}}
\newcommand{\SCtrFunc}{\SCpair{\SCsharpTr}{\SCcompFunc}}
\newcommand{\SClmaxLtwo}{\SCpair{\SCsharpMax}{\SCcompLtwo}}
\newcommand{\SClmaxSpec}{\SCpair{\SCsharpMax}{\SCcompSpec}}
\newcommand{\SClmaxPath}{\SCpair{\SCsharpMax}{\SCcompPath}}
\newcommand{\SClmaxFunc}{\SCpair{\SCsharpMax}{\SCcompFunc}}
\newcommand{\SCtrfsLtwo}{\SCpair{\SCsharpTrFS}{\SCcompLtwo}}
\newcommand{\SCtrfsSpec}{\SCpair{\SCsharpTrFS}{\SCcompSpec}}
\newcommand{\SCtrfsPath}{\SCpair{\SCsharpTrFS}{\SCcompPath}}
\newcommand{\SCtrfsFunc}{\SCpair{\SCsharpTrFS}{\SCcompFunc}}
\newcommand{\SClmaxfsLtwo}{\SCpair{\SCsharpMaxFS}{\SCcompLtwo}}
\newcommand{\SClmaxfsSpec}{\SCpair{\SCsharpMaxFS}{\SCcompSpec}}
\newcommand{\SClmaxfsPath}{\SCpair{\SCsharpMaxFS}{\SCcompPath}}
\newcommand{\SClmaxfsFunc}{\SCpair{\SCsharpMaxFS}{\SCcompFunc}}

\newcommand{\iteSCtrfsLtwo}{\SCpair{\mathit{trace(H)}_{\mathit{fs}}}{\SCcompLtwo}}
\newcommand{\iteSCtrfsSpec}{\SCpair{\mathit{trace(H)}_{\mathit{fs}}}{\SCcompSpec}}
\newcommand{\iteSCtrfsPath}{\SCpair{\mathit{trace(H)}_{\mathit{fs}}}{\SCcompPath}}
\newcommand{\iteSCtrfsFunc}{\SCpair{\mathit{trace(H)}_{\mathit{fs}}}{\SCcompFunc}}

\newcommand{\iteSClmaxfsLtwo}{\SCpair{\lambda_{\mathit{max}}(H)_{\mathit{fs}}}{\SCcompLtwo}}
\newcommand{\iteSClmaxfsSpec}{\SCpair{\lambda_{\mathit{max}}(H)_{\mathit{fs}}}{\SCcompSpec}}
\newcommand{\iteSClmaxfsPath}{\SCpair{\lambda_{\mathit{max}}(H)_{\mathit{fs}}}{\SCcompPath}}
\newcommand{\iteSClmaxfsFunc}{\SCpair{\lambda_{\mathit{max}}(H)_{\mathit{fs}}}{\SCcompFunc}}

\newcommand{\Q}{\mathcal{Q}}

\newcommand{\calP}{\mathcal{P}}

\newcommand{\SCbodysep}{%
  \textcolor{black!25}{%
    \raisebox{-2pt}[0pt][0pt]{%
      \vbox{%
        \offinterlineskip
        \hbox{\rule{0.35pt}{1.8pt}}\kern2pt
        \hbox{\rule{0.35pt}{1.8pt}}\kern2pt
        \hbox{\rule{0.35pt}{1.8pt}}%
      }%
    }%
  }%
}
\newcommand{\SCrsq}[2]{%
  \ensuremath{#1}\,{\tiny \ensuremath{#2}}%
}

\newcommand{\SCfootersep}{%
  \mbox{%
    \makecell[tc]{%
      \SCbodysep\\
      \SCbodysep\\
      \SCbodysep\\
      \SCbodysep\\
      \SCbodysep\\
      \SCbodysep\\
      \SCbodysep%
    }%
  }%
}

\title{\fontsize{20pt}{24pt}\selectfont How Far Can Sharpness and Complexity Jointly Explain Generalization?}

\author{Ziyu Cheng}
\author{Xitong Zhang}
\author{Longxiu Huang}
\author{Rongrong Wang}

\affil{Michigan State University}
\date{} 

\begin{document}

\maketitle

\begin{abstract}
Large-scale empirical studies have revealed that many widely used generalization measures correlate only weakly with generalization, suggesting that relying on a single factor is insufficient to explain it.
In this work, we examine sharpness and complexity jointly, placing the \emph{reach of the two-factor view} at the center: 
how far can they jointly explain generalization, what determines when they succeed, and where do they fall short?
We investigate these questions \emph{quantitatively}, using sixteen alternative metric pairs as probes and Pareto analysis and linear regression as evaluation tools. 
Our results show that suitably defined {\footnotesize $(S,C)$} pairs provide strong explanatory power across a range of settings.
The sharpness perturbation scale substantially affects this success, and the complexity representation influences its robustness.
The remaining failures leave open whether more refined measurements would suffice or additional factors are needed.
\end{abstract}

\section{Introduction}
Deep networks often achieve strong predictive performance in highly overparameterized regimes, yet why they generalize so well remains incompletely understood~\cite{zhang2017rethinking,bartlett2021deep}. 
A broad body of work has sought more refined explanations of generalization, including 
the implicit biases induced by architectures and optimization procedures~\cite{neyshabur2014implicit,soudry2018implicit,gunasekar2018implicit,lyu2019gradient,wei2020improved,smith2021origin}, 
norm- and margin-based bounds~\cite{neyshabur2015norm,neyshabur2018pac,golowich2020size}, 
PAC-Bayes analyses~\cite{dziugaite2017computing,neyshabur2017exploring,dziugaite2021role},
and studies of sharpness and flatness~\cite{hochreiter1997flat,keskar2017large,chaudhari2017entropysgd,dinh2017sharp,tsuzuku2020normalized,petzka2021relative,andriushchenko2022towards,andriushchenko2023modern}.

Across these diverse approaches, two recurring factors are sharpness and complexity. 
\emph{Sharpness}~$(\boldsymbol S)$ characterizes the sensitivity of a learned solution to perturbations and is commonly measured using Hessian statistics~\cite{ghorbani2019investigation,liu2022regularizing,kaur2023maximum} 
or perturbation-induced increases in loss, as in Sharpness-Aware Minimization (SAM)~\cite{foret2020sharpness,kwon2021adapsam}.
\emph{Complexity} $(\boldsymbol C)$ captures the size or effective capacity of the learned model or its hypothesis class and plays a central role in capacity-based bounds~\cite{bartlett2002rademacher}, PAC and PAC-Bayes theory~\cite{mcallester1998some,mcallester1999pac,mcallester2003simplified,guedj2019primer,alquier2024userfriendly}, and many modern generalization analyses~\cite{bartlett2017spectrally,neyshabur2018pac,arora2018stronger,liang2019fisherrao,dziugaite2020robust}. 
Common complexity metrics include weight norm, path norm, spectral norm product, and the posterior-prior KL divergence.

Neyshabur et al.~\cite{neyshabur2017exploring} showed that sharpness alone is insufficient to explain several observed generalization phenomena. Motivated by PAC-Bayes theory, they advocated considering sharpness jointly with complexity. Their bi-criteria plots supported this perspective: combining sharpness with $\ell_2\,\mathrm{norm}$ complexity accounts for generalization differences that sharpness alone failed to explain.

The subsequent large-scale study by Jiang et al.~\cite{jiang2019fantastic}, however, provided less encouraging evidence for the empirical benefit of combining the two factors.
They evaluated more than 40 generalization measures, including measures that combine sharpness and complexity through a prescribed functional form.
Two of these measures, named pacbayes-orig and sharpness-orig in their paper, 
combine norm complexity with expected sharpness and worst-case sharpness, respectively, to form scalar quantities.
They achieved evaluation scores of $0.360$ and $0.398$, suggesting limited consistency with the observed generalization ordering.
Moreover, the authors reported little change in correlation when the norm complexity was removed. 
This challenges the expected empirical benefit of the second factor in the sharpness-complexity construction advocated by~\cite{neyshabur2017exploring}, and casts doubt on whether complexity contributes explanatory information beyond sharpness itself.

These findings leave an important question open: is the two-factor perspective itself inadequate, or is explanatory information lost when sharpness and complexity are combined into a scalar quantity through a prescribed functional form?

To address this, we examine sharpness and complexity jointly \emph{without} committing to a prescribed scalarization. 
We assess the reach of the two-factor view in a systematic and quantitative manner. 
Our central question is:
{
\setlength{\abovedisplayskip}{3pt}
\setlength{\belowdisplayskip}{-1pt}
\begin{equation}
\label{central_question}
\textit{To what extent do sharpness and complexity jointly explain generalization?}
\tag{$\star$}
\end{equation}
}

The answer to this question turns out to be more nuanced than a simple yes or no, as the extent to which the two factors explain generalization varies across architectures, optimizers, and datasets. Nevertheless, we seek to identify persistent patterns that emerge across most of the settings studied using our framework. To this end, we evaluate sixteen alternative metric pairs across trained models spanning different architectures, training algorithms, and datasets, as well as in mixed settings that pool models from multiple sources. We use Pareto analysis and linear regression as evaluation tools to characterize the explanatory power of the evaluated metric pairs.

Our \textbf{main findings} are summarized below.
    \begin{enumerate}[
    leftmargin=3.5em, labelsep=0.5em, 
    label=\roman*.,
    topsep=2.0pt,      
    itemsep=1.0pt,     
    ]
    \item 
    With suitable metric choices, the two-factor view exhibits strong explanatory power across a broad range of the settings examined, including different architectures, training procedures, datasets, and challenging mixed settings.
    \item 
    Sharpness defined over a larger perturbation scale often yields substantially stronger explanatory power, whereas its classical infinitesimal counterpart can fail. The perturbation scale typically matters much more than the choice of Hessian statistic.
    \item 
    Measuring complexity at the function level yields more robust explanatory power than the parameter-level metrics considered.
    \item 
    Despite being scale-variant, the parameter norm can still be an effective explanatory metric in appropriate training regimes, such as SGD with weight decay.
    \item 
    Some regimes remain incompletely explained even by metrics that perform strongly elsewhere. 
    \end{enumerate}

We \emph{emphasize} that the investigated metric pairs and the evaluation tools 
serve as probes for examining the reach of the two-factor view. Other alternative metric definitions and evaluation criteria may also be suitable.
Our {focus} remains on the central question (\ref{central_question}).

\subsection{Related Work}\label{subsec:related_work}

Neyshabur et al.~\cite{neyshabur2017exploring} connected sharpness and norm-based complexity through a PAC-Bayes framework to explain observed variations in generalization (see Appendix~\ref{ap:pac_bayes_motivation} for details). Using expected sharpness together with an $\ell_2$-norm complexity measure, they showed that incorporating the $\ell_2$ norm can improve the explanatory power of sharpness when comparing models trained on true versus random labels across different dataset sizes. Their study was primarily qualitative, aiming to demonstrate the necessity of incorporating a second factor beyond sharpness.

In contrast, our work takes a \emph{quantitative} approach. We systematically evaluate the extent to which $(S,C)$ metric pairs can explain generalization, consider a broad range of alternative definitions for both sharpness and complexity, and study generalization on the original datasets rather than through comparisons with random-label training. Model variation in our experiments is induced by hyperparameter sweeps and different optimizers.

In the large-scale empirical study, Jiang et al.~\cite{jiang2019fantastic} evaluated more than 40 generalization measures across over 10,000 trained models.
Among these, the pacbayes-orig and sharpness-orig  
incorporate both factors. They combine expected sharpness and worst-case sharpness with norm complexity, respectively, through prescribed functional forms.
The authors assessed the correlation between each resulting scalar quantity and generalization, obtaining granulated Kendall scores of $0.360$ and $0.398$, respectively. These scores suggest that the resulting scalar measures have limited explanatory power, which challenges the expected empirical benefits of the joint perspective.
In contrast, we \emph{study the pair} $(S,C)$ directly, \emph{without prescribing a functional form} for combining the two factors. Under this two-dimensional view, our empirical findings are substantially more positive, suggesting that part of the limitation observed in prior studies may arise from the scalarization itself rather than from the two-factor perspective.

\subsection{Notation}
Let \(Z\!=\!\{z_i\}_{i=1}^{m}\) be a training dataset of \(m\) independent samples \(z_i\!=\!(\bfx_i,y_i)\) from an unknown distribution \(\mathcal{D}\), where \(\bfx_i\) and \(y_i\) denote the input and label.
For a model \(h\) and loss function \(\ell(h;z)\), the empirical (or training) and population losses are
{\small$
\ell(h;Z)\!:=\!
\frac{1}{m}\sum_{i=1}^{m}\ell(h;z_i)
$}
and
{\small$
\ell(h;\mathcal{D})\!:=\!
\mathbb{E}_{z\sim\mathcal{D}}\left[\ell(h;z)\right].
$}
The generalization gap measures the increase from training to population loss,
{\small$
\operatorname{Gap}(h)\!:=\!
\ell(h;\mathcal{D})\!-\!\ell(h;Z).
$}
Since \(\mathcal{D}\) is unknown, this gap is commonly estimated by
{\small$
\widehat{\operatorname{Gap}}(h)\!:=\!
\mathrm{test\,loss}-\mathrm{training\,loss}.
$}
We use \(w\) to denote the network parameter vector and \(f_w\) its corresponding function.

\section{A Key Distinction: Explanation vs. Regularization}
\label{sec:distinct_explanation_regularization}
Many quantities related to generalization first attract practical interest because they work well as training regularizers. 
Table~\ref{tab:regularization_explanation_examples} lists several widely used examples.

\begin{wraptable}{r}{0.40\textwidth} 
\caption{
    Common regularizers.
    }
\vspace{-10pt}
    \centering
    \footnotesize
    \setlength{\tabcolsep}{0.5pt}
    \renewcommand{\arraystretch}{1.0}

    \begin{tabularx}{\linewidth}{
        @{}
        >{\centering\arraybackslash}p{0.35\linewidth}
        >{\centering\arraybackslash}X
        @{}
    }
        \toprule
        {Regularizer}
        &
        {Generalization-related factor}
        \\
        \midrule

        noise injection
        &
        sharpness 
        \\

        SAM
        &
        sharpness 
        \\


        weight decay
        &
        complexity 
        \\

        \bottomrule
    \end{tabularx}

    \label{tab:regularization_explanation_examples}
\end{wraptable}

Weight decay in SGD provides a familiar example: regularizing the squared weight norm, \(\|w\|_2^2\), often improves test performance~\cite{krogh1991simpleweightdecay,loshchilov2019decoupled,zhang2019three,dangelo2024why}. 
However, \(\|w\|_2^2\) itself is not always a reliable complexity metric. For networks that admit function-preserving rescalings, the weights can be rescaled so that \(\|w\|_2^2\) changes substantially while the network function and its generalization behavior remain unchanged.
{This highlights an important distinction between \emph{\textbf{a quantity's effectiveness as a regularizer}} and \emph{\textbf{its ability to serve as a well-posed metric with sufficient explanatory power}}. They are related but distinct properties, and therefore motivate different research questions.}

From the perspective of regularization, the question is:
     \textit{does penalizing  $\|w\|_2^2$ in training help yield a better-generalizing model in a particular setting?} 
The explanatory analysis in this work asks:
    \textit{is a smaller $\|w\|_2^2$ {consistently} associated with better generalization across many trained models?} 

These two roles are clearly related: if a quantity exhibits strong explanatory power across many trained models, we would generally expect it to be a promising regularizer across a range of individual settings, although it need not be the best regularizer in every particular case. The converse, however, does not necessarily hold: a quantity may serve as an effective regularizer in certain settings without providing a reliable explanatory factor across a broader population of trained models.

\section{$\SC$ Metric Pairs}
\label{sec:metrics_review}

\subsection{Scaling Invariance }
\label{subsec:parameterization_ambiguity}

Deep networks exhibit \emph{parameterization ambiguity}: distinct parameter vectors \(w\) and \(w'\) can represent exactly the same function, \(f_{w'}=f_w\). A common example is \emph{scaling ambiguity}, which arises from compensating layer-wise rescalings permitted by positively homogeneous activations. For instance, in a ReLU network, multiplying the weights of one layer by \(\alpha>0\) and dividing the weights of the subsequent layer by \(\alpha\) leaves the represented function unchanged, while potentially changing parameter-dependent quantities substantially.

Ideally, a generalization metric should be invariant to such function-preserving rescalings, since the network function, and hence its predictive performance, remains unchanged. Nevertheless, we do not restrict our study exclusively to scale-invariant metrics. In highly overparameterized neural networks, rescaling represents only one source of parameterization ambiguity. Consequently, scale invariance alone does not guarantee invariance to more general function-preserving reparameterizations, nor does it by itself ensure that a metric is well posed. We therefore regard \emph{\textbf{scale invariance}} as a desirable property rather than a necessary criterion for including a metric in our study.

\subsection{Sixteen Metric Pairs — Probes For Studying The Explanatory Power}
\label{subsubsec:complexity_metrics}

\paragraph{Four Sharpness Metrics.} Sharpness describes the geometry of the loss landscape around a learned solution: a solution is considered sharp if small parameter perturbations can cause large increases in loss. 
Local curvature is commonly quantified by Hessian statistics, such as the trace and the top eigenvalue. However, the ordinary Hessian depends on the parameterization. As noted above, certain parameter rescalings can change these statistics without affecting the model’s generalization performance~\cite{dinh2017sharp}.
To reduce scaling ambiguity, we consider sharpness metrics that are \emph{adaptive to parameter magnitudes}.
That is to say, throughout this paper, unless specified otherwise, we compute sharpness metrics using the adaptive Hessian:
\begin{equation}\label{def:adap_hessian}
\scalebox{0.95}{$\displaystyle
\mathrm{H}
=
\operatorname{diag}(|w|)\,\nabla_w^2 \ell(f_{w};Z)\,\operatorname{diag}(|w|).
$}
\end{equation}

We consider both classical infinitesimal sharpness metrics and their finite-scale counterparts, motivated by our observation that their explanatory power can differ substantially.
The infinitesimal metrics are estimated directly from the adaptive Hessian, while the finite-scale metrics are evaluated through perturbed losses.

\begin{itemize}[
    leftmargin=1.5em, labelsep=0.5em,
    topsep=1.5pt,      
    itemsep=1.0pt,     
    ]
    \item 
The Hessian trace, $\SCsharpTr$, \emph{summarizes} loss curvature across parameter directions, capturing second-order changes in the infinitesimal perturbation limit. We estimate it using the Hutchinson method~\cite{hutchinson1989}.

    \item 
The Hessian top eigenvalue, $\SCsharpMax$, measures the \emph{largest} infinitesimal loss curvature among parameter directions. We estimate it using Lanczos iteration~\cite{lanczos1950}.

    \item The finite-scale Hessian trace,
\begin{equation}\label{def:trace_fs}
\scalebox{0.95}{$\displaystyle
\SCsharpTrFS(w;\sigma)
:=
\mathbb{E}_{\epsilon}
\frac{
\left[
\ell(f_{w+\sigma\epsilon};Z)-\ell(f_w;Z)
\right]}{\sigma^2},
\quad
\epsilon=|w|\odot \delta,
\quad
\delta\sim\mathcal{N}(0,I).
$}
\end{equation}
A second-order Taylor expansion gives 
$\SCsharpTrFS(w;\sigma)
\!\to\!
\frac{1}{2}\mathrm{tr(H)}$ 
as $\sigma\!\to\!0$.
Thus, up to a constant factor, this finite-scale $\SCsharpTrFS$ approaches the infinitesimal $\SCsharpTr$ as the perturbation scale approaches zero (Appendix~\ref{subap:pac_bayes_bayes_sharpness}).
We \emph{estimate the expectation} by Monte Carlo sampling: draw independent anisotropic Gaussian perturbations, average loss increases, and normalize by~$\sigma^2$.

    \item The finite-scale Hessian top eigenvalue, 
\begin{equation}\label{def:topeigen_fs}
\scalebox{0.95}{$\displaystyle
\SCsharpMaxFS(w;\sigma)
:=
\max_{\|v\|_2=1}
\frac{
\ell(f_{w+\sigma |w|\odot v};Z)
+
\ell(f_{w-\sigma |w|\odot v};Z)
-
2\ell(f_w;Z)
}{\sigma^2}.
$}
\end{equation}
A second-order Taylor expansion shows that symmetry cancels the linear terms, leaving the limiting quadratic form $v^\top\!H v$ after normalization.
Maximizing over unit directions yields the adaptive Hessian top eigenvalue: $\SCsharpMaxFS(w;\sigma)\to\lambda_{\max}\mathrm{(H)}$ as $\sigma\!\to\!0$ (Appendix~\ref{subap:pac_bayes_bayes_sharpness}).
We \emph{approximate the maximum} using a number of probe directions, refined through gradient ascent, and retain the largest value found. 

\end{itemize}

\paragraph{Four Complexity Metrics.}\label{subsec:complexity_metrics}
\vspace{-0.1cm}
\begin{itemize}[
    leftmargin=1.5em, labelsep=0.5em,
    topsep=1.2pt,      
    itemsep=1.0pt,     
    ]
\item {Parameter Norm.}
The classical norm-based complexity metric, $\|w\|_2$, corresponds to the common regularizer, weight decay in SGD. 
Parameter norm is scale-\emph{variant} by definition.

\item {Function Norm.}
As discussed in Section \ref{subsec:parameterization_ambiguity}, scaling ambiguity is just one type of parameterization ambiguity. To reduce the parameterization ambiguity as a whole, we define a function-level complexity metric.
Consider a network parameterized by $w$, and write $f_w$ for its represented function.
For classification tasks, given a fixed reference dataset \(D_{\mathrm{ref}}\!=\!\{(\bfx_i,y_i)\}_{i=1}^{n}\), let \(Z_w\) denote the logit matrix:
$$
\scalebox{0.95}{$\displaystyle
Z_w(D_{\mathrm{ref}})
:=
\begin{bmatrix}
f_w(\bfx_1)^\top\\
\vdots\\
f_w(\bfx_n)^\top
\end{bmatrix}
\in
\mathbb{R}^{n\times c},
$}
$$
where \(f_w(\bfx_i)\) is the \(c\)-class logit vector for \(\bfx_i\).
Softmax probabilities are invariant under the transformation
\(Z_w \mapsto Z_w+\mathbf{e}\mathbf{1}_c^\top\)
for any \(\mathbf{e}\in\mathbb{R}^{n}\).
We remove this additive ambiguity using the class-centering projection
$\Pi_c\!=\!I_c\!-\!\frac{1}{c}\mathbf{1}_c\mathbf{1}_c^\top$. The \emph{class-centered logit matrix} is then defined as
$$
\scalebox{0.95}{$\displaystyle
\widetilde Z_w(D_{\mathrm{ref}})
:=
Z_w(D_{\mathrm{ref}})\Pi_c
=
\begin{bmatrix}
\widetilde f_w(\bfx_1)^\top\\
\vdots\\
\widetilde f_w(\bfx_n)^\top
\end{bmatrix}
\in
\mathbb{R}^{n\times c},
$}
$$
where 
\(
\widetilde f_w(\bfx_i)
\!=\!
f_w(\bfx_i)
-
\frac{1}{c}
\mathbf{1}_c\mathbf{1}_c^\top f_w(\bfx_i) .
\)
We define the \textit{function norm} as
\begin{equation}\label{eq:def_func_norm}
\scalebox{0.95}{$\displaystyle
\mathrm{func\;norm}
:=
\bigl\| \widetilde Z_w(D_{\mathrm{ref}}) \bigr\|_{F}.
$}
\end{equation}
This quantity is not only scale-invariant but also parameterization-\emph{invariant}.
It can be interpreted as a \emph{proxy} for a simplified posterior-prior KL divergence in function space; see Appendix~\ref{ap:func_norm}.

\item {Path Norm~\cite{neyshabur2015norm}.}
Let \(\mathcal{V}\) denote all input--output paths and
$\pi_v(w)\!=\!\prod_{i\in v}w_i$
the product of multiplicative parameters along path \(v\).
The path norm is
\begin{equation}\label{eq:def_path_norm}
\scalebox{0.95}{$\displaystyle
\mathrm{path\; norm}
:=
\big( \sum_{v\in\mathcal{V}}
\pi_v(w)^2 \big)^{1/2}
=
\big(\sum_{v\in\mathcal{V}}
\prod_{i\in v}w_i^{\,2}\big)^{1/2}.
$}
\end{equation}
Unlike parameter norm, path norm is scale-\emph{invariant}.
In practice, we report the depth-normalized quantity
{\small$\mathrm{path\;norm}^{1/L_{\mathrm{eff}}}$},
where \(L_{\mathrm{eff}}\) counts effective multiplicative layers.
This normalization reduces depth-dependence and facilitates comparisons across networks. 

\item {Spectral Norm Product~\cite{bartlett2017spectrally}.}
For an \(L\)-layer network with weight matrices 
{\footnotesize\(W_1,\dots,W_L\)}, the spectral norm product is
\begin{equation}\label{eq:def_spec_norm}
\scalebox{0.95}{$\displaystyle
\mathrm{spec\; norm}
:=
\prod\nolimits_{j=1}^{L}
\|W_j\|_2,
$}
\end{equation}
where \(\|W_j\|_2\) is the spectral norm of the weight matrix at layer \(j\). 
This quantity is scale-\emph{invariant}, with smaller values suggesting a more controlled input-output mapping.
We likewise report the depth-normalized quantity
{\small$\mathrm{spec\;norm}^{1/L_{\mathrm{eff}}}$},
with \(L_{\mathrm{eff}}\) defined above.

\end{itemize}

\vspace{1.0em}
Table~\ref{tab:metric_notation} summarizes the evaluated metrics and their properties.
We evaluate $4\!\times\!4\!=\!16$ metric pairs obtained by combining the four sharpness metrics with the four complexity metrics.
\textbf{Each metric pair \emph{serves as a probe} to investigate our core question -- the reach of the two-factor view.}

\vspace{-0.1cm}
\begin{table}[h]
\caption{
Evaluated metrics and their properties.
}
\vspace{-10pt}
\centering
\small
\setlength{\tabcolsep}{10pt}
\renewcommand{\arraystretch}{0.8}

\begin{tabular}{@{}
    >{\raggedright\arraybackslash}p{1.8cm}
    >{\raggedright\arraybackslash}p{2.8cm}
    @{\hspace{4em}}
    >{\raggedright\arraybackslash}p{1.8cm}
    >{\raggedright\arraybackslash}p{3.9cm}
@{}}
\toprule

\multicolumn{2}{l}{Sharpness metrics}
& \multicolumn{2}{l}{Complexity metrics} \\

\midrule
$\SCsharpTr$
& scale-invariant
& param norm
& scale-\emph{variant} \\ 

$\SCsharpMax$
& scale-invariant
& func norm 
& parameterization-invariant\\ 

$\SCsharpTrFS$
& scale-invariant
& path norm
& scale-invariant\\ 

$\SCsharpMaxFS$
& scale-invariant
& spec norm 
& scale-invariant\\ 

\bottomrule
\end{tabular}
\label{tab:metric_notation}
\end{table}

\section{Evaluation Tools} 

\paragraph{Linear Regression.}
For a fixed architecture and training procedure, we sweep hyperparameters to obtain \emph{a collection of well-trained models}, each serving as one regression point. 
For model $i$, let \(S_i\) and \(C_i\) denote its sharpness and complexity under the evaluated metric pair, and \(\mathrm{Gap}_i\) its generalization gap (test loss minus training loss).
We fit
\begin{equation*}
\scalebox{0.95}{$\displaystyle
    \mathrm{Gap}_i
    =
    \beta_0
    +
    \beta_S S_i
    +
    \beta_C C_i
    +
    \varepsilon_i,
    \qquad i=1,\ldots,N,
$}
\end{equation*}
where \(N\) is the collection size and \(\varepsilon_i\) is the regression residual.
We report \(R^2\) and the signs of \(\beta_S\) and \(\beta_C\) in Section~\ref{sec:numerical_results}.
A high \(R^2\) with positive coefficients indicates strong explanatory power under a linear approximation.

Linear regression provides a straightforward assessment but is limited to linear relationships.
Our goal, however, is to assess the joint view \emph{without} prescribing a functional form (Section~\ref{subsec:related_work}).
A poor linear fit may reflect nonlinearity rather than failure of the two-factor view.
This motivates our \textbf{{methodological contribution}}:
an order-based evaluation tool that directly tests the monotonic ordering implied by the two-factor view {without} prescribing a functional form.

\paragraph{Pareto Analysis.}
To avoid imposing any functional combination, we introduce {Pareto analysis}. The main idea is to compare the ordering induced by \(\SC\) with the ordering of generalization gaps. This analysis relies only on the \textbf{\emph{monotonic relationship}} suggested by the two-factor view:
\begin{center}
\vspace{-4pt}
\textit{if model \(i\) has {both lower S and lower C} than model \(j\), then model \(i\) should {generalize better}.}
\vspace{-4pt}
\end{center}

Given two trained models \(i\) and \(j\), we call \((i,j)\) a \textbf{candidate pair} if model~\(i\) has both lower $S$ and lower $C$ than model \(j\), i.e., $S_i\!<\!S_j \text{ and } C_i\!<\!C_j .$
Based on the expected {monotonic relationship}, this ordering implies
$\mathrm{Gap}_i\!<\!\mathrm{Gap}_j$.
We focus on such candidate pairs because only these pairs induce a clear expected ordering from the monotonicity. 
If one model has lower $S$ but higher $C$ than the other, no clear generalization ordering follows without specifying a particular functional form, which we do not impose in this paper.
A candidate pair is called a \textbf{contradictory pair} if this expected ordering is violated:
\(
S_i\!<\!S_j,C_i\!<\!C_j,\text{but }
\mathrm{Gap}_i\!>\!\mathrm{Gap}_j.
\)
Such a contradictory pair indicates that the ordering induced by the evaluated metric pair \emph{disagrees} with the generalization ordering.
The frequency of such violations
is quantified by \textbf{\emph{Pareto contradiction rate}},
{
\setlength{\abovedisplayskip}{3pt}
\setlength{\belowdisplayskip}{1pt}
\begin{equation*}\label{eq:pareto_contradiction_rate}
\scalebox{0.90}{$\displaystyle
    \mathrm{PCR}
    :=
    \frac{
    \#\{\text{contradictory pairs}\}
    }{
    \#\{\text{candidate pairs}\}
    } .
$}
\end{equation*}
}

\begin{figure}[h]
\captionsetup{skip=0.5pt}
\captionsetup[subfigure]{font=scriptsize, skip=2pt}

    \centering
    \begin{minipage}{0.85\textwidth}
    \centering     
    \begin{subfigure}[T]{0.30\textwidth}
        \centering
        \includegraphics[width=\linewidth]{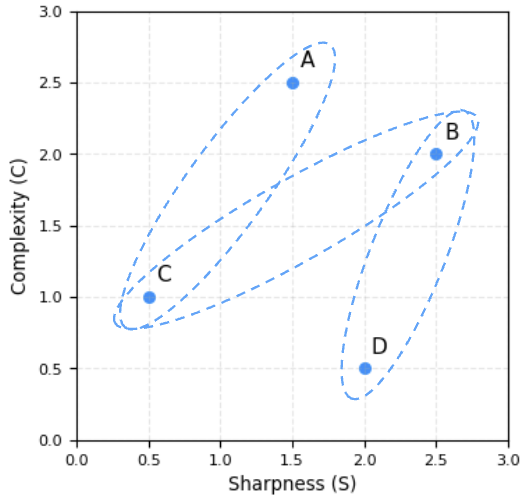}
        \caption{
        \makecell{\emph{candidate pairs}: \tiny{\{AC,BC,BD\}}, \\count 3. 
        }}
        \label{fig:sub1}
    \end{subfigure}
    \hfill
    \begin{subfigure}[T]{0.30\textwidth}
        \centering
        \includegraphics[width=\linewidth]{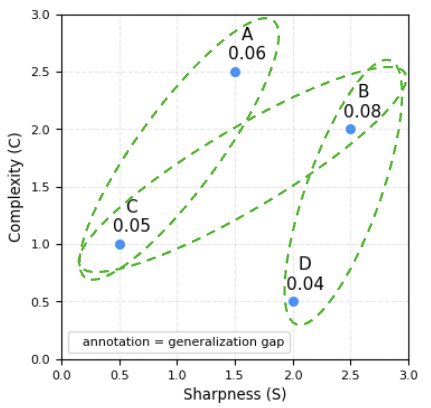}
        \caption{
        \makecell{\emph{contradictory pairs}: None,\\count 0. $\;\text{PCR}\!=\!\frac{0}{3}\!=\!0\%$.
        }}
        \label{fig:sub2}
    \end{subfigure}
    \hfill
    \begin{subfigure}[T]{0.30\textwidth}
        \centering
        \includegraphics[width=\linewidth]{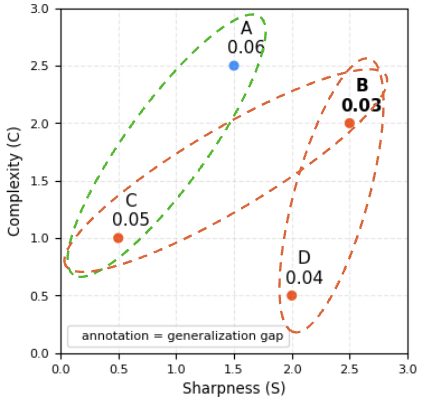}
        \caption{
        \makecell{\emph{contradictory pairs}:\tiny{\{BC,BD\}},\\
        count 2. $\;\text{PCR}\!=\!\frac{2}{3}\!=\!66.7\%$.
        }}
        \label{fig:sub3}
    \end{subfigure}
    
    \end{minipage}
    \caption{
    Computing PCR for an artificial collection of 
    trained models {\footnotesize $\{A,B,C,D\}$}. 
    Each point denotes a trained model, annotated by its generalization gap.
    The middle: no violations occur, yielding low PCR of 0\%. 
    The right: frequent violations occur, yielding high PCR of 66.7\%.
    }
    \label{fig:PCA_illustration}
\end{figure}

A high PCR means frequent violations of the monotonic relationship expected from the two-factor view. Conversely,
\emph{a lower PCR indicates better agreement} between the \(\SC\)-induced ordering and the generalization ordering, and therefore \emph{stronger} explanatory power of the metric pair evaluated. 
Figure~\ref{fig:PCA_illustration} illustrates how PCR is computed.
Numerical PCR results are reported in Section~\ref{sec:numerical_results}. 
Visualized PCRs are presented in
Figure~\ref{fig:selected_PCR_visual_resnet18} and Figure~\ref{fig:PCR_visual_resnet18} in the Appendix.
\vspace{-0.15cm}

\vspace{-1.8em}
\paragraph{Evaluation Criteria:} interpreting the two analytical tools together.
\label{subsec:evaluation_criteria}
Pareto analysis and linear regression provide complementary assessments of the explanatory power of an evaluated metric pair.
PCR measures violations of the monotonic ordering predicted by the two-factor view (\emph{lower is better}), while $R^2$ evaluates how well a linear model captures the relationship (\emph{higher is better}).
The outcomes of interpreting their results jointly are summarized in Table~\ref{tab:joint_diagnostic_interpretation}.

\vspace{-0.05cm}
\begin{table}[h]
    \caption{Joint interpretation of Pareto analysis and linear regression results.}
    \label{tab:joint_diagnostic_interpretation}
    \vspace{-10pt}
    \centering
    \small
    \renewcommand{\arraystretch}{0.9}
    \setlength{\tabcolsep}{8pt}

    \resizebox{0.99\textwidth}{!}{%
\begin{tabular}{
    >{\centering\arraybackslash}p{1.0cm}
    >{\raggedleft\arraybackslash}p{1.2cm}
    >{\raggedright\arraybackslash}p{3.2cm}
    >{\raggedright\arraybackslash}p{6.8cm}
    }

    \toprule

    \multicolumn{3}{c}{%
        {Assessments of Two-Factor Explanatory Power}%
    }
    & \multirow{2}{*}{Interpretation} \\[-1.5pt]
    \cmidrule(lr){1-3}

    {PCR}$\,\downarrow$
    & $R^2\,\uparrow$
    & {coeffi. signs of $(S,C)$}
    &  \\[-1.0pt]
    \midrule

    low & high & positive
    & order-consistent, approximately linear \\

    low & low & positive
    & order-consistent, likely nonlinear \\

    high & high & positive
    & order-inconsistent, despite linear trend \\

    high & low & positive
    & order-inconsistent, no linear trend \\

    any & any & at least one negative
    & directionally inconsistent with two-factor view \\

    \bottomrule
    \end{tabular}%
    }
\end{table}
\vspace{-0.05cm}

We \emph{\textbf{primarily use}} PCR to assess explanatory power because it does not assume a functional relationship between \(S,C\), and the generalization gap. We \emph{{retain}} linear regression as it provides a complementary diagnostic of whether the relationship is approximately additive. 
We regard low PCR as favorable, and high \(R^2\) together with positive regression coefficients provides additional evidence of an approximately linear relationship. See Appendix~\ref{ap:diagnostic_criteria} for a detailed discussion.

\section{Numerical Results and Observations}
\label{sec:numerical_results}

\paragraph{Experimental Setup.}
We evaluate the explanatory power of the two-factor view across a diverse collection of trained networks. 
For each experimental setting, we fix the dataset, architecture, and training algorithm, and obtain a population of trained models by varying hyperparameters. We then retain well-trained models whose training loss is sufficiently small; see Appendix~\ref{ap:experiment_protocol} for details.
For every well-trained model, we record its generalization gap (test loss minus training loss) and compute its sharpness and complexity values under the evaluated metrics described in Section~\ref{sec:metrics_review}.
We then assess how well each metric pair explains the variation in generalization gaps across the resulting model population using two  complementary assessments: 
(i) PCR, which measures how frequently the expected ordering of generalization gaps is violated; and 
(ii) \(R^2\) together with the coefficient signs from linear regression of the generalization gap.

\begin{figure}[h]
\captionsetup{skip=-8pt, belowskip=-4pt} %
    \centering
    
    \begin{subfigure}[b]{0.249\textwidth}
        \centering
        \includegraphics[width=\linewidth]{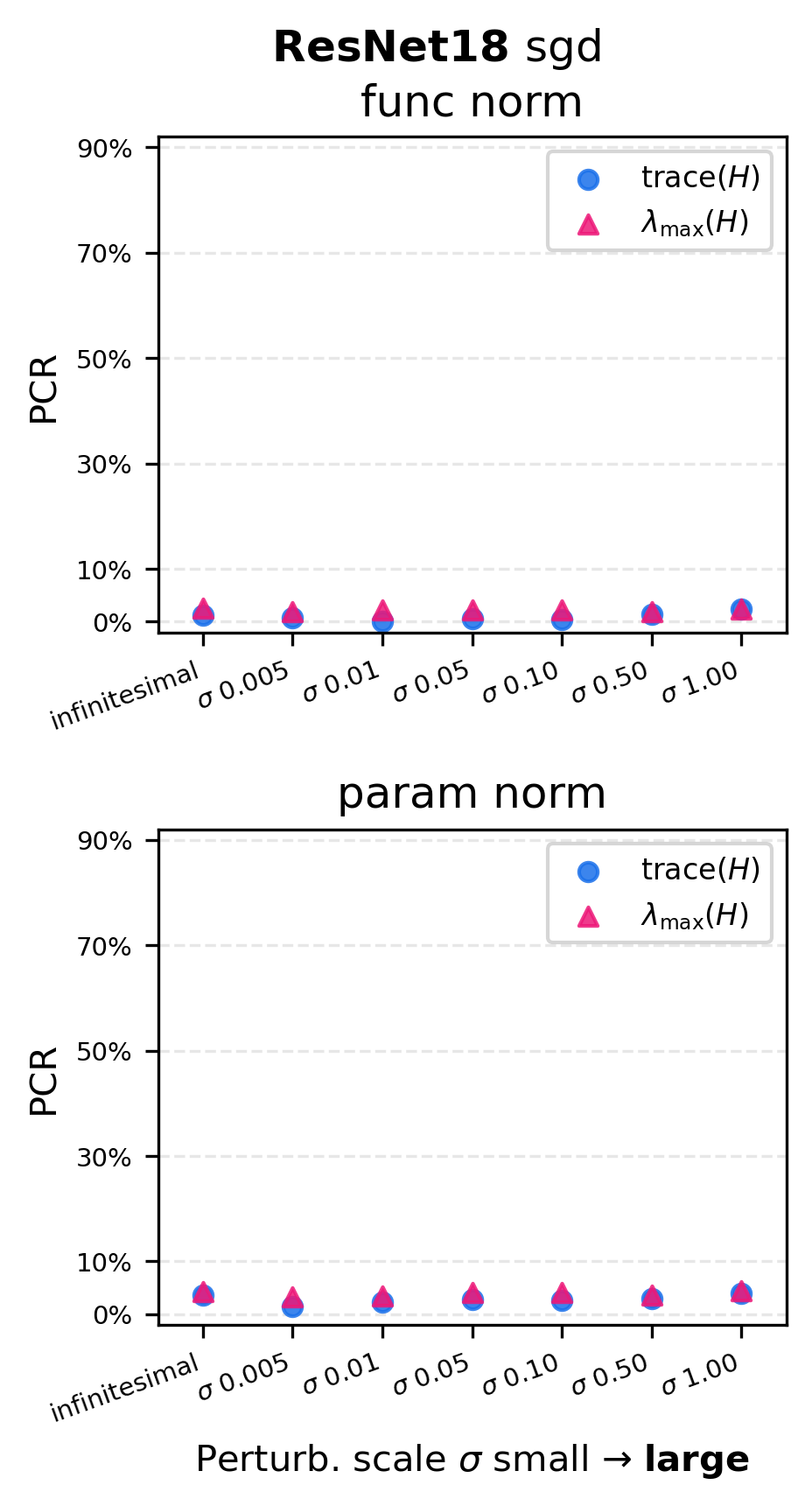}
        \label{fig:panel_a}
    \end{subfigure}\hfill
    \begin{subfigure}[b]{0.249\textwidth}
        \centering
        \includegraphics[width=\linewidth]{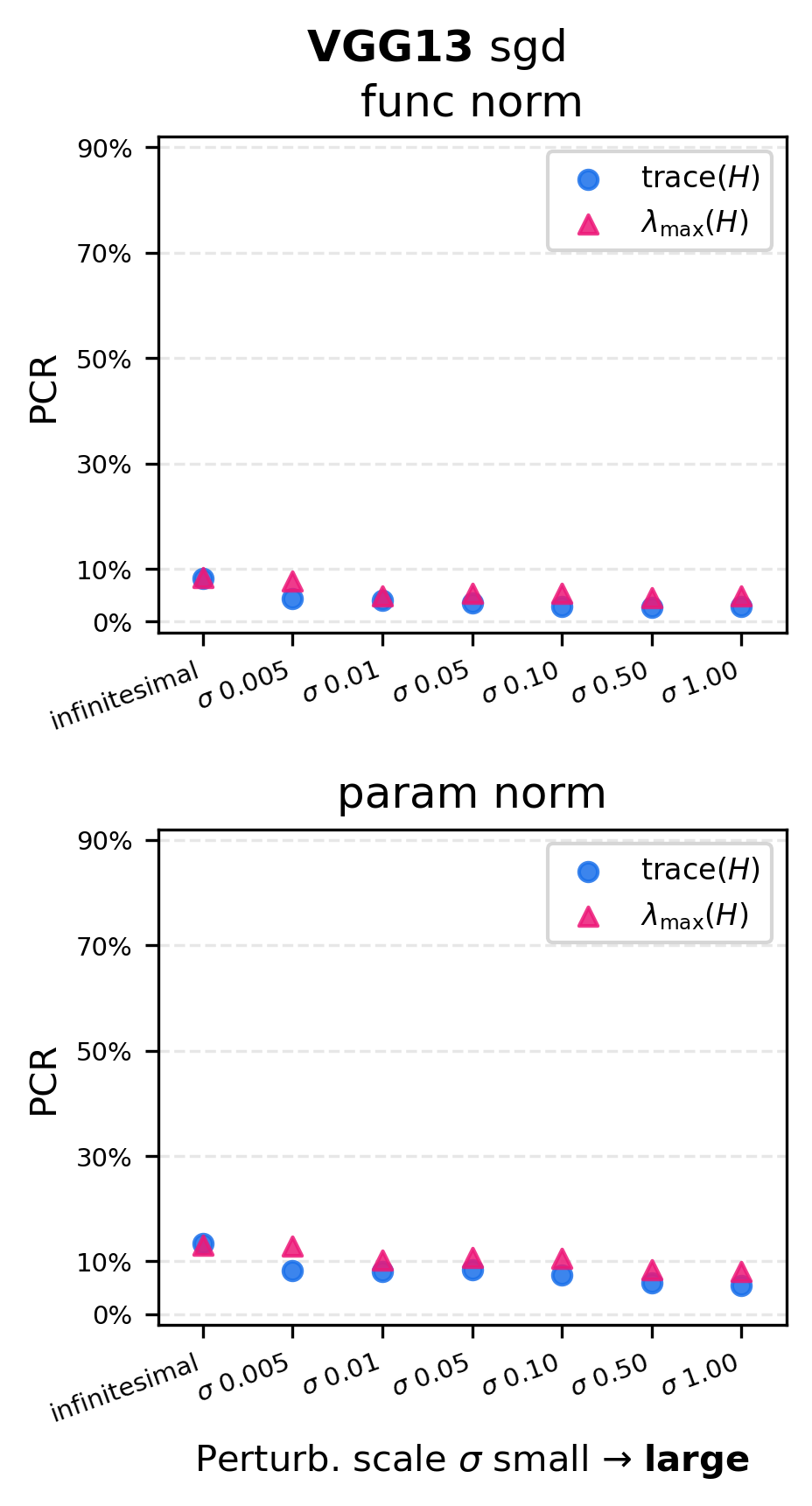}
        \label{fig:panel_b}
    \end{subfigure}\hfill
    \begin{subfigure}[b]{0.249\textwidth}
        \centering
        \includegraphics[width=\linewidth]{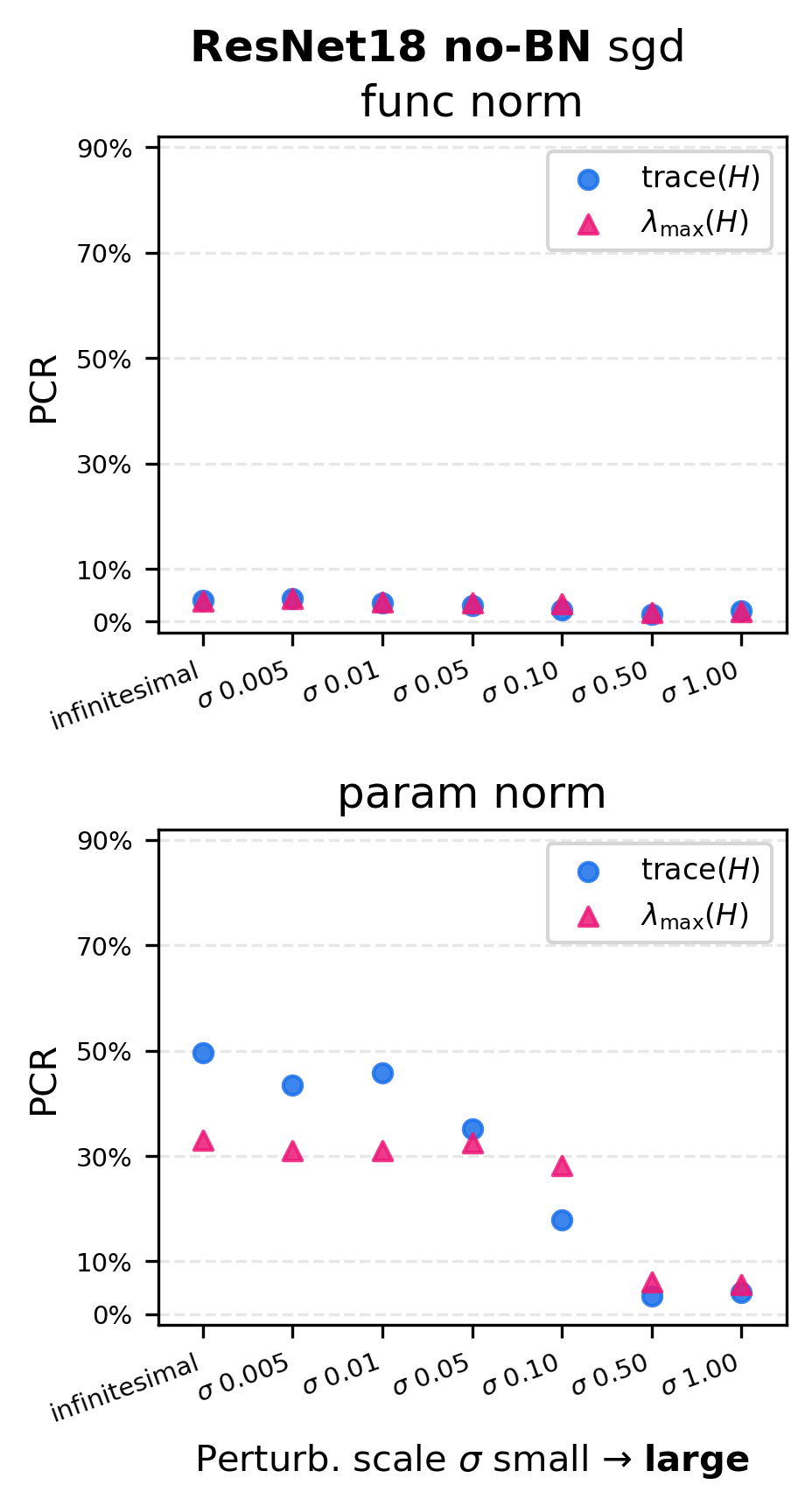}
        \label{fig:panel_c}
    \end{subfigure}\hfill
    \begin{subfigure}[b]{0.249\textwidth}
        \centering
        \includegraphics[width=\linewidth]{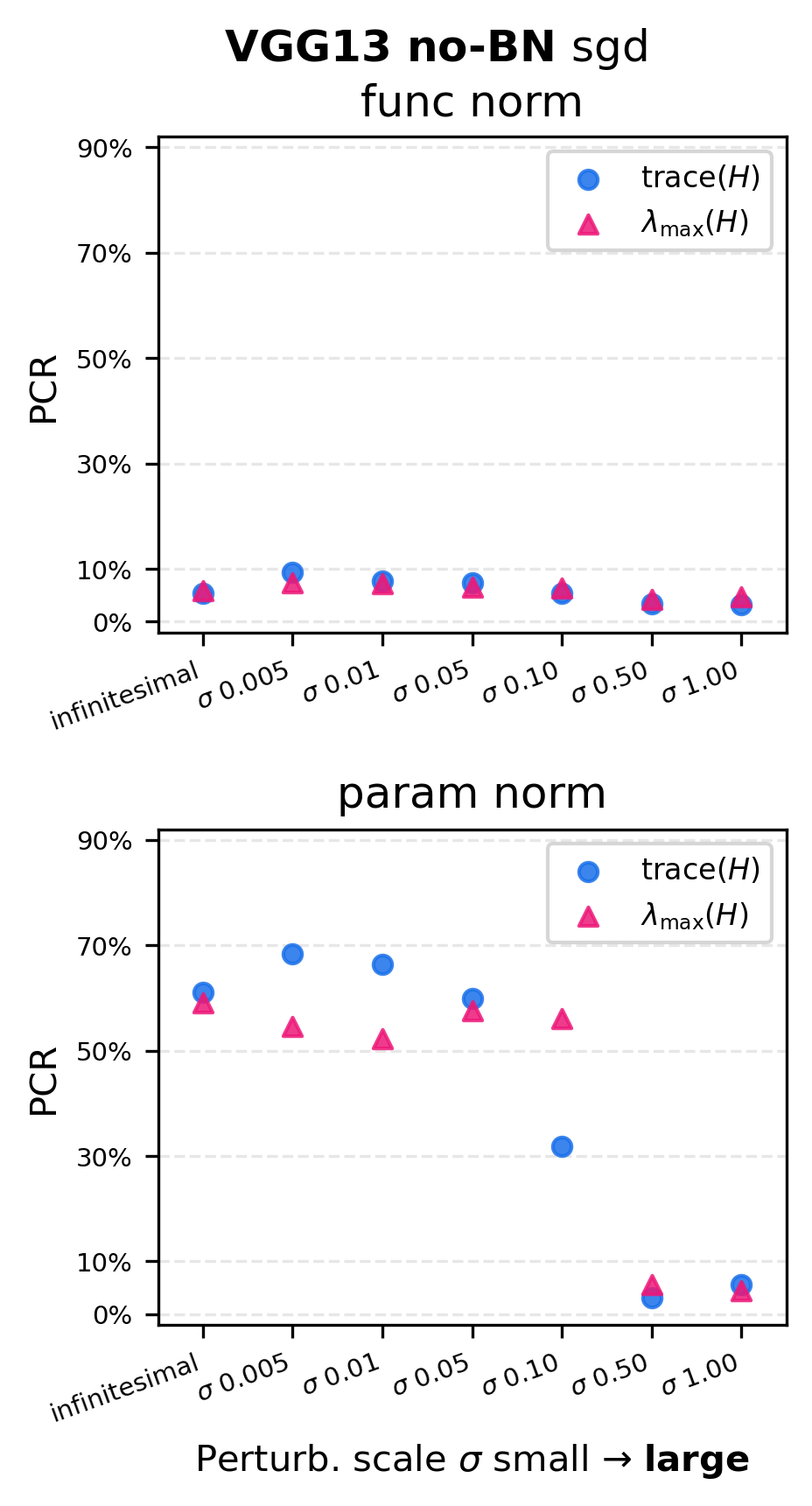}
        \label{fig:panel_d}
    \end{subfigure}
\caption{
How PCR varies as the sharpness perturbation radius increases from an infinitesimal scale to small and then large finite scale.
Each column corresponds to an architecture. 
In each subplot, with the complexity metric fixed, sharpness defined over a \emph{larger} neighborhood often yields lower PCR, indicating stronger explanatory power and supporting Observation~\ref{obs:larger_scale_much_better}.
The row-wise comparisons support Observations~\ref{obs:func_norm_more_stable}, \ref{obs:param_norm_still_good}, and~\ref{obs:path_spec_norm_bad}.
Results of path norm and spectral norm are provided in Figure~\ref{fig:varying_perturb_scale_full}.
}
\label{fig:varying_perturb_scale}
\end{figure}

We repeat this analysis across different architectures, datasets, and training procedures. 
To further test the robustness of the observed patterns, we additionally consider mixed settings, in which models from multiple sources are pooled together. 
The main results are reported in Tables~\ref{tab:main_result_table}, \ref{tab:main_result_table_mixedsetting}, \ref{tab:trained_by_adamw},  Figure~\ref{fig:varying_perturb_scale}, and Figure~\ref{fig:varying_perturb_scale_full}.

\vspace{-0.5em}
\paragraph{Observations based on the results.}

\begin{enumerate}
    [leftmargin=1.5em, labelsep=0.5em,
    topsep=2.5pt,      
    itemsep=1.5pt,     
    parsep=0.25pt,      
    ]
    \item \textbf{The two-factor view provides \emph{strong} explanatory power across different settings.} 
    Overall, our findings are encouraging: with appropriate metric choices, sharpness and complexity can jointly account for a substantial fraction of the variation in generalization across a range of settings.
    The results in Tables~\ref{tab:main_result_table}, \ref{tab:main_result_table_mixedsetting}, \ref{tab:trained_by_adamw}, Figure~\ref{fig:varying_perturb_scale} and~\ref{fig:varying_perturb_scale_full} make this answer concrete. For nearly every setting evaluated, at least one metric pair exhibits strong explanatory power.
    WideResNet no-BN is the only exception where no evaluated metric pair shows strong explanatory power (Observation~\ref{obs:only_bad_exception}).

    \item \textbf{The two-factor explanatory power \emph{remains strong} when pooling across architectures, optimizers and even datasets, but this depends strongly on the choice of complexity metric.}\label{obs:mixed_setting}
    Some metric pairs perform well within individual settings (Tables~\ref{tab:main_result_table} and~\ref{tab:trained_by_adamw}) but deteriorate substantially in mixed settings (Table~\ref{tab:main_result_table_mixedsetting}).
    Pairs involving function norm remain much more stable: paired with finite-scale sharpness, function norm consistently yields low PCR, whereas parameter norm can yield much higher PCR in the mixed settings.

    \item \textbf{Sharpness defined over a \emph{larger} neighborhood has much stronger explanatory power.} \label{obs:larger_scale_much_better}
    Conventional sharpness metrics, $\SCsharpTr$ and $\SCsharpMax$, characterize local loss changes as the perturbation radius approaches zero and show relatively weak explanatory power. In contrast, their finite-scale counterparts, $\SCsharpTrFS$ and $\SCsharpMaxFS$, can provide substantially stronger explanatory power at larger perturbation scales.
    Holding the complexity metric fixed clearly reveals this \emph{infinitesimal versus finite-scale} contrast.
    For ResNet18 no-BN in Table~\ref{tab:main_result_table}, with parameter norm as complexity, replacing $\SCsharpTr$ with $\SCsharpTrFS$ \emph{reduces} PCR from 49.5\% to 3.4\% and \emph{increases} $R^2$ from 0.21 to 0.68.
    We \textbf{\emph{further examine}} how PCR changes as the sharpness perturbation radius increases from an \emph{infinitesimal} scale to \emph{small} and then \emph{large} finite scale.
    Figures~\ref{fig:varying_perturb_scale} and~\ref{fig:varying_perturb_scale_full} show this progression, with the complexity metric fixed in each subplot. 
    For sharpness based on either Hessian trace (blue circle) or top eigenvalue (pink triangle), PCR often decreases at larger scales, indicating stronger explanatory power.
    This trend is especially pronounced when a sharpness metric is paired with parameter norm, path norm, or spectral norm.

    \item \textbf{The \emph{perturbation scale} matters much more than the choice of Hessian statistic.}\label{obs:scale_matter_more}
    In most four-by-two mini-blocks in the tables, which compare four sharpness metrics at a fixed complexity metric, the trace and top eigenvalue often give similar results, while \emph{the improvement from infinitesimal to finite-scale sharpness is much larger}. This suggests that the scale at which the loss landscape is probed plays a substantially greater role than the way its curvature is summarized.
    For example, in the parameter norm mini-block for sgd-trained VGG13 no-BN in Table~\ref{tab:main_result_table}, $\SCsharpTr$ and $\SCsharpMax$ produce PCRs $61.0\%$ and $59.0\%$, while their finite-scale counterparts, $\SCsharpTrFS$ and $\SCsharpMaxFS$, yield much lower PCRs $3.0\%$ and $5.6\%$.
    Figure~\ref{fig:varying_perturb_scale} and~\ref{fig:varying_perturb_scale_full} provide more visible evidence for this observation.
    PCR differences between $\SCsharpTr$ (blue circle) and $\SCsharpMax$ (pink triangle) are often small or even tiny, whereas those between small-scale sharpness {\scriptsize$(\sigma\!\le\!0.10)$} and large-scale sharpness {\scriptsize$(\sigma\!\geq\!0.50)$} can be substantial.

\begin{table}[H]
\caption{
\textbf{Main results.} Assessments of two-factor explanatory power by Pareto analysis and linear regression.
Each vertical block, separated by dashed lines, corresponds to a fixed architecture, optimizer, and dataset. For each block, models are generated by varying hyperparameters, and sixteen metric pairs are evaluated using PCR based on generalization-gap rankings and \(R^2\) from linear regression against generalization gap. 
\emph{Low} PCR is prioritized as evidence of \emph{strong explanatory power},
and high $R^2$ with positive coefficients further supports an approximately linear relationship (Section~\ref{subsec:evaluation_criteria}).
Finite-scale sharpness metrics use the large perturbation scale {\footnotesize$\sigma\!=\!0.50$}.
Path norm and spectral norm lack natural extensions to ViT, so the corresponding entries are left blank.
See visualized PCRs in Figure~\ref{fig:PCR_visual_resnet18}.
}
\vspace{-9pt}
\centering
\footnotesize
\setlength{\tabcolsep}{5pt}
\setlength{\fboxsep}{0.2pt}
\renewcommand{\arraystretch}{1.15}

\resizebox{0.99\linewidth}{!}{%
\begin{tabular}{
    @{}l
    *{3}{ 
        >{\raggedleft\arraybackslash}p{1.0cm}
        >{\centering\arraybackslash}p{1.45cm}
        !{\SCbodysep}
    }
    >{\raggedleft\arraybackslash}p{1.00cm}
    >{\centering\arraybackslash}p{1.45cm}
    @{}
}
\toprule 
\multicolumn{1}{@{}c}{%
    \multirow{4}{*}{%
        \makecell[c]{%
            $\boldsymbol{(S,C)}$\\
            \textbf{Metric Pair}%
        }%
    }%
}
& \multicolumn{8}{c}{\textbf{CIFAR-10}} \\[-3pt]

& \multicolumn{2}{c}{\textbf{ResNet18} \scriptsize{sgd}}
& \multicolumn{2}{c}{\textbf{VGG13} \scriptsize{sgd}}
& \multicolumn{2}{c}{\textbf{ResNet18 no-BN} \scriptsize{sgd}}
& \multicolumn{2}{c}{\textbf{VGG13 no-BN} \scriptsize{sgd}} \\[-3pt]

\cmidrule(lr){2-3}
\cmidrule(lr){4-5}
\cmidrule(lr){6-7}
\cmidrule(lr){8-9}

& \multicolumn{1}{r}{\textbf{PCR}}
& \multicolumn{1}{c}{$\boldsymbol{R^2}\,(S,C)$}

& \multicolumn{1}{r}{\textbf{PCR}}
& \multicolumn{1}{c}{$\boldsymbol{R^2}\,(S,C)$}

& \multicolumn{1}{r}{\textbf{PCR}}
& \multicolumn{1}{c}{$\boldsymbol{R^2}\,(S,C)$}

& \multicolumn{1}{r}{\textbf{PCR}}
& \multicolumn{1}{c}{$\boldsymbol{R^2}\,(S,C)$}
\\[-3pt]

& \multicolumn{8}{c}{%
    \scriptsize 
    \qquad PCR$\downarrow$: lower is better 
    \qquad
    $R^2$$\uparrow$: higher with {\tiny$(+,+)$} is better %
} \\[-3pt]

\midrule

\SCtrFunc
& 1.2\% & \SCrsq{0.48}{(+,+)}
& 8.2\% & \SCrsq{0.31}{(+,+)}
& 4.1\% & \SCrsq{0.79}{(+,+)}
& 5.4\% & \SCrsq{0.68}{(+,+)}\\

\SClmaxFunc
& 2.6\% & \SCrsq{0.45}{(+,+)}
& 8.4\% & \SCrsq{0.31}{(+,+)}
& 3.9\% & \SCrsq{0.79}{(+,+)} 
& 5.8\% & \SCrsq{0.68}{(+,+)}\\

\iteSCtrfsFunc
& 1.5\% & \SCrsq{0.95}{(+,+)}
& 2.8\% & \SCrsq{0.97}{(+,+)} 
& 1.5\% & \SCrsq{0.85}{(+,+)} 
& 3.4\% & \SCrsq{0.85}{(+,+)}\\

\iteSClmaxfsFunc
& 1.9\% & \SCrsq{0.92}{(+,+)}
& 4.5\% & \SCrsq{0.87}{(+,+)}
& 1.7\% & \SCrsq{0.82}{(+,+)} 
& 4.3\% & \SCrsq{0.85}{(+,+)}\\

\addlinespace[2pt]

\SCtrLtwo
& 3.5\%  & \SCrsq{0.84}{(+,+)}
& 13.4\% & \SCrsq{0.70}{(+,+)} 
& 49.5\% & \SCrsq{0.21}{(-,+)} 
& 61.0\% & \SCrsq{0.42}{(-,+)} \\

\SClmaxLtwo
& 4.4\%  & \SCrsq{0.84}{(+,+)}
& 13.0\% & \SCrsq{0.69}{(+,+)} 
& 33.0\% & \SCrsq{0.20}{(-,+)}  
& 59.0\% & \SCrsq{0.42}{(+,+)} \\

\iteSCtrfsLtwo
& 3.0\% & \SCrsq{0.98}{(+,+)}
& 6.0\% & \SCrsq{0.95}{(+,+)}
& 3.4\% & \SCrsq{0.68}{(+,+)}  
& 3.0\% & \SCrsq{0.72}{(+,-)} \\

\iteSClmaxfsLtwo
& 3.6\% & \SCrsq{0.97}{(+,+)}
& 8.5\% & \SCrsq{0.92}{(+,+)}
& 6.0\% & \SCrsq{0.56}{(+,+)} 
& 5.6\% & \SCrsq{0.70}{(+,-)} \\

\addlinespace[2pt]

\SCtrPath 
& 15.1\% & \SCrsq{0.90}{(+,+)}
& 31.1\% & \SCrsq{0.85}{(-,+)}
& 61.0\% & \SCrsq{0.19}{(-,+)}
& 66.7\% & \SCrsq{0.42}{(-,+)}\\

\SClmaxPath
& 15.7\% & \SCrsq{0.90}{(+,+)}
& 31.0\% & \SCrsq{0.84}{(-,+)}
& 47.2\% & \SCrsq{0.18}{(-,+)}
& 68.3\% & \SCrsq{0.42}{(-,+)}\\ 

\iteSCtrfsPath
& 8.6\% & \SCrsq{0.96}{(+,+)}
& 16.8\% &\SCrsq{0.86}{(+,+)}
& 5.1\% & \SCrsq{0.68}{(+,+)}
& 3.2\% & \SCrsq{0.72}{(+,-)}\\

\iteSClmaxfsPath
& 9.9\% & \SCrsq{0.96}{(+,+)}
& 19.6\% &\SCrsq{0.84}{(+,+)}
& 7.7\% & \SCrsq{0.56}{(+,+)}
& 6.1\% & \SCrsq{0.70}{(+,-)}\\

\addlinespace[2pt]

\SCtrSpec
& 54.0\% & \SCrsq{0.07}{(+,+)}
& 53.6\% & \SCrsq{0.17}{(-,+)} 
& 80.6\% & \SCrsq{0.20}{(-,+)} 
& 69.1\% & \SCrsq{0.44}{(-,+)} \\

\SClmaxSpec
& 55.3\% & \SCrsq{0.05}{(+,+)}
& 54.1\% & \SCrsq{0.17}{(-,+)}
& 74.0\% & \SCrsq{0.17}{(-,+)} 
& 72.2\% & \SCrsq{0.44}{(-,+)} \\

\iteSCtrfsSpec
& 14.0\% & \SCrsq{0.88}{(+,-)}
& 21.0\% & \SCrsq{0.83}{(+,-)}
&  4.4\% & \SCrsq{0.66}{(+,+)} 
&  3.4\% & \SCrsq{0.75}{(+,-)} \\

\iteSClmaxfsSpec
& 30.5\% & \SCrsq{0.82}{(+,-)}
& 31.4\% & \SCrsq{0.58}{(+,-)} 
&  7.8\% & \SCrsq{0.53}{(+,+)} 
&  6.7\% & \SCrsq{0.73}{(+,-)} \\[-2pt]

\midrule

\multicolumn{1}{@{}c}{%
    \multirow{3}{*}{%
        \makecell[c]{%
            $\boldsymbol{(S,C)}$\\
            \textbf{Metric Pair}%
        }%
    }%
}
& \multicolumn{8}{c}{\textbf{CIFAR-100}} \\[-3pt]

& \multicolumn{2}{c}{\textbf{WideResNet} \scriptsize{sgd}}
& \multicolumn{2}{c}{\textbf{ViT} \scriptsize{adamw}}
& \multicolumn{2}{c}{\textbf{WideResNet no-BN} \scriptsize{sgd}}
& \multicolumn{2}{c}{\textbf{ViT no-LN} \scriptsize{adamw}} \\[-3pt]

\cmidrule(lr){2-3}
\cmidrule(lr){4-5}
\cmidrule(lr){6-7}
\cmidrule(lr){8-9}

& \multicolumn{1}{r}{\textbf{PCR}}
& \multicolumn{1}{c}{$\boldsymbol{R^2}\,(S,C)$}
& \multicolumn{1}{r}{\textbf{PCR}}
& \multicolumn{1}{c}{$\boldsymbol{R^2}\,(S,C)$}
& \multicolumn{1}{r}{\textbf{PCR}}
& \multicolumn{1}{c}{$\boldsymbol{R^2}\,(S,C)$}
& \multicolumn{1}{r}{\textbf{PCR}}
& \multicolumn{1}{c}{$\boldsymbol{R^2}\,(S,C)$}
\\[-3pt]


\midrule

\SCtrFunc
&  4.6\% & \SCrsq{0.88}{(+,+)} 
& 40.6\% & \SCrsq{0.80}{(-,+)} 
& 27.7\% & \SCrsq{0.40}{(+,+)} 
& 16.7\% & \SCrsq{0.70}{(+,+)}  \\

\SClmaxFunc
&  2.9\% & \SCrsq{0.88}{(+,+)} 
& 26.2\% & \SCrsq{0.80}{(-,+)} 
& 31.1\% & \SCrsq{0.29}{(+,+)} 
& 16.5\% & \SCrsq{0.72}{(+,+)} \\

\iteSCtrfsFunc
& 1.7\% & \SCrsq{0.95}{(+,+)} 
& 9.4\% & \SCrsq{0.82}{(+,+)} 
& 31.3\% &\SCrsq{0.32}{(-,+)} 
& 12.8\% & \SCrsq{0.74}{(+,+)} \\

\iteSClmaxfsFunc
& 1.8\% & \SCrsq{0.92}{(+,+)} 
& 14.5\% & \SCrsq{0.80}{(-,+)} 
& 31.5\% & \SCrsq{0.36}{(-,+)} 
& 13.2\% & \SCrsq{0.74}{(+,+)}  \\

\addlinespace[2pt]

\SCtrLtwo
& 2.4\% & \SCrsq{0.93}{(+,+)} 
& 39.8\% & \SCrsq{0.78}{(-,+)} 
& 51.0\% & \SCrsq{0.05}{(+,+)} 
& 24.5\% & \SCrsq{0.04}{(+,+)} \\

\SClmaxLtwo
& 1.5\% & \SCrsq{0.91}{(+,+)} 
& 18.3\% & \SCrsq{0.78}{(-,+)} 
& 54.4\% & \SCrsq{0.01}{(-,-)} 
& 17.9\% & \SCrsq{0.13}{(+,+)}\\

\iteSCtrfsLtwo
& 2.1\% & \SCrsq{0.96}{(+,+)} 
& 7.0\% & \SCrsq{0.78}{(+,+)}  
& 40.0\% & \SCrsq{0.01}{(-,+)} 
& 39.5\% & \SCrsq{0.47}{(+,-)}\\

\iteSClmaxfsLtwo
& 2.3\% & \SCrsq{0.95}{(+,+)} 
& 11.0\% & \SCrsq{0.78}{(-,+)} 
& 41.4\% & \SCrsq{0.04}{(-,+)} 
& 20.2\% & \SCrsq{0.35}{(+,-)}\\

\addlinespace[2pt]

\SCtrPath
& 12.8\% & \SCrsq{0.90}{(+,+)} 
& -- & -- 
& 56.9\% & \SCrsq{0.03}{(+,+)} 
& -- & -- \\

\SClmaxPath
& 8.4\% & \SCrsq{0.91}{(+,+)} 
& -- & -- 
& 61.8\% & \SCrsq{0.02}{(-,-)} 
& -- & -- \\

\iteSCtrfsPath
& 4.7\% & \SCrsq{0.96}{(+,+)} 
& -- & -- 
& 41.9\% & \SCrsq{0.01}{(+,-)} 
& -- & -- \\

\iteSClmaxfsPath
& 4.3\% & \SCrsq{0.95}{(+,+)} 
& -- & -- 
& 43.3\% & \SCrsq{0.03}{(-,-)} 
& -- & -- \\

\addlinespace[2pt]

\SCtrSpec
& 47.4\% & \SCrsq{0.43}{(+,-)} 
& -- & -- 
& 53.0\% & \SCrsq{0.03}{(+,+)} 
& -- & -- \\

\SClmaxSpec
& 43.9\% & \SCrsq{0.53}{(+,-)} 
& -- & -- 
& 57.4\% & \SCrsq{0.03}{(-,-)} 
& -- & -- \\

\iteSCtrfsSpec
& 5.8\% & \SCrsq{0.90}{(+,-)} 
& -- & -- 
& 38.5\% & \SCrsq{0.01}{(-,-)} 
& -- & -- \\

\iteSClmaxfsSpec
& 10.8\% & \SCrsq{0.83}{(+,-)} 
& -- & -- 
& 39.6\% & \SCrsq{0.05}{(-,-)} 
& -- & -- \\
\bottomrule
\end{tabular}
}
\label{tab:main_result_table}
\end{table}

    \item 
    \textbf{Function norm exhibits \emph{more stable} explanatory power than parameter norm.}\label{obs:func_norm_more_stable}
    This stability appears in two ways.
    \textbf{\emph{First}}, the explanatory power of function norm is \emph{less sensitive to sharpness choice}, particularly infinitesimal versus finite-scale sharpness.
    For VGG13 no-BN in Table~\ref{tab:main_result_table}, function norm maintains strong explanatory power with both $\SCsharpMax$ and $\SCsharpMaxFS$ ({\footnotesize PCRs $5.8\%$ and $4.3\%$}, respectively), whereas parameter norm varies markedly ({\footnotesize $59.0\%$ and $5.6\%$}).
    The function norm subplots in the first row of Figure~\ref{fig:varying_perturb_scale} likewise show highly consistent explanatory power across sharpness choices and scales, from infinitesimal to large.
    \textbf{\emph{Second}}, function norm retains strong and robust explanatory power when pooling models across architectures, optimizers, or datasets into \emph{mixed settings}, whereas parameter norm can collapse; Table~\ref{tab:main_result_table_mixedsetting} shows this contrast. 

    \item \textbf{Although scale-\emph{variant}, parameter norm can still explain generalization effectively in appropriate training regimes.}\label{obs:param_norm_still_good}
    SGD with weight decay may encourage balanced weight scales during training, limiting the influence of parameterization ambiguity.
    In such regimes, lacking scale invariance does not necessarily weaken the explanatory power of parameter norm.
    For example, sgd-trained WideResNet in Table~\ref{tab:main_result_table}, pairing $\SCsharpMaxFS$ with parameter norm gives \(R^2\) of 0.95, versus \(0.92\) with function norm, and PCRs are both below {$3\%$}.

    \item \textbf{Path norm and spectral norm exhibit less consistent explanatory power across settings.}\label{obs:path_spec_norm_bad}
    Although both can perform well in certain individual settings,
    their performance varies considerably across architectures and
    training settings.
    Many corresponding pairs exhibit large
    PCR, lower $R^2$, or negative regression coefficients, especially with infinitesimal sharpness, as seen for sgd-trained ResNet18 no-BN in Table~\ref{tab:main_result_table}. 
    Finite-scale sharpness often improves these poor results, but they remain less consistent across settings than those obtained with function norm.

    \item \textbf{Some regimes remain incompletely explained, even by the strongest metric pairs.} \label{obs:only_bad_exception}
    For WideResNet no-BN in Table~\ref{tab:main_result_table}, finite-scale sharpness paired with function norm still gives weak results: 
    $(\SCsharpTrFS,\text{func norm})$ yields PCR 31.3\% and $R^2$ 0.32,
    with similarly poor results for $(\SCsharpMaxFS,\text{func norm})$. 
    This suggests that, despite strong explanatory power across many settings, the two-factor view with the evaluated metrics does not yet fully explain generalization.
\end{enumerate}

\begin{table}[H]
\caption{
\textbf{Main results} (mixed settings).
Metric pairs and evaluation criteria are the same as in Table~\ref{tab:main_result_table}; here each vertical block reports results for \emph{a mixed setting}: pooling models trained from multiple sources into one 
collection.
{Cross-architecture} pools models across architectures, holding the dataset and optimizer fixed.
{Cross-optimizer \& cross-architecture} pools models across architectures and optimizers.
{Cross-dataset \& cross-architecture} pools models across architectures and datasets.
See details in Appendix~\ref{subap:mixed_setting}.
}
\vspace{-9pt}
\centering
\footnotesize
\setlength{\tabcolsep}{5pt}
\setlength{\fboxsep}{0.2pt}
\renewcommand{\arraystretch}{1.15}

\resizebox{0.99\linewidth}{!}{%
\begin{tabular}{
    @{}l
    *{3}{
        >{\raggedleft\arraybackslash}p{1.05cm}
        >{\centering\arraybackslash}p{1.50cm}
        !{\SCbodysep}
    }
    >{\raggedleft\arraybackslash}p{1.00cm}
    >{\centering\arraybackslash}p{1.40cm}
    @{}
}
\toprule

\multicolumn{1}{@{}c}{%
    \multirow{2}{*}{%
        \makecell[c]{%
            $\boldsymbol{(S,C)}$\\
            \textbf{Metric Pair}%
        }%
    }%
}
& \multicolumn{2}{c}{%
    \makecell[c]{%
        \multirow{2}{*}{\textbf{Cross-architecture}}\\
        {}\\
        {\scriptsize mixed four}
    }%
}
& \multicolumn{2}{c}{%
    \makecell[c]{%
        \textbf{Cross-optimizer}\\
        \textbf{\;\;\& Cross-architecture}\\
        {\scriptsize mixed eight}
    }%
}
& \multicolumn{4}{c}{%
    \makecell[c]{%
        \multirow{2}{*}{\textbf{Cross-dataset \& Cross-architecture}}\\
        {}\\
        {\scriptsize mixed three\qquad\qquad\qquad\qquad mixed six}
    }%
} \\[-1pt]

\cmidrule(lr){2-3}
\cmidrule(lr){4-5}
\cmidrule(lr){6-9}

& \multicolumn{1}{r}{\textbf{PCR}}
& \multicolumn{1}{c}{$\boldsymbol{R^2}\,(S,C)$}
& \multicolumn{1}{r}{\textbf{PCR}}
& \multicolumn{1}{c}{$\boldsymbol{R^2}\,(S,C)$}
& \multicolumn{1}{r}{\textbf{PCR}}
& \multicolumn{1}{c}{$\boldsymbol{R^2}\,(S,C)$}
& \multicolumn{1}{r}{\textbf{PCR}}
& \multicolumn{1}{c}{$\boldsymbol{R^2}\,(S,C)$}
\\[-3pt]


\midrule

\SCtrFunc 
& 4.1\% & \SCrsq{0.81}{(+,+)} 
& 11.5\%& \SCrsq{0.26}{(-,+)} 
& 2.3\% & \SCrsq{0.83}{(+,+)} 
& 2.2\% & \SCrsq{0.86}{(+,+)} \\
\SClmaxFunc     
& 4.1\% & \SCrsq{0.81}{(+,+)}  
& 11.6\%& \SCrsq{0.25}{(-,+)} 
& 2.2\% & \SCrsq{0.81}{(+,+)} 
& 2.9\% & \SCrsq{0.86}{(+,+)} \\
\SCtrfsFunc    
& 2.1\% & \SCrsq{0.88}{(+,+)}  
& 6.6\% & \SCrsq{0.32}{(+,+)}
& 3.0\% & \SCrsq{0.92}{(+,+)}  
& 2.7\% & \SCrsq{0.87}{(+,+)} \\
\SClmaxfsFunc    
& 1.7\% & \SCrsq{0.89}{(+,+)}  
& 6.1\% & \SCrsq{0.25}{(+,+)}
& 2.4\% & \SCrsq{0.92}{(+,+)}  
& 2.4\% & \SCrsq{0.87}{(+,+)} \\

\addlinespace[2pt]

\SCtrLtwo       
& 65.6\% &\SCrsq{0.18}{(-,+)}  
& 64.6\%& \SCrsq{0.01}{(-,-)} 
& 8.5\% & \SCrsq{0.81}{(+,+)} 
& 41.0\% & \SCrsq{0.05}{(+,+)} \\
\SClmaxLtwo   
& 62.3\% &\SCrsq{0.17}{(-,+)}  
& 63.5\%& \SCrsq{0.01}{(-,-)} 
& 9.8\% & \SCrsq{0.81}{(+,+)}  
& 39.8\% & \SCrsq{0.03}{(+,+)} \\
\SCtrfsLtwo      
& 29.9\% &\SCrsq{0.17}{(+,+)}  
& 27.0\%& \SCrsq{0.14}{(+,-)} 
& 9.2\% & \SCrsq{0.89}{(+,+)}  
& 30.7\% &\SCrsq{0.04}{(+,-)} \\
\SClmaxfsLtwo   
& 27.7\% & \SCrsq{0.19}{(+,+)}  
& 26.3\% & \SCrsq{0.05}{(+,-)} 
& 8.4\% & \SCrsq{0.88}{(+,+)} 
& 27.8\% & \SCrsq{0.11}{(+,-)} \\

\addlinespace[2pt]

\SCtrPath      
& 70.1\% & \SCrsq{0.11}{(-,+)}   
& 49.4\%& \SCrsq{0.61}{(-,+)} 
& 28.2\% & \SCrsq{0.35}{(-,+)} 
& 51.9\% & \SCrsq{0.04}{(+,-)}\\
\SClmaxPath      
& 68.5\% & \SCrsq{0.10}{(-,+)}  
& 49.1\%& \SCrsq{0.61}{(-,+)} 
& 30.8\% & \SCrsq{0.35}{(-,+)} 
& 51.6\% & \SCrsq{0.02}{(+,-)}\\
\SCtrfsPath     
& 36.8\% & \SCrsq{0.10}{(+,+)}  
& 25.8\%& \SCrsq{0.61}{(+,+)} 
& 17.1\% & \SCrsq{0.75}{(+,+)} 
& 36.1\% & \SCrsq{0.05}{(+,-)}\\
\SClmaxfsPath  
& 36.8\% & \SCrsq{0.11}{(+,+)}  
& 25.2\%& \SCrsq{0.61}{(+,+)} 
& 17.9\% & \SCrsq{0.67}{(+,+)}  
& 34.2\% & \SCrsq{0.12}{(+,-)} \\

\addlinespace[2pt]

\SCtrSpec        
& 66.8\% & \SCrsq{0.11}{(-,+)}  
& 47.5\%& \SCrsq{0.26}{(-,+)} 
& 43.4\% & \SCrsq{0.11}{(-,+)} 
& 53.3\% & \SCrsq{0.04}{(+,+)}  \\
\SClmaxSpec       
& 64.5\% & \SCrsq{0.11}{(-,+)}  
& 47.2\%& \SCrsq{0.26}{(-,+)} 
& 47.5\% & \SCrsq{0.11}{(-,+)} 
& 53.0\% & \SCrsq{0.02}{(+,+)} \\
\SCtrfsSpec       
& 28.2\% & \SCrsq{0.12}{(+,+)}  
& 19.6\%& \SCrsq{0.35}{(+,+)} 
& 14.8\% & \SCrsq{0.70}{(+,+)} 
& 32.7\% & \SCrsq{0.04}{(+,+)}  \\
\SClmaxfsSpec     
& 29.7\% & \SCrsq{0.14}{(+,+)}  
& 19.3\%& \SCrsq{0.30}{(+,+)}  
& 20.1\% & \SCrsq{0.61}{(+,+)} 
& 31.4\% & \SCrsq{0.11}{(+,+)}  \\[-1pt]

\bottomrule
\end{tabular}%
}
\label{tab:main_result_table_mixedsetting}
\end{table}

\subsection{Complementary Single-Factor Analysis}

To complement the two-factor joint view, we evaluate the explanatory power of single factors as well. 
Specifically, we assess
each of the four sharpness metrics and the four complexity metrics \emph{individually},
using an analogous order-based criterion and single-factor linear regression.
Observations~\ref{obs:larger_scale_much_better}, \ref{obs:scale_matter_more}, \ref{obs:func_norm_more_stable}, and~\ref{obs:param_norm_still_good} discussed above are also partially supported by the single-factor results; see Appendix~\ref{subap:single_factor_analysis} for details.

\section{Conclusion}
Focusing on the central question posed in the paper's title,
we quantitatively examined the reach of the two-factor view in modern deep networks.
In contrast to the pessimistic findings of previous studies on scalar generalization measures, including scalarized sharpness-complexity measures, our results show that suitably defined $\SC$ metric pairs can provide strong explanatory power across a range of settings. 
We find that finite-scale sharpness often exhibits much stronger explanatory power than its infinitesimal counterpart, with perturbation scale typically mattering more than the choice of Hessian statistic. 
Function norm offers more stable explanatory behavior than parameter norm in many settings, particularly in models without normalization and in mixed settings.
Despite being scale-variant, parameter norm remains highly effective in some training regimes, highlighting that the advantage of complexity representation depends on the setting. This observation is consistent with the practical effectiveness of weight decay in SGD training.
The remaining failures are also informative. 
An important next step is to investigate whether the unexplained regimes call for more refined metrics or for additional factors beyond sharpness and complexity. 
We hope these findings can guide further investigation toward a more comprehensive understanding of generalization.

\newpage
\bibliographystyle{plain}
\bibliography{refs}

@article{bartlett2021deep,
  title   = {Deep learning: A statistical viewpoint},
  author  = {Bartlett, Peter L. and Montanari, Andrea and Rakhlin, Alexander},
  journal = {Acta Numerica},
  volume  = {30},
  pages   = {87--201},
  year    = {2021},
}

@inproceedings{wei2020improved,
  title     = {Improved Sample Complexities for Deep Networks and
               Robust Classification via an All-Layer Margin},
  author    = {Wei, Colin and Ma, Tengyu},
  booktitle = {International Conference on Learning Representations},
  year      = {2020}
}

@inproceedings{zhang2017rethinking,
  title={Understanding Deep Learning Requires Rethinking Generalization},
  author={Zhang, Chiyuan and Bengio, Samy and Hardt, Moritz and Recht, Benjamin and Vinyals, Oriol},
  booktitle={International Conference on Learning Representations},
  year={2017}
}

@inproceedings{jiang2019fantastic,
  title={Fantastic Generalization Measures and Where to Find Them},
  author={Jiang, Yiding and Neyshabur, Behnam and Mobahi, Hossein and Krishnan, Dilip and Bengio, Samy},
  booktitle={International Conference on Learning Representations},
  year={2019}
}

@inproceedings{dziugaite2020robust,
  author    = {Dziugaite, Gintare Karolina and Drouin, Alexandre and
               Neal, Brady and Rajkumar, Nitarshan and Caballero, Ethan and
               Wang, Linbo and Mitliagkas, Ioannis and Roy, Daniel M.},
  title     = {In Search of Robust Measures of Generalization},
  booktitle = {Advances in Neural Information Processing Systems},
  volume    = {33},
  year      = {2020},
}

@inproceedings{andriushchenko2023modern,
  author    = {Andriushchenko, Maksym and Croce, Francesco and
               M{\"u}ller, Maximilian and Hein, Matthias and
               Flammarion, Nicolas},
  title     = {A Modern Look at the Relationship between Sharpness and Generalization},
  booktitle = {Proceedings of the 40th International Conference on Machine Learning},
  series    = {Proceedings of Machine Learning Research},
  volume    = {202},
  pages     = {840--902},
  publisher = {PMLR},
  year      = {2023},

  
}

@inproceedings{chaudhari2017entropysgd,
  title={{Entropy-SGD}: Biasing Gradient Descent Into Wide Valleys},
  author={Chaudhari, Pratik and Choromanska, Anna and Soatto, Stefano and LeCun, Yann and Baldassi, Carlo and Borgs, Christian and Chayes, Jennifer and Sagun, Levent and Zecchina, Riccardo},
  booktitle={International Conference on Learning Representations},
  year={2017},
}

@article{neyshabur2014implicit,
  title={In Search of the Real Inductive Bias: On the Role of Implicit Regularization in Deep Learning},
  author={Neyshabur, Behnam and Tomioka, Ryota and Srebro, Nathan},
  journal={arXiv preprint arXiv:1412.6614},
  year={2014}
}

@article{soudry2018implicit,
  title={The Implicit Bias of Gradient Descent on Separable Data},
  author={Soudry, Daniel and Hoffer, Elad and Nacson, Mor Shpigel and Gunasekar, Suriya and Srebro, Nathan},
  journal={Journal of Machine Learning Research},
  volume={19},
  number={70},
  pages={1--57},
  year={2018}
}

@inproceedings{gunasekar2018implicit,
  title={Implicit Bias of Gradient Descent on Linear Convolutional Networks},
  author={Gunasekar, Suriya and Lee, Jason D. and Soudry, Daniel and Srebro, Nathan},
  booktitle={Advances in Neural Information Processing Systems},
  year={2018}
}

@inproceedings{lyu2019gradient,
  title={Gradient Descent Maximizes the Margin of Homogeneous Neural Networks},
  author={Lyu, Kaifeng and Li, Jian},
  booktitle={International Conference on Learning Representations},
  year={2020}
}

@inproceedings{smith2021origin,
  title={On the Origin of Implicit Regularization in Stochastic Gradient Descent},
  author={Smith, Samuel L. and Dherin, Benoit and Barrett, David G. T. and De, Soham},
  booktitle={International Conference on Learning Representations},
  year={2021}
}

@inproceedings{tsuzuku2020normalized,
  title={Normalized Flat Minima: Exploring Scale Invariant Definition of Flat Minima for Neural Networks using {PAC-Bayesian} Analysis},
  author={Tsuzuku, Yusuke and Sato, Issei and Sugiyama, Masashi},
  booktitle={International Conference on Machine Learning},
  pages={9636--9647},
  year={2020},
  organization={PMLR}
}

@inproceedings{ghorbani2019investigation,
  title={An Investigation into Neural Net Optimization via Hessian Eigenvalue Density},
  author={Ghorbani, Behrooz and Krishnan, Shankar and Xiao, Ying},
  booktitle={International Conference on Machine Learning},
  pages={2232--2241},
  year={2019},
  organization={PMLR}
}

@inproceedings{petzka2021relative,
  title={Relative Flatness and Generalization},
  author={Petzka, Henning and Kamp, Michael and Sch{\"a}fer, Linus and Salam, Asif and Fischer, Christian},
  booktitle={Advances in Neural Information Processing Systems},
  volume={34},
  pages={10420--10432},
  year={2021},
}

@article{hochreiter1997flat,
  title={Flat Minima},
  author={Hochreiter, Sepp and Schmidhuber, J{\"u}rgen},
  journal={Neural Computation},
  volume={9},
  number={1},
  pages={1--42},
  year={1997}
}

@inproceedings{keskar2017large,
  title={On Large-Batch Training for Deep Learning: Generalization Gap and Sharp Minima},
  author={Keskar, Nitish Shirish and Mudigere, Dheevatsa and Nocedal, Jorge and Smelyanskiy, Mikhail and Tang, Ping Tak Peter},
  booktitle={International Conference on Learning Representations},
  year={2017}
}

@article{golowich2020size,
  title={Size-independent sample complexity of neural networks},
  author={Golowich, Noah and Rakhlin, Alexander and Shamir, Ohad},
  journal={Information and Inference: A Journal of the IMA},
  volume={9},
  number={2},
  pages={473--501},
  year={2020},
  publisher={Oxford University Press}
}

@inproceedings{arora2018stronger,
  title     = {Stronger Generalization Bounds for Deep Nets via a Compression Approach},
  author    = {Arora, Sanjeev and Ge, Rong and Neyshabur, Behnam and Zhang, Yi},
  booktitle = {Proceedings of the 35th International Conference on Machine Learning},
  series    = {Proceedings of Machine Learning Research},
  volume    = {80},
  pages     = {254--263},
  year      = {2018},
  publisher = {PMLR},
}

@inproceedings{dinh2017sharp,
  title={Sharp Minima Can Generalize For Deep Nets},
  author={Dinh, Laurent and Pascanu, Razvan and Bengio, Samy and Bengio, Yoshua},
  booktitle={International Conference on Machine Learning},
  pages={1019--1028},
  year={2017}
}

@inproceedings{liang2019fisherrao,
  author    = {Liang, Tengyuan and Poggio, Tomaso and Rakhlin, Alexander and
               Stokes, James},
  title     = {{Fisher-Rao} Metric, Geometry, and Complexity of Neural Networks},
  booktitle = {Proceedings of the Twenty-Second International Conference on Artificial Intelligence and Statistics},
  series    = {Proceedings of Machine Learning Research},
  volume    = {89},
  pages     = {888--896},
  publisher = {PMLR},
  year      = {2019},
}

@inproceedings{kaur2023maximum,
  title={On the Maximum Hessian Eigenvalue and Generalization},
  author={Kaur, Simran and Cohen, Jeremy M. and Lipton, Zachary C.},
  booktitle={Proceedings of the Thirty-Ninth Conference on Uncertainty in Artificial Intelligence},
  year={2023},
}

@article{liu2022regularizing,
  title={Regularizing Deep Neural Networks with Stochastic Estimators of Hessian Trace},
  author={Liu, Yucong and Yu, Shixing and Lin, Tong},
  journal={arXiv preprint arXiv:2208.05924},
  year={2022}
}

@inproceedings{andriushchenko2022towards,
  title={Towards Understanding Sharpness-Aware Minimization},
  author={Andriushchenko, Maksym and Flammarion, Nicolas},
  booktitle={International Conference on Machine Learning},
  year={2022},
}

@inproceedings{foret2020sharpness,
  title={Sharpness-aware Minimization for Efficiently Improving Generalization},
  author={Foret, Pierre and Kleiner, Ariel and Mobahi, Hossein and Neyshabur, Behnam},
  booktitle={International Conference on Learning Representations},
  year={2020}
}

@inproceedings{kwon2021adapsam,
  title={ASAM: Adaptive Sharpness-Aware Minimization for Scale-Invariant Learning of Deep Neural Networks},
  author={Kwon, Jungmin and Kim, Jeongmin and Park, Hyunseo and Choi, In Kwon},
  booktitle={International Conference on Machine Learning},
  year={2021},
}

@inproceedings{mcallester1998some,
  title={Some pac-bayesian theorems},
  author={McAllester, David A},
  booktitle={Proceedings of the eleventh annual conference on Computational learning theory},
  pages={230--234},
  year={1998}
}

@inproceedings{mcallester1999pac,
  title={PAC-Bayesian model averaging},
  author={McAllester, David A},
  booktitle={Proceedings of the twelfth annual conference on Computational learning theory},
  pages={164--170},
  year={1999}
}

@inproceedings{mcallester2003simplified,
  title={Simplified {PAC-Bayesian} Margin Bounds},
  author={McAllester, David A.},
  booktitle={Learning Theory and Kernel Machines (COLT)},
  year={2003},
}

@inproceedings{dziugaite2017computing,
  title={Computing Nonvacuous Generalization Bounds for Deep (Stochastic) Neural Networks with Many More Parameters than Training Data},
  author={Dziugaite, Gintare Karolina and Roy, Daniel M.},
  booktitle={Proceedings of the Thirty-Third Conference on Uncertainty in Artificial Intelligence},
  year={2017}
}

@inproceedings{dziugaite2021role,
  title={On the Role of Data in {PAC-Bayes} Bounds},
  author={Dziugaite, Gintare Karolina and Hsu, Kyle and Gharbieh, Waseem and Arpino, Gabriel and Roy, Daniel},
  booktitle={Proceedings of the 24th International Conference on Artificial Intelligence and Statistics (AISTATS)},
  pages={604--612},
  volume={130},
  year={2021},
  organization={PMLR},
}

@article{alquier2024userfriendly,
  title   = {User-friendly introduction to PAC-Bayes bounds},
  author  = {Alquier, Pierre},
  journal = {Foundations and Trends in Machine Learning},
  volume  = {17},
  number  = {2},
  pages   = {174--303},
  year    = {2024},
}

@article{guedj2019primer,
  title={A primer on PAC-Bayesian learning},
  author={Guedj, Benjamin},
  journal={arXiv preprint arXiv:1901.05353},
  year={2019}
}

@article{neyshabur2017exploring,
  title={Exploring generalization in deep learning},
  author={Neyshabur, Behnam and Bhojanapalli, Srinadh and McAllester, David and Srebro, Nati},
  journal={Advances in neural information processing systems},
  volume={30},
  year={2017}
}

@inproceedings{neyshabur2018pac,
  title={A PAC-Bayesian Approach to Spectrally-Normalized Margin Bounds for Neural Networks},
  author={Neyshabur, Behnam and Bhojanapalli, Srinadh and McAllester, David and Srebro, Nathan},
  booktitle={International Conference on Learning Representations},
  year={2018}
}

@inproceedings{neyshabur2015norm,
  title={Norm-based capacity control in neural networks},
  author={Neyshabur, Behnam and Tomioka, Ryota and Srebro, Nathan},
  booktitle={Conference on Learning Theory},
  year={2015},
}

@article{bartlett2002rademacher,
  author  = {Bartlett, Peter L. and Mendelson, Shahar},
  title   = {Rademacher and Gaussian Complexities: Risk Bounds and Structural Results},
  journal = {Journal of Machine Learning Research},
  year    = {2002},
  volume  = {3},
  pages   = {463--482},
}

@article{bartlett2017spectrally,
  title={Spectrally-normalized margin bounds for neural networks},
  author={Bartlett, Peter L. and Foster, Dylan J. and Telgarsky, Matus},
  journal={Advances in neural information processing systems},
  volume={30},
  year={2017}
}

@inproceedings{krogh1991simpleweightdecay,
 author = {Krogh, Anders and Hertz, John},
 booktitle = {Advances in Neural Information Processing Systems},
 title = {A Simple Weight Decay Can Improve Generalization},
 volume = {4},
 year = {1991}
}

@inproceedings{loshchilov2019decoupled,
  title={Decoupled Weight Decay Regularization},
  author={Loshchilov, Ilya and Hutter, Frank},
  booktitle={International Conference on Learning Representations},
  year={2019}
}

@inproceedings{zhang2019three,
  title={Three Mechanisms of Weight Decay Regularization},
  author={Zhang, Guodong and Wang, Chaoqi and Xu, Bowen and Grosse, Roger},
  booktitle={International Conference on Learning Representations},
  year={2019}
}

@inproceedings{dangelo2024why,
  title     = {Why Do We Need Weight Decay in Modern Deep Learning?},
  author    = {D'Angelo, Francesco and Andriushchenko, Maksym and Varre, Aditya and Flammarion, Nicolas},
  booktitle = {Advances in Neural Information Processing Systems},
  volume    = {37},
  year      = {2024},
}

@article{hutchinson1989,
  author  = {Hutchinson, M. F.},
  title   = {A stochastic estimator of the trace of the influence matrix
             for {Laplacian} smoothing splines},
  journal = {Communications in Statistics - Simulation and Computation},
  volume  = {18},
  number  = {3},
  pages   = {1059--1076},
  year    = {1989},
}

@article{lanczos1950,
  author  = {Lanczos, Cornelius},
  title   = {An iteration method for the solution of the eigenvalue problem
             of linear differential and integral operators},
  journal = {Journal of Research of the National Bureau of Standards},
  volume  = {45},
  number  = {4},
  pages   = {255--282},
  year    = {1950},
}

\newpage
\appendix

\section*{Appendix}

\section{Theoretical motivation of the two-factor view}
\label{ap:pac_bayes_motivation}

In this section, we explain the theoretical motivation of the two-factor view.

\subsection{PAC-Bayes Bounds}

PAC-Bayes bounds theoretically characterize the generalization gap in terms of sharpness and complexity. 
A typical PAC-Bayes bound relates the test performance of a randomized predictor to its training performance and complexity; see Section 3.2.2 in~\cite{alquier2024userfriendly}.

Let \(Z=\{z_i\}_{i=1}^m\) be a training set containing \(m\) independent samples drawn from the data distribution \(\mathcal{D}\). Let \(\mathcal{P}\) be a prior distribution over predictors, chosen independently of \(Z\), and let \(\mathcal{Q}\) be a posterior distribution that may depend on \(Z\). Assuming that the loss is bounded in \([0,1]\), with probability at least \(1-\delta\) over the random draw of \(Z\), the following inequality holds for all posterior distributions \(\mathcal{Q}\):
\begin{equation}
\label{eq:general_bound_model_space}
L_{\mathcal{D}}(\mathcal{Q})
\quad \leq \quad
\widehat{L}_Z(\mathcal{Q})
+
\sqrt{
\frac{
\mathrm{KL}(\mathcal{Q}\Vert\mathcal{P})
+\ln\!\left(\frac{2\sqrt{m}}{\delta}\right)
}{2m}
}.
\end{equation}
Here,
\begin{equation}
\label{def:expected_pop_loss}
L_{\mathcal{D}}(\mathcal{Q})
:=
\mathbb{E}_{h\sim\mathcal{Q}}\ell(h;\mathcal{D})
=
\mathbb{E}_{h\sim\mathcal{Q}}
\mathbb{E}_{z\sim\mathcal{D}}
\bigl[\ell(h;z)\bigr]
\end{equation}
is the \emph{expected} population loss (or test loss) of a predictor \(h\) sampled from \(\mathcal{Q}\). 
Similarly,
\begin{equation}
\label{def:expected_emp_loss}
\widehat{L}_Z(\mathcal{Q})
:=
\mathbb E_{h \sim \mathcal{Q}} 
\big[\ell(h;Z)\big]
=
\mathbb{E}_{h\sim\mathcal{Q}}
\big[
\tfrac{1}{m}
\sum\nolimits_{i=1}^{m}
\ell(h;z_i)
\big]
\end{equation}
is the \emph{expected} empirical loss of a predictor \(h\) sampled from \(\mathcal{Q}\).

The KL divergence \(\mathrm{KL}(\mathcal{Q}\Vert\mathcal{P})\) measures how different the posterior distribution \(\mathcal{Q}\), which represents the trained predictors, is from the data-independent prior distribution \(\mathcal{P}\). It therefore serves as a measure of model complexity.

In simple terms, the PAC-Bayes bound~(\ref{eq:general_bound_model_space}) states that \emph{the test loss is small when two quantities are small}: 
the expected empirical loss \(\widehat{L}_Z(\mathcal{Q})\) and the complexity term \(\mathrm{KL}(\mathcal{Q}\Vert\mathcal{P})\).
The bound also becomes tighter as the number of training samples \(m\) becomes larger.

As mentioned in Section~\ref{subsec:parameterization_ambiguity},
for deep networks, the model $h$ is usually represented through parameterization, i.e., $h=f_w$. 
This leads to two levels of description.
In \textbf{parameter space}, we denote the prior and posterior distributions by \(\calP_{w}\) and \(\Q_{w}\).
Through the map $w\mapsto f_w$,
they induce corresponding \textbf{function space} distributions, denoted by \(\calP_f\) and \(\Q_f\). 


\subsection{Origin of the Two Factors}
\label{subap:origin_two_factor}

PAC-Bayes bounds control the expected population loss $L_{\mathcal{D}}(\Q)$ mainly through two terms: 
\begin{center}
\emph{the expected empirical loss} $\hat{L}_{Z}(\Q)$ and 
\emph{the posterior-prior deviation} 
$\mathrm{KL}(\Q\|\calP)$.
\end{center}

The first term is closely related to the \textbf{sharpness} factor.

Applying Taylor expansion in the parameter space to $\hat{L}_{Z}(\Q)$, we obtain
\begin{equation}\label{eq:sharpness_taylorexpansion}
\widehat{L}_Z(\Q_{w})
\,=\,
\mathbb E_{w \sim \mathcal{Q}_{w}} 
\big[\ell(f_{w};Z)\big]
\;
=
\ell(f_{\hat w};Z)
+ 
\tfrac{\sigma^2}{2}
\operatorname{tr}
\Big(
\nabla_w^2 \ell(f_{\hat w};Z)
\Big)
+o(\sigma^2)
,
\end{equation}
where the posterior $\mathcal{Q}_{w}$ is centered at the learned solution 
$\hat w$ with covariance $\Sigma_w=\sigma^2I$; see the proof below.
This shows that, when the training loss $\ell(f_{\hat w};Z)$ becomes sufficiently small, the dominant term depends on the \emph{local curvature} around the learned solution $\hat w$ , which provides intuition for the sharpness factor.

The second term is closely related to the \textbf{complexity} factor: 
KL divergence $\mathrm{KL}(\Q\|\calP)$  measures how much deviation is needed to move from the prior to the learned posterior.

KL divergence admits a closed form under a Gaussian posterior and prior. 
Suppose $\Q=\mathcal{N}(\mu_\Q,  \Sigma_\Q)$ and
$\calP=\mathcal{N}(\mu_\calP, \Sigma_\calP)$, then
\begin{equation}
\label{eq: KL_closedform}
\mathrm{KL}(\Q\| \calP)
=
\frac12
\Big(
-\!d 
+ \mathrm{tr}
\big(\Sigma_\Q\;{\Sigma_\calP}^{-1}
\big)
+ \ln\frac{\det({\Sigma_\calP})}{\det(\Sigma_\Q)}
+ 
(\mu_{\Q}-\mu_{\calP})^\top {\Sigma_\calP}^{-1} (\mu_{\Q}-\mu_{\calP})
\Big)\;,
\end{equation}
where \(d\) is the dimension of the distribution.

In particular, suppose that the parameter-level prior and posterior are Gaussian with the same isotropic covariance
\(\Sigma_{\Q_w}=\Sigma_{\calP_w}=\sigma^2 I\).
Let the posterior be centered at the learned solution $\mu_{\Q_w}=\hat w$ and the prior at the origin $\mu_{\calP_w}=0$. Then the parameter-level KL divergence reduces to the squared weight norm up to a constant factor:
\begin{equation}\label{eq: KL_closedform_weightdecay}
\mathrm{KL}(\Q_{w}\|\calP_{w})
=\frac{\|\hat w\|^2}{2\sigma^2}.
\end{equation}
This corresponds to the \emph{weight decay} in SGD training, and thus explains why weight decay can be viewed as a simple parameter-space complexity regularizer.

\begin{proof}[Proof of Eq.~(\ref{eq:sharpness_taylorexpansion})]
Let the posterior $\Q_{ w}$ be centered at the learned solution $\hat w$ with covariance $\Sigma_ w=\sigma^2I$.
The expected empirical loss can be written as
\begin{equation*}\label{sharpness_taylor_expansion}
\begin{aligned}
\hat L_{Z}(\Q_{ w})
:=
\mathbb{E}_{ w\sim \Q_ w}\big[\ell(f_ w;Z)\big] \stackrel{ w=\hat w+\Delta w}{=}
\mathbb{E}_{\Delta w\sim \widetilde{\Q}_ w}
\big[\ell(f_{\hat w+\Delta w};Z)\big],
\end{aligned}
\end{equation*}
where \(\widetilde{\Q}_ w\) is the distribution of
\(\Delta w= w-\hat w\), satisfying
\[
\mathbb{E}_{\widetilde{\Q}_ w}[\Delta w]=0\;,
\quad
\mathbb{E}_{\widetilde{\Q}_ w}[\Delta w\Delta w^\top]=\sigma^2I.
\]
Applying a second-order Taylor expansion around \(\hat w\), we obtain
\begin{equation*}
\begin{aligned}
\hat L_{Z}(\Q_{ w})
&=
\mathbb{E}_{\Delta w\sim \widetilde{\Q}_ w}
\Big[\ell(f_{\hat w+\Delta w};Z)\Big]\\
&=
\mathbb{E}_{\Delta w\sim \widetilde{\Q}_ w}
\Big[
\ell(f_{\hat w};Z)
+
\Delta w^\top \nabla_ w \ell(f_{\hat w};Z)
+
\tfrac{1}{2}
\Delta w^\top
\nabla_ w^2 \ell(f_{\hat w};Z)
\Delta w
+
o(\|\Delta w\|^2)
\Big] \\
&=
\ell(f_{\hat w};Z)
+0+
\tfrac{1}{2}
\operatorname{tr}
\left(
\nabla_ w^2 \ell(f_{\hat w};Z)
\,
\mathbb{E}_{\Delta w\sim \widetilde{\Q}_ w}
[\Delta w\Delta w^\top]
\right) 
+ \mathbb{E}_{\Delta w\sim \widetilde{\Q}_{ w}} [o(\|\Delta w\|^2)]\\
&=
\underbrace{\ell(f_{\hat w};Z)}_{\text{training loss}} 
+ \quad
\tfrac{\sigma^2}{2}
\operatorname{tr}
\Big(
\nabla_ w^2 \ell(f_{\hat w};Z)
\Big)
+o(\sigma^2).
\end{aligned}
\end{equation*}
\end{proof}

\subsection{Finite-scale sharpness}
\label{subap:pac_bayes_bayes_sharpness}

For convenience, denote
\(
T:=\operatorname{diag}(|w|).
\)
The adaptive Hessian defined in Eq.~(\ref{def:adap_hessian}) is then
\[
\mathrm{H}=T\nabla_w^2\ell(f_w;Z)T.
\]

\paragraph{Finite-scale Hessian trace.}
As introduced in Section~\ref{sec:metrics_review},
the finite-scale Hessian trace is
\begin{equation}\label{def:trace_fs2}
\SCsharpTrFS(w;\sigma)
:=
\mathbb{E}_{\epsilon}
\left[
\frac{
\ell(f_{w+\sigma\epsilon};Z)-\ell(f_w;Z)
}{\sigma^2}
\right],
\qquad
\epsilon=T\delta,
\quad
\delta\sim\mathcal{N}(0,I).
\end{equation}
We show that this quantity approaches one-half of the
infinitesimal adaptive Hessian trace as the perturbation
scale approaches zero.

\begin{proof}
Let \(\Delta w=\sigma T \delta\). Then
\[
\mathbb{E}[\Delta w]=0,
\qquad
\mathbb{E}[\Delta w\Delta w^\top]=\sigma^2T^2.
\]
Taking the expectation of the Taylor expansion, we obtain
\begin{equation*}
\begin{aligned}
\mathbb{E}_{\Delta w}\big[\ell(f_{w+\Delta w};Z)\big]
&=
\mathbb{E}_{\Delta w}
\Big[
\ell(f_w;Z)
+
\Delta w^\top\nabla_w\ell(f_w;Z)
+
\tfrac12\Delta w^\top\nabla_w^2\ell(f_w;Z)\Delta w
+
o(\| \Delta w\|_2^2) 
\Big]
\\
&=
\ell(f_w;Z)
+
0
+
\tfrac12
\operatorname{tr}
\left(
\nabla_w^2\ell(f_w;Z)\,
\mathbb{E}[\Delta w\Delta w^\top]
\right)
+
o(\sigma^2)
\\
&=
\ell(f_w;Z)
+
\tfrac{\sigma^2}{2}
\operatorname{tr}
\left(
\nabla_w^2\ell(f_w;Z)T^2
\right)
+
o(\sigma^2)
\\
&=
\ell(f_w;Z)
+
\tfrac{\sigma^2}{2}\operatorname{tr}(H)
+
o(\sigma^2).
\end{aligned}
\end{equation*}
Subtracting \(\ell(f_w;Z)\) and dividing by \(\sigma^2\) yields
\[
\SCsharpTrFS(w;\sigma)
=
\tfrac12\mathrm{tr(H)}+o(1)
\;\longrightarrow\;
\tfrac12\mathrm{tr(H)}
\qquad
\text{as }\sigma\to0.
\]
Thus, up to a constant factor, the finite-scale
Hessian trace approaches its infinitesimal counterpart.
\end{proof}

\paragraph{Finite-scale Hessian top eigenvalue.}
The finite-scale Hessian top eigenvalue is
\begin{equation}\label{def:topeigen_fs2}
\SCsharpMaxFS(w;\sigma)
:=
\max_{\|v\|_2=1}
\frac{
\ell(f_{w+\sigma Tv};Z)
+
\ell(f_{w-\sigma Tv};Z)
-
2\ell(f_w;Z)
}{\sigma^2}.
\end{equation}
We show that this quantity approaches the infinitesimal
adaptive Hessian top eigenvalue as the perturbation scale
approaches zero.

\begin{proof}
For any unit vector \(v\), a second-order Taylor expansion gives
\begin{equation*}
\ell(f_{w\pm\sigma Tv};Z)
=
\ell(f_w;Z)
\pm
\sigma v^\top T\nabla_w\ell(f_w;Z)
+
\tfrac{\sigma^2}{2}v^\top Hv
+
o(\sigma^2).
\end{equation*}
Adding the two expansions cancels the linear terms, yielding
\begin{equation*}
\frac{
\ell(f_{w+\sigma Tv};Z)
+
\ell(f_{w-\sigma Tv};Z)
-
2\ell(f_w;Z)
}{\sigma^2}
=
v^\top Hv+o(1).
\end{equation*}
Taking the maximum over unit vectors and using the
Rayleigh characterization, we obtain
\begin{equation*}
\begin{aligned}
\SCsharpMaxFS(w;\sigma)
&=
\max_{\|v\|_2=1}v^\top Hv+o(1)
\\
&=
\lambda_{\max}\mathrm{(H)}+o(1)
\;\longrightarrow\;
\lambda_{\max}\mathrm{(H)}
\qquad
\text{as }\sigma\to0.
\end{aligned}
\end{equation*}
\end{proof}

\subsection{Further interpretations of the function norm}
\label{ap:func_norm}

The complexity term $\mathrm{KL}(\Q\|\mathcal{P})$ in PAC-Bayes bounds is fundamentally \emph{defined
over distributions of predictors} rather than over a particular
parameter representation. However, in deep networks, the prior and
posterior are commonly specified in parameter space, leading to the
parameter-level complexity
\(
\textrm{KL}\!\left(
\Q_{w}\|\mathcal{P}_{w}
\right).
\)
Because the map from parameters to predictors $w \mapsto f_{w} $ is generally non-injective in deep networks,
this quantity may distinguish parameter distributions that induce the
same predictive behavior.

To obtain a complexity metric defined closer to function space, we
map parameters to their centered function outputs on 
a fixed reference dataset \(D_{\mathrm{ref}}\!=\!\{(\bfx_i,y_i)\}_{i=1}^{n}\). Let \(Z_w\) denote the logit matrix:
$$
\scalebox{0.95}{$\displaystyle
Z_w(D_{\mathrm{ref}})
:=
\begin{bmatrix}
f_w(\bfx_1)^\top\\
\vdots\\
f_w(\bfx_n)^\top
\end{bmatrix}
\in
\mathbb{R}^{n\times c},
$}
$$
where \(f_w(\bfx_i)\) is the \(c\)-class logit vector for \(\bfx_i\).
Softmax probabilities are invariant under the transformation
\(Z_w \mapsto Z_w+\mathbf{e}\mathbf{1}_c^\top\)
for any \(\mathbf{e}\in\mathbb{R}^{n}\).
We remove this additive ambiguity using the class-centering projection
$\Pi_c\!=\!I_c\!-\!\frac{1}{c}\mathbf{1}_c\mathbf{1}_c^\top$. The \emph{class-centered logit matrix} is then defined as
$$
\scalebox{0.95}{$\displaystyle
\widetilde Z_w(D_{\mathrm{ref}})
:=
Z_w(D_{\mathrm{ref}})\Pi_c
=
\begin{bmatrix}
\widetilde f_w(\bfx_1)^\top\\
\vdots\\
\widetilde f_w(\bfx_n)^\top
\end{bmatrix}
\in
\mathbb{R}^{n\times c},
$}
$$
where 
\(
\widetilde f_w(\bfx_i)
\!=\!
f_w(\bfx_i)
-
\frac{1}{c}
\mathbf{1}_c\mathbf{1}_c^\top f_w(\bfx_i) .
\) 

Define the centered function-output lifting map as
\begin{equation}
\label{eq:centered_function_lifting_pac_bayes}
\widetilde F_{D_{\mathrm{ref}}}( w)
:=
\operatorname{vec}
\left(
\widetilde Z_{ w}(D_{\mathrm{ref}})
\right)
=
\operatorname{vec}\left(
\begin{bmatrix}
\widetilde f_{ w}(\bfx_1)^{\top}\\
\vdots\\
\widetilde f_{ w}(\bfx_n)^{\top}
\end{bmatrix}\right).
\end{equation}
Through the lifting map in Eq.~(\ref{eq:centered_function_lifting_pac_bayes}), the parameter-space
prior and posterior induce corresponding distributions over the centered
function-outputs:
\begin{equation}
\label{eq:centered_function_pushforwards}
\mathcal{P}_f^{D_{\mathrm{ref}}}
:=
\left(
\widetilde F_{D_{\mathrm{ref}}}
\right)_{\#}\mathcal{P}_{ w},
\qquad
\Q_f^{D_{\mathrm{ref}}}
:=
\left(
\widetilde F_{D_{\mathrm{ref}}}
\right)_{\#}\Q_{ w}.
\end{equation}
Therefore, a natural \textbf{function-space counterpart} of the parameter-space complexity
is 
\begin{equation}
\label{eq:function_space_kl}
\mathrm{KL}\!\left(
\Q_f^{D_{\mathrm{ref}}}
\,\|\,
\mathcal{P}_f^{D_{\mathrm{ref}}}
\right).
\end{equation}
Because the centered function-output distributions are obtained by applying a
deterministic map to the parameter distributions, the data-processing
inequality gives
\begin{equation*}
\mathrm{KL}\!\left(
\Q_f^{D_{\mathrm{ref}}}
\,\|\,
\calP_f^{D_{\mathrm{ref}}}
\right)
\leq
\mathrm{KL}\!\left( 
\Q_{ w}
\,\|\,
\calP_{ w}
\right).
\end{equation*}
The function-space KL divergence~(\ref{eq:function_space_kl}) \emph{discards distinctions between
parameter distributions} that disappear after mapping them to the same
predictor on the reference inputs.

Directly estimating the function-level KL divergence~(\ref{eq:function_space_kl}) can be computationally demanding. 
We therefore consider a \textit{simplified proxy}. 
The centered function outputs lie in the subspace
\[
\mathcal{H}_{D_{\mathrm{ref}}}
:=
\left\{
\operatorname{vec}(U)
:
U \in \mathbb{R}^{n \times c},\; U\mathbf{1}_c=0
\right\}.
\]
Suppose that the induced function-space posterior and prior are approximated on this subspace by Gaussian distributions with a shared isotropic covariance, and that the prior is centered at zero. We write these approximations as follows:
\[
\Q_f^{D_{\mathrm{ref}}}
\approx
\mathcal{N}_{\mathcal{H}_{D_{\mathrm{ref}}}}
\left(
\mu_f,\sigma_f^2 I
\right),
\qquad
\calP_f^{D_{\mathrm{ref}}}
\approx
\mathcal{N}_{\mathcal{H}_{D_{\mathrm{ref}}}}
\left(
0,\sigma_f^2 I
\right).
\]
Under these assumptions, according to Eq.~(\ref{eq: KL_closedform}), 
the function-space KL divergence~(\ref{eq:function_space_kl}) reduces to
\begin{equation}
\label{eq:simplified_centered_function_kl}
\mathrm{KL}\!\left(
\Q_f^{D_{\mathrm{ref}}}
\,\|\,
\calP_f^{D_{\mathrm{ref}}}
\right)
=
\frac{\left\|
\mu_f
\right\|_2^2}{2\sigma_f^2}
.
\end{equation}
When the parameter posterior is centered at the learned
solution (i.e., $\Q_{ w}$ has mean $\hat w$), its induced function-level posterior mean can be
approximated by the centered function-outputs at the trained model:
\[
\mu_f
=
\mathbb{E}_{ w\sim\Q_{ w}}
\left[
\widetilde F_{D_{\mathrm{ref}}}( w)
\right]
\approx
\widetilde F_{D_{\mathrm{ref}}}(\hat w).
\]
As a result, 
\begin{equation}
\label{eq:function_space_kl_proxy}
\mathrm{KL}\!\left(
\Q_f^{D_{\mathrm{ref}}}
\,\|\,
\calP_f^{D_{\mathrm{ref}}}
\right)
\approx
\frac{1}{2\sigma_f^2}
\|\widetilde F_{D_{\mathrm{ref}}}(\hat w)\|_2^2.
\end{equation}
Recall that the \textbf{function norm} in Section~\ref{subsec:complexity_metrics} is defined as
\begin{equation*}
\scalebox{0.95}{$\displaystyle
\mathrm{func\;norm}
:=
\bigl\| \widetilde Z_w(D_{\mathrm{ref}}) \bigr\|_{F}.
$}
\end{equation*}
We use the unsquared counterpart of the norm in Eq.~(\ref{eq:function_space_kl_proxy}), omitting the fixed positive factor, as a function-level complexity metric. This preserves Pareto orderings, although the corresponding linear-regression fit may differ.

This centered function-output norm can be viewed as a practical \textbf{\emph{proxy}} for
a simplified function-space posterior-prior KL divergence. It is not
an exact evaluation of~(\ref{eq:function_space_kl}); rather, it is obtained under the
approximations of a shared isotropic
function-space covariance, a zero-centered function prior, and a concentrated posterior.


\newpage
\section{Detailed explanation of the evaluation criteria}
\label{ap:diagnostic_criteria}

Pareto analysis and linear regression provide complementary assessments of the explanatory power of an evaluated metric pair.

PCR measures violations of the monotonic ordering predicted by the two-factor view (\emph{lower is better}), while $R^2$ evaluates how well a linear model captures the relationship (\emph{higher is better}).

The possible outcomes of interpreting their results jointly are summarized in Table~\ref{tab:joint_diagnostic_interpretation} and discussed below.


\begin{enumerate}
[   label=\roman*.,  
    leftmargin=3.0em, labelsep=0.5em,
    topsep=3pt,
    itemsep=2pt,
    parsep=0pt,
    partopsep=0pt
]

\item \emph{Low PCR, high \(R^2\), and positive regression coefficients.}
This is the ideal case.
The metric pair induces an ordering that agrees with the generalization ordering,
and the relationship is well captured by a linear model.
Together, these diagnostics provide strong evidence that the evaluated metric pair has strong explanatory power for generalization.

\item \emph{Low PCR, low \(R^2\), and positive regression coefficients.}
The metric pair induces an ordering that is consistent with the two-factor view,
but the relationship is not well captured by a linear model.
In this case, poor regression results do not necessarily weaken the two-factor view, which may retain strong explanatory power even when the dependence of generalization on the evaluated metrics is nonlinear.

\item \emph{High PCR, high \(R^2\), and positive regression coefficients.} 
The regression captures a strong global linear trend, but the high PCR reveals frequent violations of the expected monotonic ordering.
Because \(R^2\) describes the quality of an \emph{approximate} global linear fit, 
small residuals can violate the ordering of nearby models and worsen PCR while barely affecting \(R^2\). Thus, high \(R^2\) does not guarantee low PCR.
In this case, the linear trend captures part of the relationship, while important information may still be missing.

\item \emph{High PCR and low \(R^2\), or any negative regression coefficient.}
High PCR together with low \({R^2}\) indicates that the metric pair is neither
order-consistent nor well described by a linear model.
A negative regression coefficient constitutes an additional sign failure,
because it contradicts the expected direction of the corresponding sharpness
or complexity effect.
This indicates that the evaluated metrics are poor realizations of the two factors 
in the given setting. 
\end{enumerate}

\paragraph{Greater emphasis on PCR.}
The joint interpretation \emph{places greater weight on} PCR than on \(R^2\), because PCR does not rely on a linear modeling assumption. 
For example, an evaluation result with low PCR but low \(R^2\) is still considered favorable: it suggests strong agreement with the monotonic ordering, even though the relationship is not well captured by a linear model. 

\paragraph{Why retain linear regression assessments when PCR takes priority?}
Although we prioritize PCR when evaluating explanatory power, linear regression provides additional information about the form of the relationship beyond pairwise ordering.
This may help guide regularization, where selected quantities are often added to the training objective as a weighted sum; see the distinction between regularization and explanation in Section~\ref{sec:distinct_explanation_regularization}.
Low PCR together with high $R^2$ and positive coefficients supports a simple additive approximation, motivating the corresponding weighted sum as a promising candidate training penalty.
With low PCR but a poor linear fit, the monotonic ordering remains informative even though a fixed weighted sum does not adequately capture the relationship.
Therefore, linear regression \emph{plays a complementary role}: it helps assess how the two factors might be combined, providing additional information that PCR alone cannot supply.

\section{Experimental protocol} 
\label{ap:experiment_protocol}

We report several implementation choices that are important for conducting
the experiments.

\begin{enumerate}
[   
    leftmargin=1.5em, labelsep=0.5em,
    topsep=2pt,
    itemsep=2pt,
    parsep=0pt,
    partopsep=0pt
]
    \item \textbf{Broad hyperparameter sweeps.}
    For each individual setting in Table~\ref{tab:main_result_table} and Table~\ref{tab:trained_by_adamw}, we perform a broad hyperparameter sweep over learning rate, batch size, weight decay, and momentum, and obtain at least 48 trained models.
    The specific hyperparameter ranges may differ across architectures to
    account for differences in training stability and optimization behavior.

    \item \textbf{Filtering under-trained models.}
    To avoid confounding effects from under-trained models, we 
    retain well-trained models that achieve sufficiently small training loss, using a threshold of \(\mathrm{train\,loss}\!\leq\!0.01\).
    Runs that do not satisfy this criterion are not used in the analytical assessment.    
    This filtering step helps ensure that the measured generalization gap
    primarily reflects the variations in generalization, rather than
    artifacts of incomplete training or optimization failure.
    For each individual setting in Table~\ref{tab:main_result_table} and Table~\ref{tab:trained_by_adamw}, we {retain at least 24 well-trained models} for analytical evaluation. 
    This ensures that the Pareto analysis and linear regression are
    conducted on a diverse collection of well-trained models.

    \item \textbf{Fixed reference data for metric computation.}
    The complexity metric, function norm, requires reference data. It is
    computed using the same fixed reference set for all trained models.
    This reduces sampling noise and ensures that differences in metric values
    are attributable to the trained models, rather than to variations in the
    reference data used for evaluation.


    %
\end{enumerate}

\paragraph{}
Table~\ref{tab:trained_model_counts} summarizes the counts of trained models and  well-trained models that are retained in each individual setting.

\begin{table}[h]
\caption{
Counts of trained models and retained well-trained models
in each individual setting.
For the last column, the denominator denotes the total number of trained models obtained from the hyperparameter sweep, and the numerator denotes the number of well-trained models that are retained for analysis ($\text{training loss}\!\le\!0.01$).
}
\centering
\small
\setlength{\tabcolsep}{18pt}
\renewcommand{\arraystretch}{1.0}

\begin{tabular}{llcc}
\toprule
\textbf{Dataset} 
& \textbf{Architecture} 
& \textbf{Optimizer} 
& \textbf{Retained / Total} \\
\midrule

\multirow{8}{*}{CIFAR-10}
& ResNet18        & SGD   & 30 / 48 \\
& VGG13           & SGD   & 34 / 48 \\
& ResNet18 no-BN  & SGD   & 58 / 60 \\
& VGG13 no-BN     & SGD   & 30 / 50 \\
& ResNet18        & AdamW & 46 / 48 \\
& VGG13           & AdamW & 39 / 48 \\
& ResNet18 no-BN  & AdamW & 35 / 48 \\
& VGG13 no-BN     & AdamW & 37 / 48 \\
\midrule

\multirow{4}{*}{CIFAR-100}
& WideResNet       & SGD   & 24 / 48 \\
& ViT              & AdamW & 47 / 48 \\
& WideResNet no-BN & SGD   & 47 / 48 \\
& ViT no-LN        & AdamW & 28 / 48 \\


\bottomrule
\end{tabular}
\label{tab:trained_model_counts}
\end{table}

\section{All numerical results}
\label{ap:numerical_results_full}

\subsection{Optimizer adamw} 
\label{subap:trained_by_adamw}
\vspace{-1.0em}
\begin{table}[H]
\caption{
\textbf{Main results} (optimizer=Adamw). 
Experimental settings, evaluated metric pairs and evaluation criteria are the same as in Table~\ref{tab:main_result_table}.
Here, models are trained using AdamW, rather than SGD.
}
\vspace{-8pt}

\centering
\footnotesize
\setlength{\tabcolsep}{5pt}
\setlength{\fboxsep}{0.2pt}
\renewcommand{\arraystretch}{1.2}
\resizebox{0.99\linewidth}{!}{%
\begin{tabular}{
    @{}l
    *{3}{
        >{\raggedleft\arraybackslash}p{1.05cm}
        >{\centering\arraybackslash}p{1.50cm}
        !{\SCbodysep}
    }
    >{\raggedleft\arraybackslash}p{1.00cm}
    >{\centering\arraybackslash}p{1.40cm}
    @{}
}
\toprule

\multicolumn{1}{@{}c}{%
    \multirow{3}{*}{%
        \makecell[c]{%
            $\boldsymbol{(S,C)}$\\
            \textbf{Metric Pair}%
        }%
    }%
}
& \multicolumn{8}{c}{\textbf{CIFAR-10}} \\

& \multicolumn{2}{c}{\textbf{ResNet18} \scriptsize{adamw}}
& \multicolumn{2}{c}{\textbf{VGG13} \scriptsize{adamw}}
& \multicolumn{2}{c}{\textbf{ResNet18 no-BN} \scriptsize{adamw}}
& \multicolumn{2}{c}{\textbf{VGG13 no-BN} \scriptsize{adamw}} \\[-3pt]

\cmidrule(lr){2-3}
\cmidrule(lr){4-5}
\cmidrule(lr){6-7}
\cmidrule(lr){8-9}

& \multicolumn{1}{r}{\textbf{PCR}}
& \multicolumn{1}{c}{$\boldsymbol{R^2}\,(S,C)$}
& \multicolumn{1}{r}{\textbf{PCR}}
& \multicolumn{1}{c}{$\boldsymbol{R^2}\,(S,C)$}
& \multicolumn{1}{r}{\textbf{PCR}}
& \multicolumn{1}{c}{$\boldsymbol{R^2}\,(S,C)$}
& \multicolumn{1}{r}{\textbf{PCR}}
& \multicolumn{1}{c}{$\boldsymbol{R^2}\,(S,C)$}
\\[-3pt]


\midrule

\SCtrFunc
& 44.1\% & \SCrsq{0.59}{(+,+)}
& 18.0\% & \SCrsq{0.83}{(+,+)}
& 32.5\% & \SCrsq{0.69}{(-,+)}
& 28.4\% & \SCrsq{0.15}{(+,+)}\\

\SClmaxFunc
& 46.1\% & \SCrsq{0.59}{(+,+)}
& 17.1\% & \SCrsq{0.83}{(+,+)}
& 27.5\% & \SCrsq{0.67}{(-,+)}
& 29.9\% & \SCrsq{0.15}{(+,+)}\\

\iteSCtrfsFunc
& 3.6\% & \SCrsq{0.70}{(+,+)}
& 6.7\% & \SCrsq{0.91}{(+,+)}
& 6.0\% & \SCrsq{0.75}{(+,+)}
& 6.0\% & \SCrsq{0.78}{(+,+)}\\

\iteSClmaxfsFunc
& 4.3\% & \SCrsq{0.64}{(+,+)}
& 5.5\% & \SCrsq{0.81}{(+,+)}
& 6.7\% & \SCrsq{0.73}{(+,+)}
& 10.8\% &\SCrsq{0.66}{(+,+)}\\

\addlinespace[2pt]

\SCtrLtwo
& 28.9\% & \SCrsq{0.78}{(+,+)}
& 8.8\% & \SCrsq{0.57}{(+,+)}
& 42.5\% & \SCrsq{0.51}{(-,+)}
& 17.0\% & \SCrsq{0.79}{(+,+)}\\

\SClmaxLtwo
& 30.1\% & \SCrsq{0.78}{(+,+)}
& 8.1\% & \SCrsq{0.57}{(+,+)}
& 36.7\% & \SCrsq{0.49}{(-,+)}
& 17.0\% & \SCrsq{0.79}{(+,+)}\\

\iteSCtrfsLtwo
& 10.1\% & \SCrsq{0.76}{(+,+)}
& 7.3\% & \SCrsq{0.91}{(+,+)}
& 10.5\% & \SCrsq{0.43}{(-,+)}
& 4.5\% & \SCrsq{0.85}{(+,+)}\\

\iteSClmaxfsLtwo
& 9.7\% & \SCrsq{0.78}{(+,+)}
& 7.1\% & \SCrsq{0.81}{(+,+)}
& 9.6\% & \SCrsq{0.51}{(-,+)}
& 5.9\% & \SCrsq{0.93}{(+,+)}\\

\addlinespace[2pt]

\SCtrPath 
& 56.2\% & \SCrsq{0.48}{(-,+)}
& 31.0\% & \SCrsq{0.70}{(-,+)}
& 43.9\% & \SCrsq{0.59}{(-,+)}
& 16.6\% & \SCrsq{0.73}{(+,+)}\\

\SClmaxPath
& 56.7\% & \SCrsq{0.48}{(-,+)}
& 29.5\% & \SCrsq{0.70}{(-,+)}
& 37.9\% & \SCrsq{0.56}{(-,+)}
& 16.5\% & \SCrsq{0.73}{(+,+)}\\

\iteSCtrfsPath
& 29.3\% & \SCrsq{0.60}{(-,+)}
& 19.6\% & \SCrsq{0.89}{(+,-)}
& 10.0\% & \SCrsq{0.55}{(+,+)}
& 4.6\% & \SCrsq{0.84}{(+,+)}\\

\iteSClmaxfsPath
& 28.3\% & \SCrsq{0.48}{(-,+)}
& 17.9\% & \SCrsq{0.89}{(+,+)}
& 9.5\% & \SCrsq{0.56}{(-,+)}
& 6.1\% & \SCrsq{0.91}{(+,+)}\\

\addlinespace[2pt]

\SCtrSpec
& 42.2\% & \SCrsq{0.64}{(+,+)}
& 17.8\% & \SCrsq{0.66}{(+,+)}
& 44.9\% & \SCrsq{0.56}{(-,+)}
& 16.1\% & \SCrsq{0.66}{(+,+)}\\

\SClmaxSpec
& 44.1\% & \SCrsq{0.64}{(+,+)}
& 15.9\% & \SCrsq{0.66}{(+,+)}
& 38.3\% & \SCrsq{0.52}{(-,+)}
& 16.3\% & \SCrsq{0.66}{(+,+)}\\

\iteSCtrfsSpec
& 12.7\% & \SCrsq{0.69}{(+,+)}
& 8.9\% & \SCrsq{0.94}{(+,+)}
& 9.7\% & \SCrsq{0.49}{(+,+)}
& 4.7\% & \SCrsq{0.83}{(+,+)}\\

\iteSClmaxfsSpec
& 12.6\% & \SCrsq{0.69}{(+,+)}
& 8.7\% & \SCrsq{0.87}{(+,+)}
& 9.4\% & \SCrsq{0.49}{(-,+)}
& 6.0\% & \SCrsq{0.89}{(+,+)}\\

\bottomrule
\end{tabular}%

}
\label{tab:trained_by_adamw}
\end{table}


\subsection{Mixed settings} 
\label{subap:mixed_setting}
\vspace{-1.0em}


\begin{table}[H]
\caption{
A detailed version of Table~\ref{tab:main_result_table_mixedsetting}.
The last row lists models being pooled in each mixed setting, with the well-trained model counts in parentheses.
}
\vspace{-8pt}
\centering
\footnotesize
\setlength{\tabcolsep}{5pt}
\setlength{\fboxsep}{0.2pt}
\renewcommand{\arraystretch}{1.15}

\resizebox{0.99\linewidth}{!}{%
\begin{tabular}{
    @{}l
    *{3}{
        >{\raggedleft\arraybackslash}p{1.05cm}
        >{\centering\arraybackslash}p{1.50cm}
        !{\SCbodysep}
    }
    >{\raggedleft\arraybackslash}p{1.00cm}
    >{\centering\arraybackslash}p{1.40cm}
    @{}
}
\toprule

\multicolumn{1}{@{}c}{%
    \multirow{2}{*}{%
        \makecell[c]{%
            $\boldsymbol{(S,C)}$\\
            \textbf{Metric \textit{Pair}}%
        }%
    }%
}
& \multicolumn{2}{c}{%
    \makecell[c]{%
        \multirow{2}{*}{\textbf{Cross-architecture}}\\
        {}\\
        {\scriptsize mixed four}
    }%
}
& \multicolumn{2}{c}{%
    \makecell[c]{%
        \textbf{Cross-optimizer}\\
        \textbf{\;\;\& Cross-architecture}\\
        {\scriptsize mixed eight}
    }%
}
& \multicolumn{4}{c}{%
    \makecell[c]{%
        \multirow{2}{*}{\textbf{Cross-dataset \& Cross-architecture}}\\
        {}\\
        {\scriptsize mixed three\qquad\qquad\qquad\qquad mixed six}
    }%
}   \\[-1pt]

\cmidrule(lr){2-3}
\cmidrule(lr){4-5}
\cmidrule(lr){6-9}

& \multicolumn{1}{r}{\textbf{PCR}}
& \multicolumn{1}{c}{$\boldsymbol{R^2}\,(S,C)$}
& \multicolumn{1}{r}{\textbf{PCR}}
& \multicolumn{1}{c}{$\boldsymbol{R^2}\,(S,C)$}
& \multicolumn{1}{r}{\textbf{PCR}}
& \multicolumn{1}{c}{$\boldsymbol{R^2}\,(S,C)$}
& \multicolumn{1}{r}{\textbf{PCR}}
& \multicolumn{1}{c}{$\boldsymbol{R^2}\,(S,C)$}
\\[-3pt]


\midrule

\SCtrFunc 
& 4.1\% & \SCrsq{0.81}{(+,+)} 
& 11.5\%& \SCrsq{0.26}{(-,+)} 
& 2.3\% & \SCrsq{0.83}{(+,+)} 
& 2.2\% & \SCrsq{0.86}{(+,+)} \\
\SClmaxFunc     
& 4.1\% & \SCrsq{0.81}{(+,+)}  
& 11.6\%& \SCrsq{0.25}{(-,+)} 
& 2.2\% & \SCrsq{0.81}{(+,+)} 
& 2.9\% & \SCrsq{0.86}{(+,+)} \\
\SCtrfsFunc    
& 2.1\% & \SCrsq{0.88}{(+,+)}  
& 6.6\% & \SCrsq{0.32}{(+,+)}
& 3.0\% & \SCrsq{0.92}{(+,+)}  
& 2.7\% & \SCrsq{0.87}{(+,+)} \\
\SClmaxfsFunc    
& 1.7\% & \SCrsq{0.89}{(+,+)}  
& 6.1\% & \SCrsq{0.25}{(+,+)}
& 2.4\% & \SCrsq{0.92}{(+,+)}  
& 2.4\% & \SCrsq{0.87}{(+,+)} \\

\addlinespace[2pt]

\SCtrLtwo       
& 65.6\% &\SCrsq{0.18}{(-,+)}  
& 64.6\%& \SCrsq{0.01}{(-,-)} 
& 8.5\% & \SCrsq{0.81}{(+,+)} 
& 41.0\% & \SCrsq{0.05}{(+,+)} \\
\SClmaxLtwo   
& 62.3\% &\SCrsq{0.17}{(-,+)}  
& 63.5\%& \SCrsq{0.01}{(-,-)} 
& 9.8\% & \SCrsq{0.81}{(+,+)}  
& 39.8\% & \SCrsq{0.03}{(+,+)} \\
\SCtrfsLtwo      
& 29.9\% &\SCrsq{0.17}{(+,+)}  
& 27.0\%& \SCrsq{0.14}{(+,-)} 
& 9.2\% & \SCrsq{0.89}{(+,+)}  
& 30.7\% &\SCrsq{0.04}{(+,-)} \\
\SClmaxfsLtwo   
& 27.7\% & \SCrsq{0.19}{(+,+)}  
& 26.3\% & \SCrsq{0.05}{(+,-)} 
& 8.4\% & \SCrsq{0.88}{(+,+)} 
& 27.8\% & \SCrsq{0.11}{(+,-)} \\

\addlinespace[2pt]

\SCtrPath      
& 70.1\% & \SCrsq{0.11}{(-,+)}   
& 49.4\%& \SCrsq{0.61}{(-,+)} 
& 28.2\% & \SCrsq{0.35}{(-,+)} 
& 51.9\% & \SCrsq{0.04}{(+,-)}\\
\SClmaxPath      
& 68.5\% & \SCrsq{0.10}{(-,+)}  
& 49.1\%& \SCrsq{0.61}{(-,+)} 
& 30.8\% & \SCrsq{0.35}{(-,+)} 
& 51.6\% & \SCrsq{0.02}{(+,-)}\\
\SCtrfsPath     
& 36.8\% & \SCrsq{0.10}{(+,+)}  
& 25.8\%& \SCrsq{0.61}{(+,+)} 
& 17.1\% & \SCrsq{0.75}{(+,+)} 
& 36.1\% & \SCrsq{0.05}{(+,-)}\\
\SClmaxfsPath  
& 36.8\% & \SCrsq{0.11}{(+,+)}  
& 25.2\%& \SCrsq{0.61}{(+,+)} 
& 17.9\% & \SCrsq{0.67}{(+,+)}  
& 34.2\% & \SCrsq{0.12}{(+,-)} \\

\addlinespace[2pt]

\SCtrSpec        
& 66.8\% & \SCrsq{0.11}{(-,+)}  
& 47.5\%& \SCrsq{0.26}{(-,+)} 
& 43.4\% & \SCrsq{0.11}{(-,+)} 
& 53.3\% & \SCrsq{0.04}{(+,+)}  \\
\SClmaxSpec       
& 64.5\% & \SCrsq{0.11}{(-,+)}  
& 47.2\%& \SCrsq{0.26}{(-,+)} 
& 47.5\% & \SCrsq{0.11}{(-,+)} 
& 53.0\% & \SCrsq{0.02}{(+,+)} \\
\SCtrfsSpec       
& 28.2\% & \SCrsq{0.12}{(+,+)}  
& 19.6\%& \SCrsq{0.35}{(+,+)} 
& 14.8\% & \SCrsq{0.70}{(+,+)} 
& 32.7\% & \SCrsq{0.04}{(+,+)}  \\
\SClmaxfsSpec     
& 29.7\% & \SCrsq{0.14}{(+,+)}  
& 19.3\%& \SCrsq{0.30}{(+,+)}  
& 20.1\% & \SCrsq{0.61}{(+,+)} 
& 31.4\% & \SCrsq{0.11}{(+,+)}  \\[-2pt]

\midrule

\makecell[tc]{%
    \textbf{Models being}\\
    \textbf{pooled together}\\
    {(the number of}\\ 
    {well-trained models}\\
    {in the collection)}%
}
& \multicolumn{2}{c!{\SCfootersep}}{%
    \scriptsize \makecell[tl]{%
    CIFAR-10, {sgd}\\
        \quad ResNet18 \tiny{$\;(30)$}\\
        \quad VGG13 \tiny{$\;(34)$} \\
        \quad ResNet18 no-BN \tiny{$\;(58)$} \\
        \quad VGG13 no-BN \tiny{$\;(30)$}%
    }%
}
& \multicolumn{2}{c!{\SCfootersep}}{%
    \scriptsize \makecell[tl]{%
    CIFAR-10, {sgd \& adamw}\\
        \quad ResNet18 \tiny{$\;(76)$}\\
        \quad VGG13 \tiny{$\;(73)$}\\
        \quad ResNet18 no-BN \tiny{$\;(93)$}\\
        \quad VGG13 no-BN \tiny{$\;(67)$}%
    }%
}
& \multicolumn{2}{c!{\SCfootersep}}{%
    \scriptsize \makecell[tl]{%
    CIFAR-10, {sgd}\\
        \quad ResNet18 \tiny{$\;(30)$}\\
        \quad VGG13 \tiny{$\;(34)$}\\
    CIFAR-100, {sgd}\\
        \quad WideResNet \tiny{$\;(24)$}%
    }%
}
& \multicolumn{2}{c}{%
    \scriptsize \makecell[tl]{%
    CIFAR-10, {sgd}\\
        \quad ResNet18 \tiny{$\;(30)$}\\
        \quad VGG13 \tiny{$\;(34)$}\\
        \quad ResNet18 no-BN \tiny{$\;(58)$}\\
        \quad VGG13 no-BN \tiny{$\;(30)$}\\
    CIFAR-100, {sgd}\\
        \quad WideResNet \tiny{$\;(24)$}\\
        \quad WideResNet no-BN \tiny{$(47)$}\\%
    }%
} \\[-3pt]

\makecell[c]{%
    \textbf{Collection size}
}
& \multicolumn{2}{c!{\SCbodysep}}{$152$}
& \multicolumn{2}{c!{\SCbodysep}}{$309$}
& \multicolumn{2}{c!{\SCbodysep}}{$88$}
& \multicolumn{2}{c}{$223$}
\\[-2pt]

\bottomrule
\end{tabular}%
}
\label{tab:main_result_table_mixedsetting_full}
\end{table}

\subsection{Varying sharpness perturbation scales} 
\label{subap:vary_scale_fullfigure}

\begin{figure}[H]
    \centering
    
    \begin{subfigure}[b]{0.249\textwidth}
        \centering
        \includegraphics[width=\linewidth]{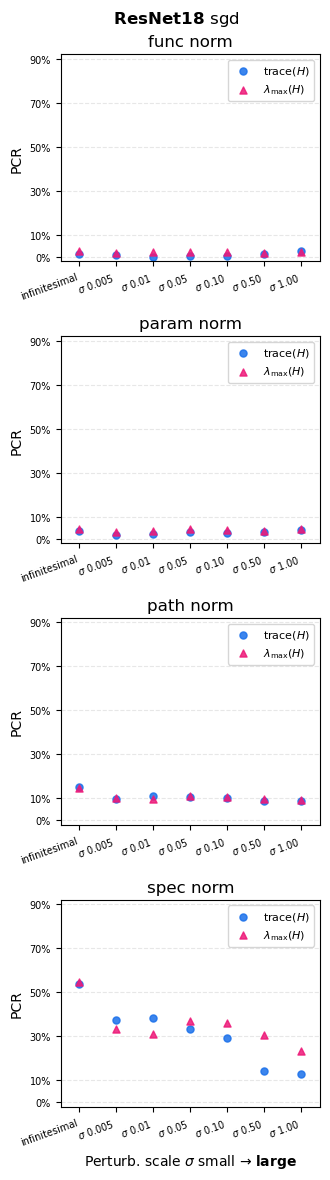}
        \label{fig:panel_a}
    \end{subfigure}\hfill
    \begin{subfigure}[b]{0.249\textwidth}
        \centering
        \includegraphics[width=\linewidth]{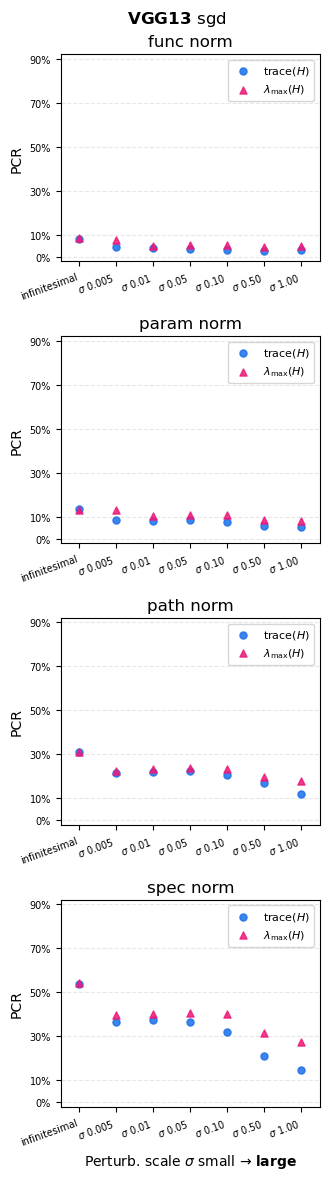}
        \label{fig:panel_b}
    \end{subfigure}\hfill
    \begin{subfigure}[b]{0.249\textwidth}
        \centering
        \includegraphics[width=\linewidth]{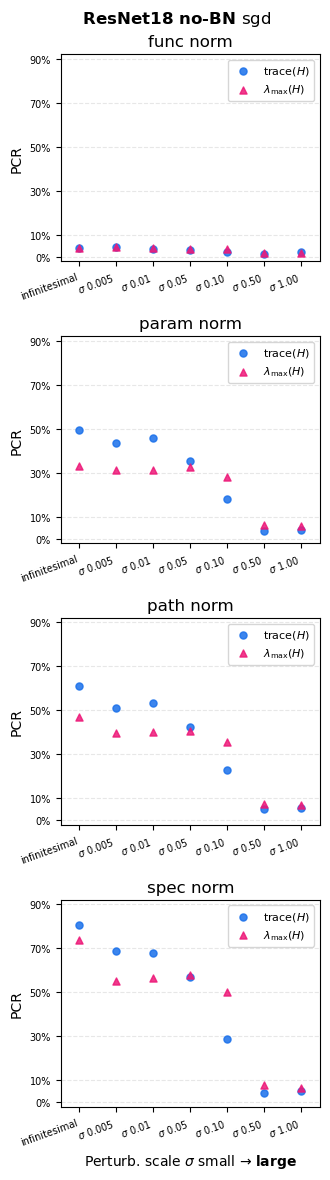}
        \label{fig:panel_c}
    \end{subfigure}\hfill
    \begin{subfigure}[b]{0.249\textwidth}
        \centering
        \includegraphics[width=\linewidth]{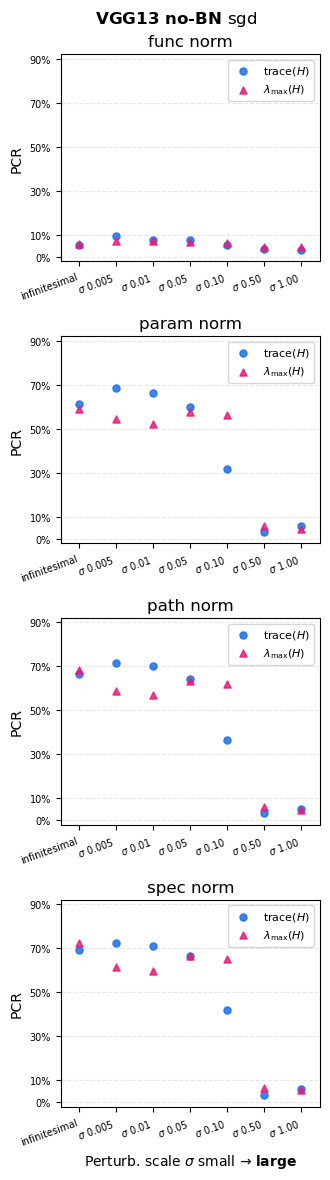}
        \label{fig:panel_d}
    \end{subfigure}
\caption{ Full version of Figure~\ref{fig:varying_perturb_scale}.
How PCR varies as the sharpness perturbation radius increases from an infinitesimal scale to small and then large finite scale.
Each column corresponds to an architecture. 
Within each subplot, with the complexity metric fixed, sharpness defined over a \emph{larger} neighborhood often yields lower PCR, indicating stronger explanatory power and supporting Observation~\ref{obs:larger_scale_much_better}.
The row-wise comparisons of subplots support Observations~\ref{obs:func_norm_more_stable}, \ref{obs:param_norm_still_good}, and~\ref{obs:path_spec_norm_bad}.
}
\label{fig:varying_perturb_scale_full}
\end{figure}



\subsection{Visualized PCRs} 
\label{subap:Visualized_PCRs}

\begin{figure}[H]
    \centering
    \begin{minipage}{0.99\linewidth} 
        \centering

        \begin{subfigure}[b]{0.45\linewidth}
            \centering
            \includegraphics[width=\linewidth]{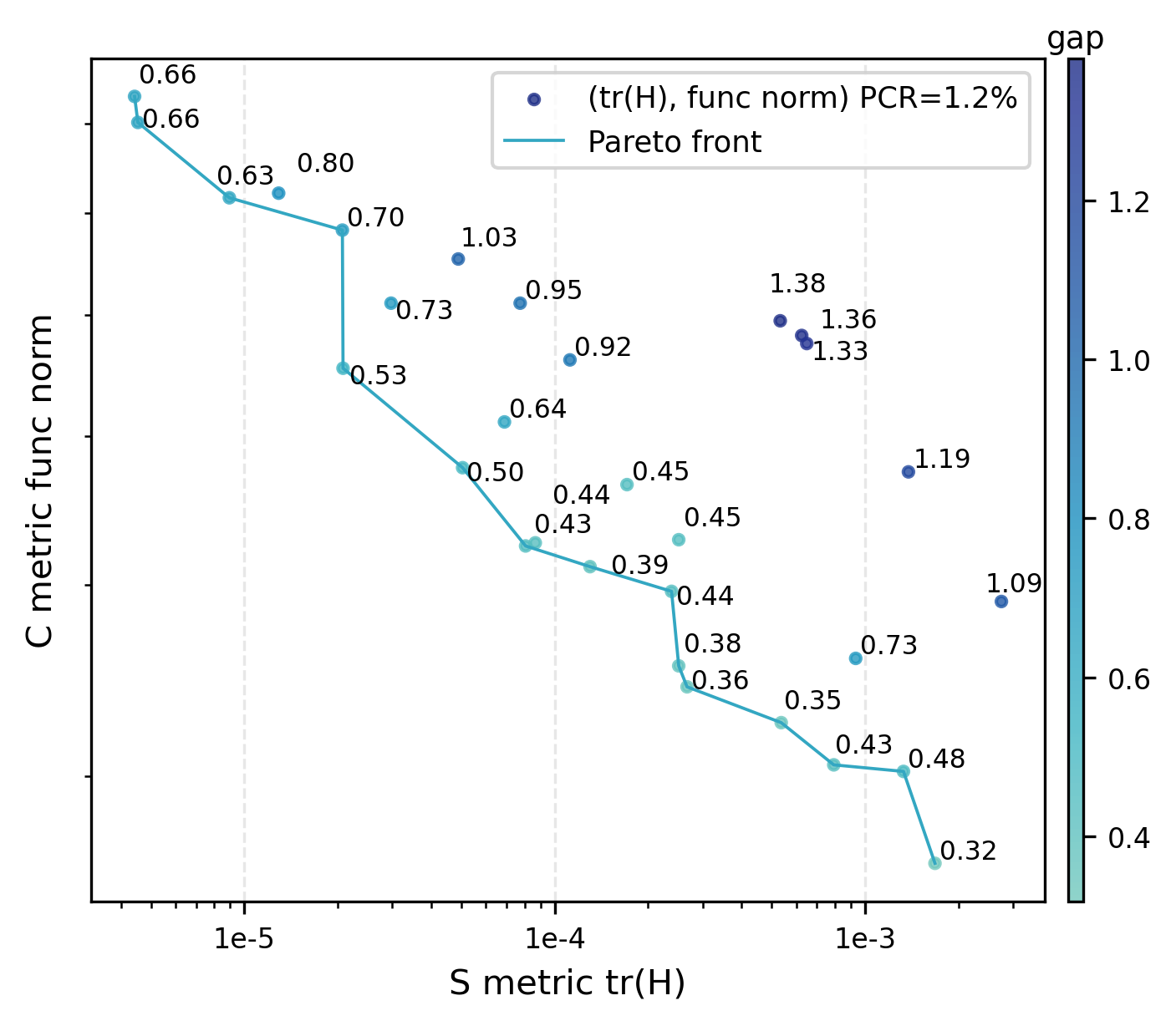}
            \caption{(tr(H), func norm), PCR$\!=\!1.2\%$.}
        \end{subfigure}\hfill
        \begin{subfigure}[b]{0.45\linewidth}
            \centering
            \includegraphics[width=\linewidth]{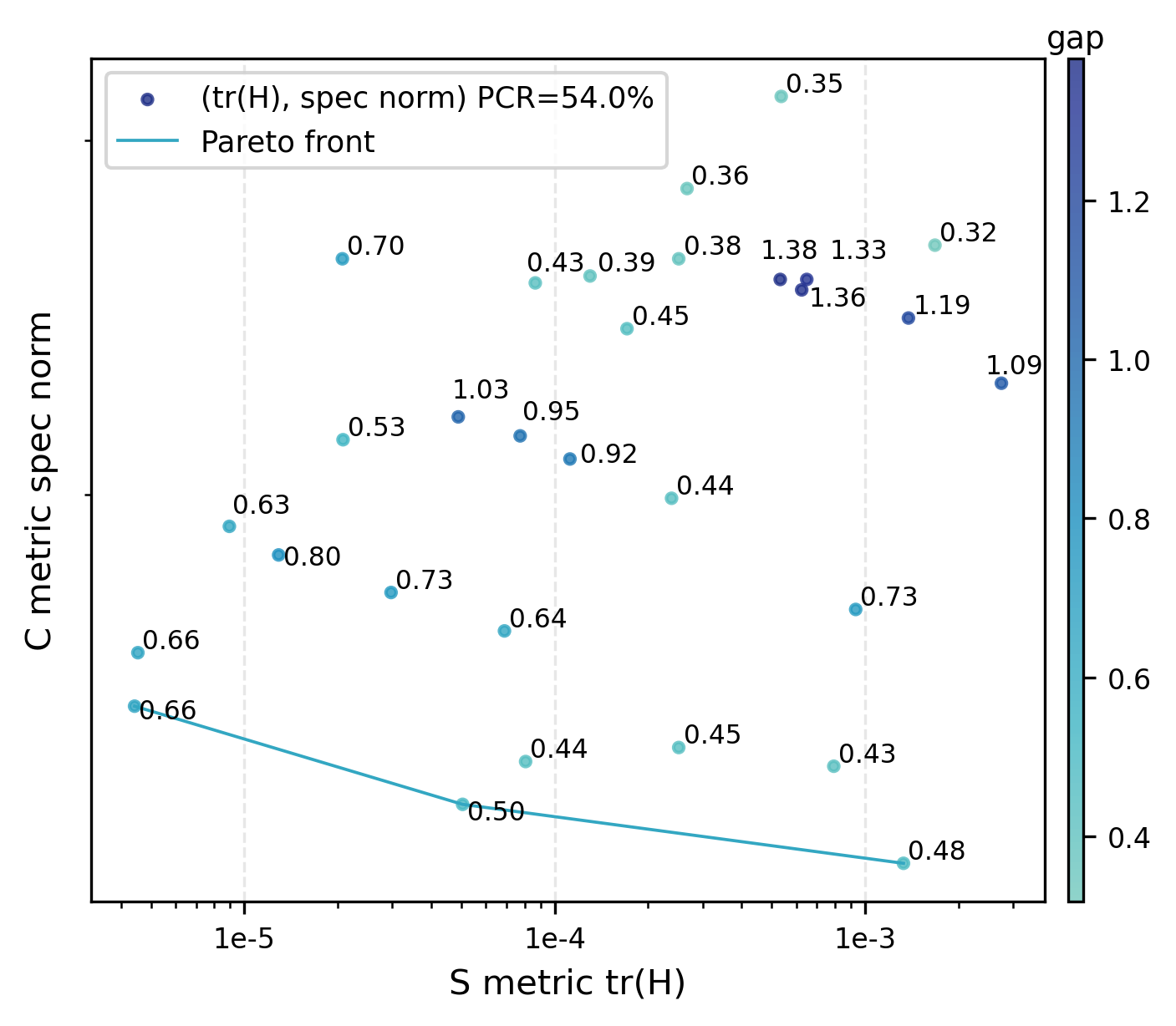}
            \caption{(tr(H), spec norm), PCR$\!=\!54.0\%$.}
        \end{subfigure}

        \caption{Two representative PCR visualizations in the sgd-trained ResNet18 setting, corresponding to the two PCR values listed in the first vertical block in Table~\ref{tab:main_result_table}.
        Lighter colors indicate smaller generalization gaps.
        Under the expected monotonic relationship, the lower-left point in each candidate pair should have a smaller generalization gap and therefore a lighter color, than the upper-right point.
        The left, metric pair (tr(H), func norm): few violations of the expected ordering, achieving low PCR 1.2\%.
        The right, metric pair (tr(H), spec norm): severe violations of the expected ordering, achieving high PCR 54.0\%.
        }
        \label{fig:selected_PCR_visual_resnet18}
    \end{minipage}
\end{figure}

\begin{figure}[H]

    \centering

    \includegraphics[width=0.248\linewidth]{figs/resnet18bn_trH__func_norm.png}\hfill
    \includegraphics[width=0.248\linewidth]{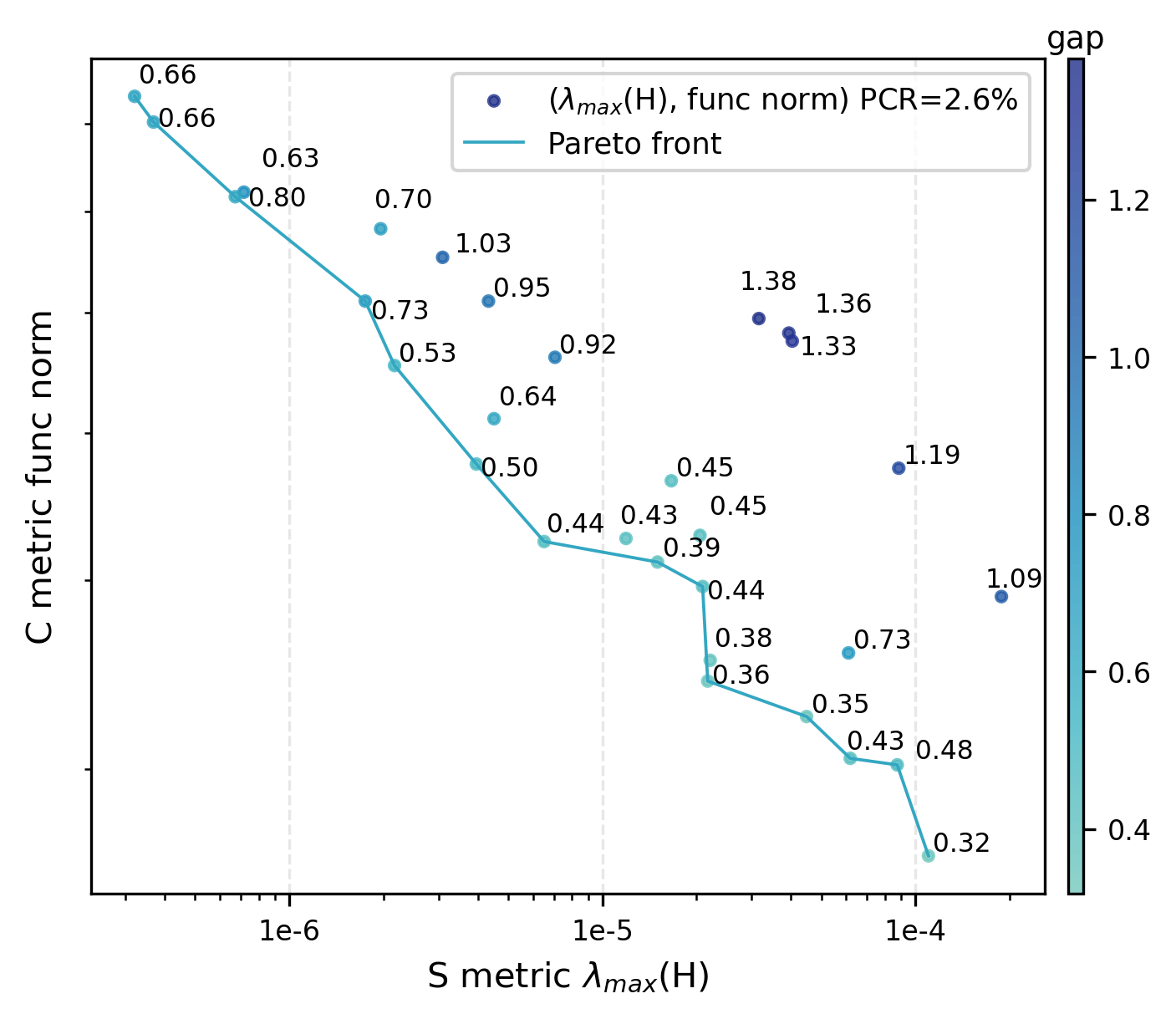}\hfill
    \includegraphics[width=0.248\linewidth]{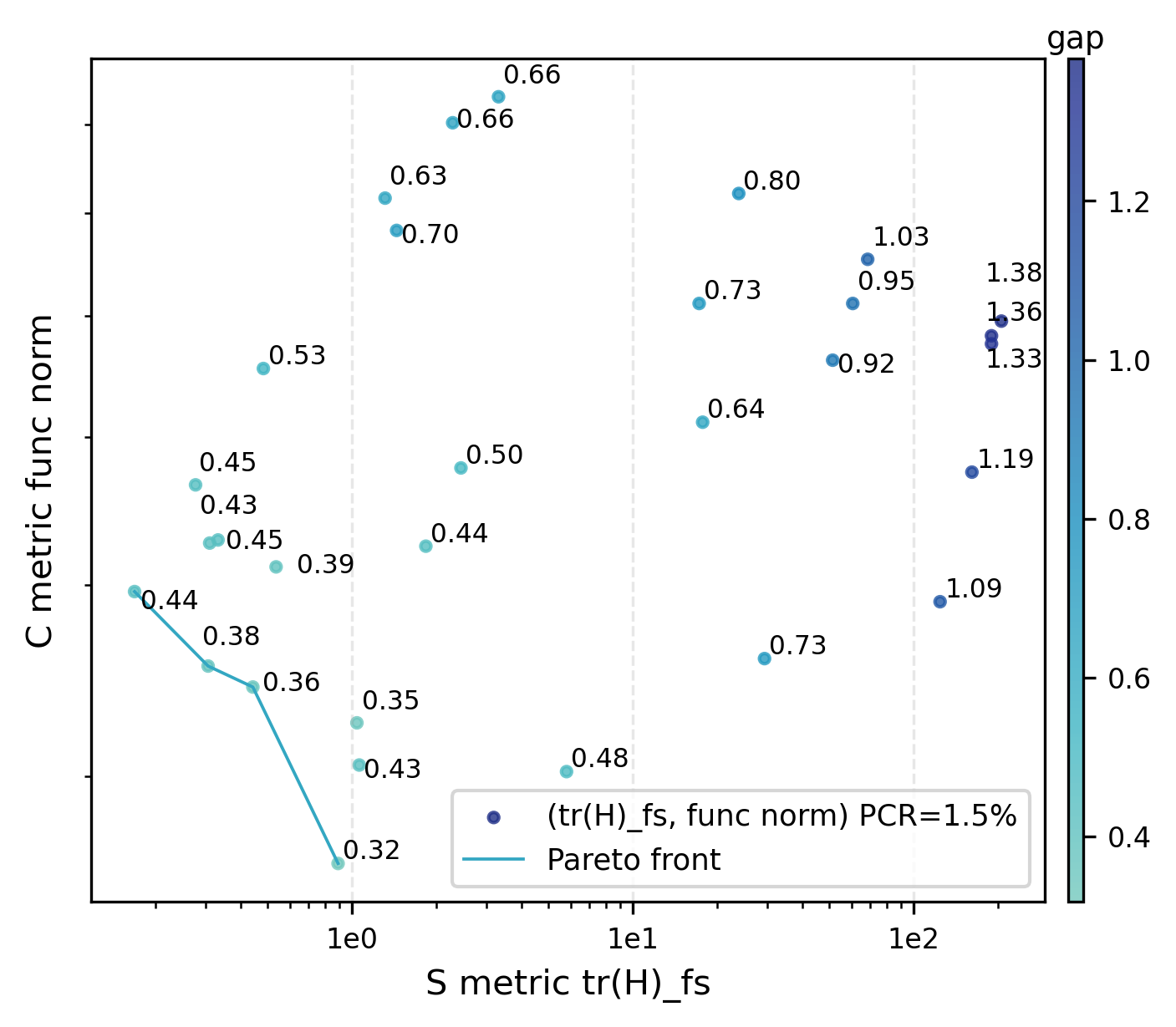}\hfill
    \includegraphics[width=0.248\linewidth]{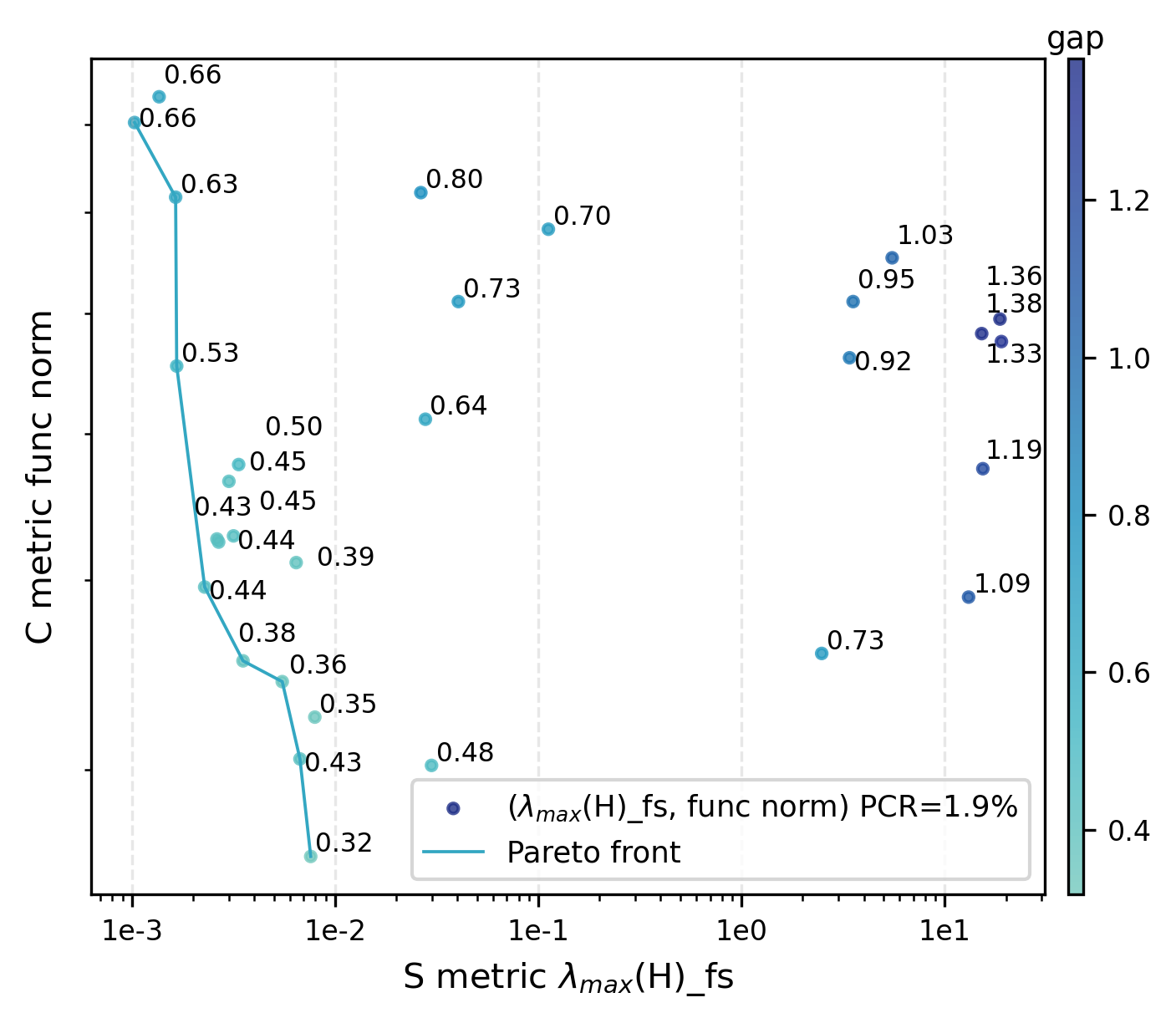}\hfill
    \par\smallskip
    {\scriptsize Func norm paired with four sharpness metrics.\par}
    \medskip

    \includegraphics[width=0.248\linewidth]{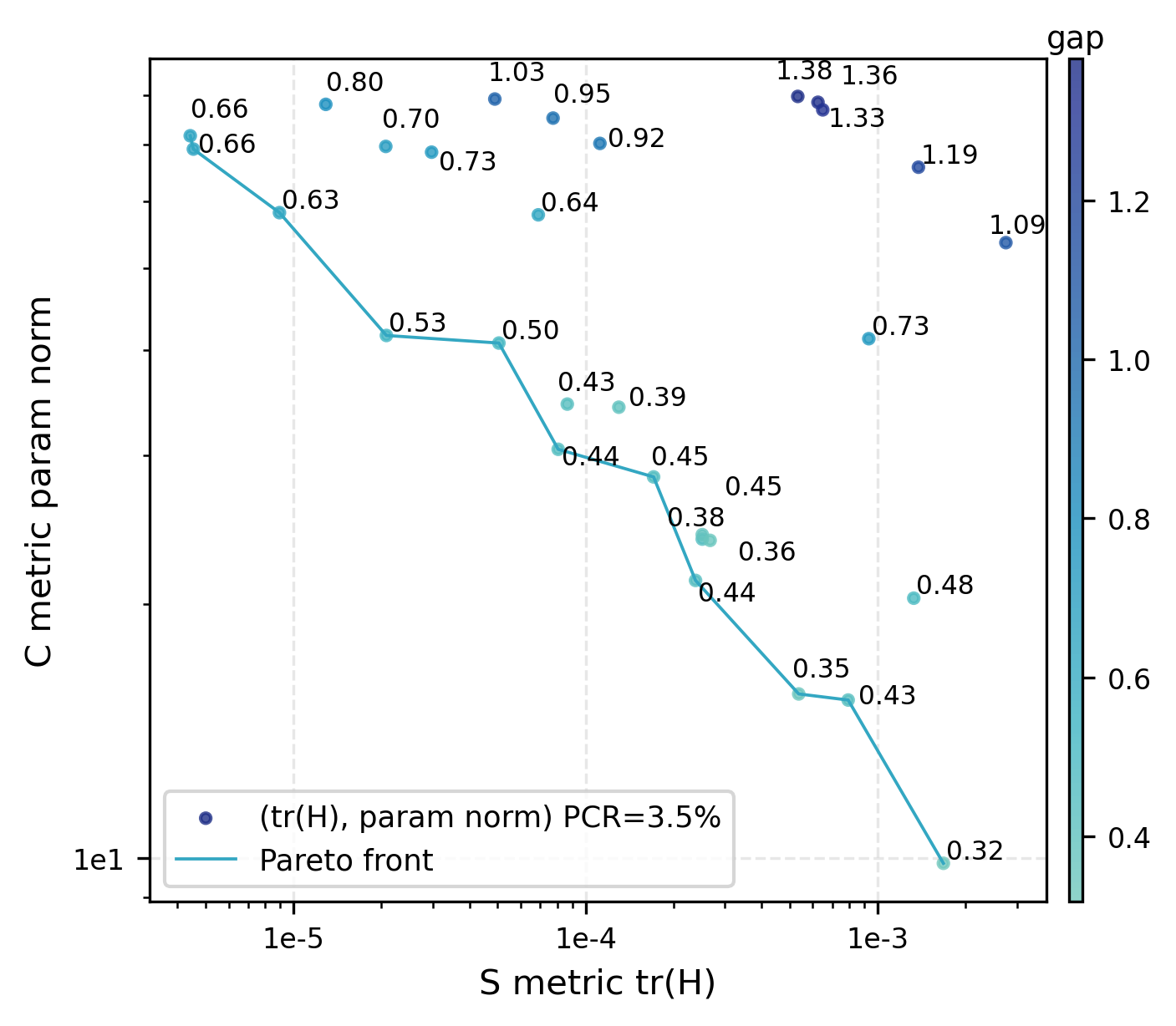}\hfill
    \includegraphics[width=0.248\linewidth]{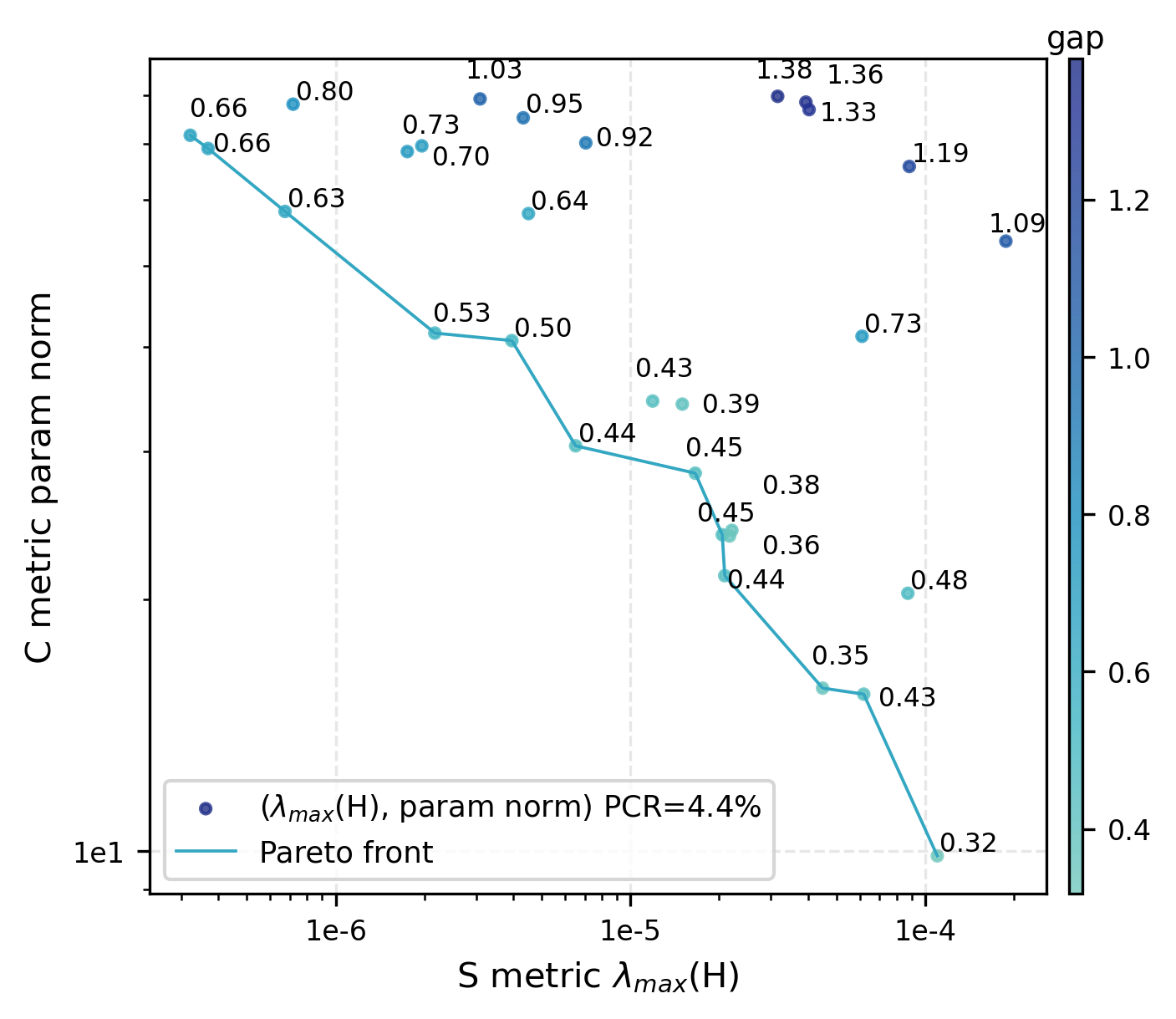}\hfill
    \includegraphics[width=0.248\linewidth]{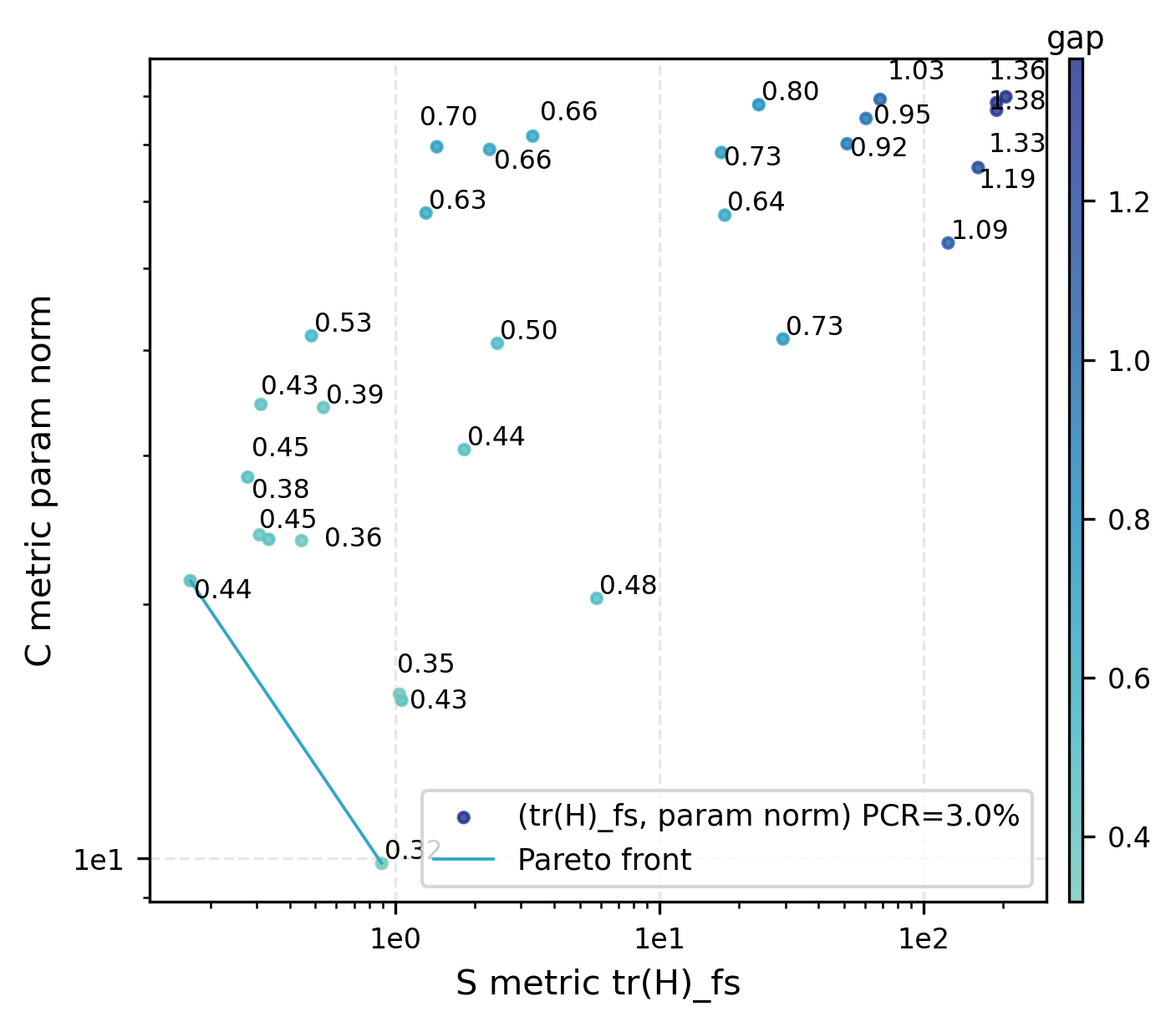}\hfill
    \includegraphics[width=0.248\linewidth]{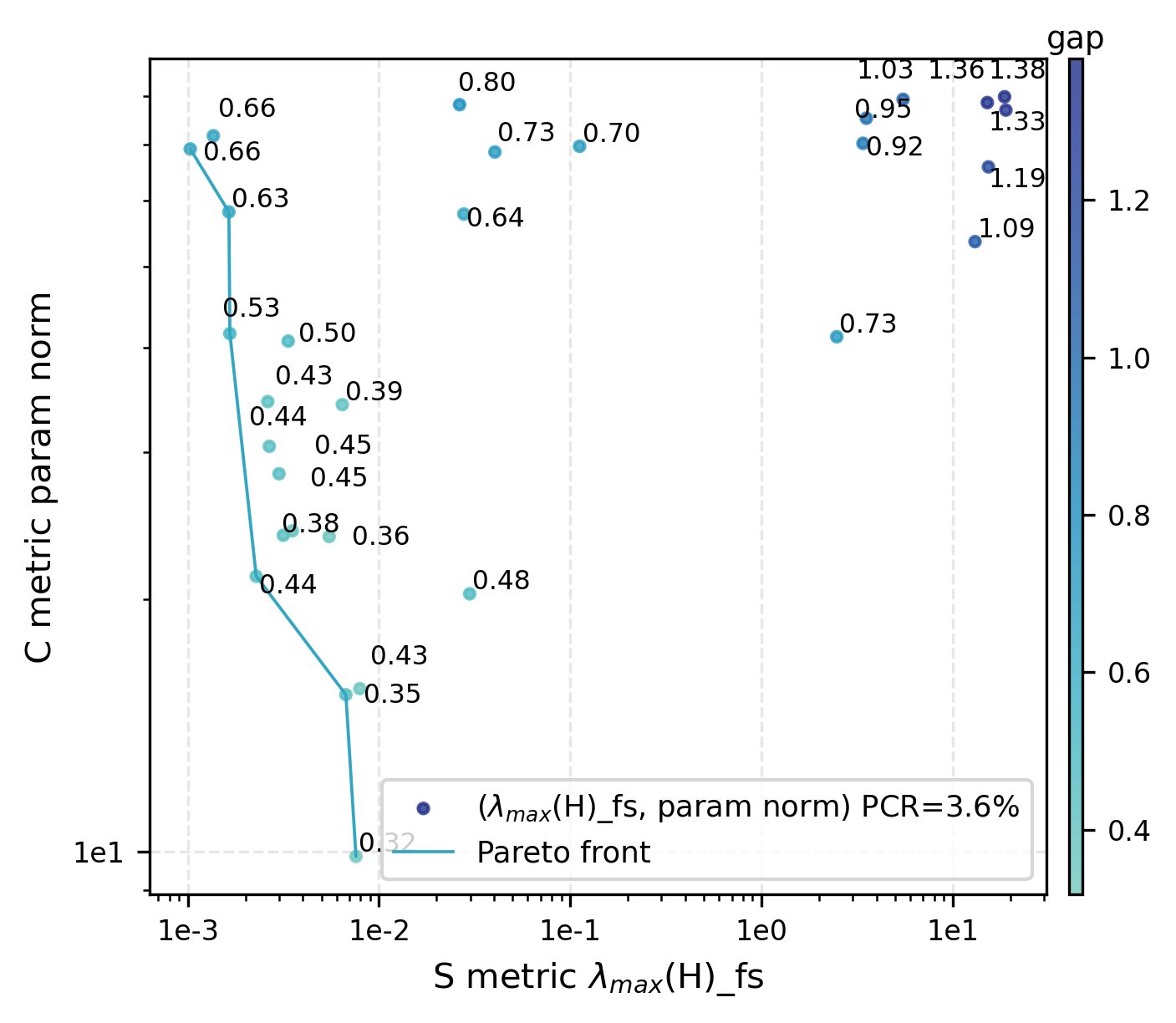}\hfill
    \par\smallskip
    {\scriptsize Param norm paired with four sharpness metrics.\par}
    \medskip

    \includegraphics[width=0.248\linewidth]{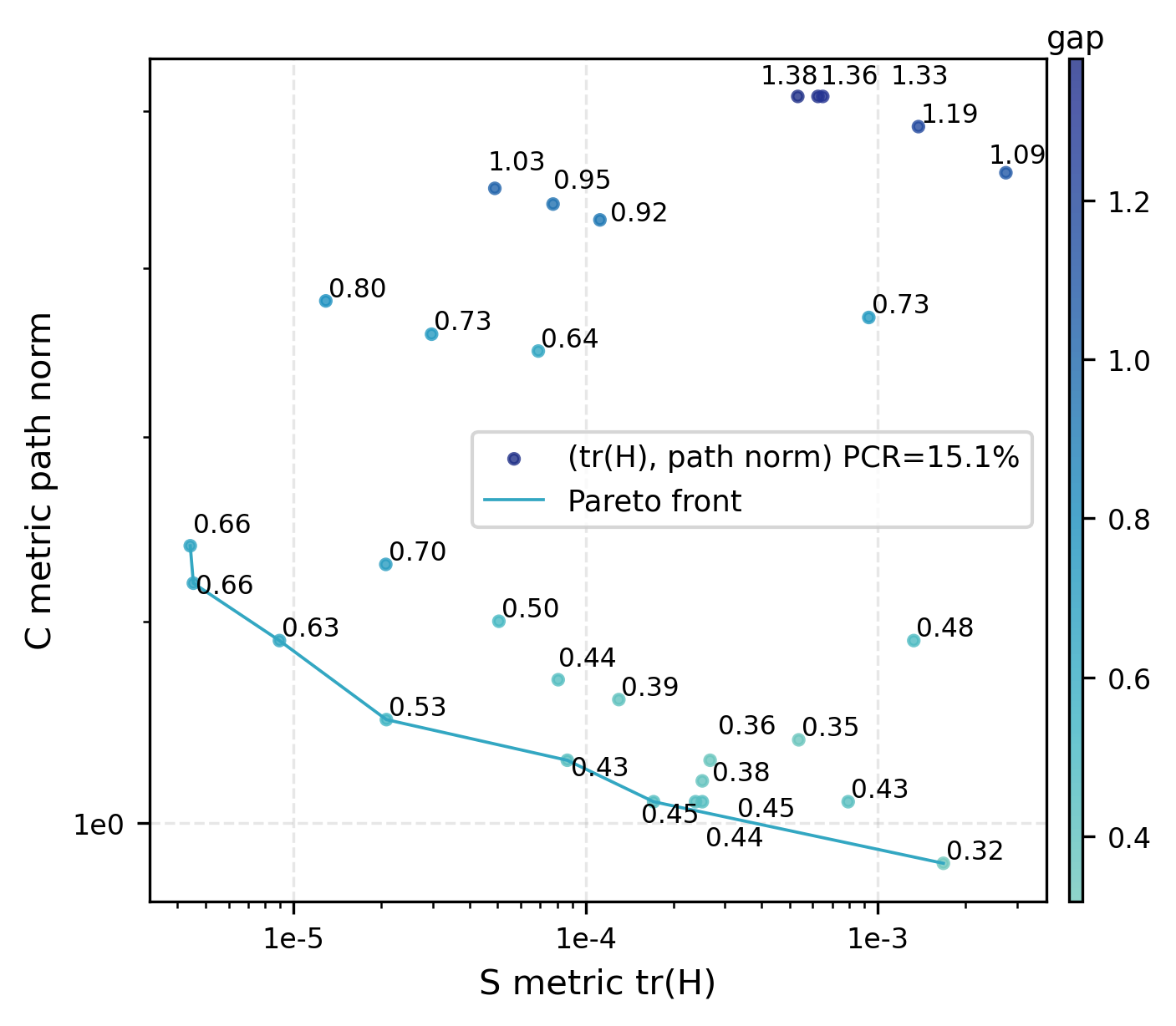}\hfill
    \includegraphics[width=0.248\linewidth]{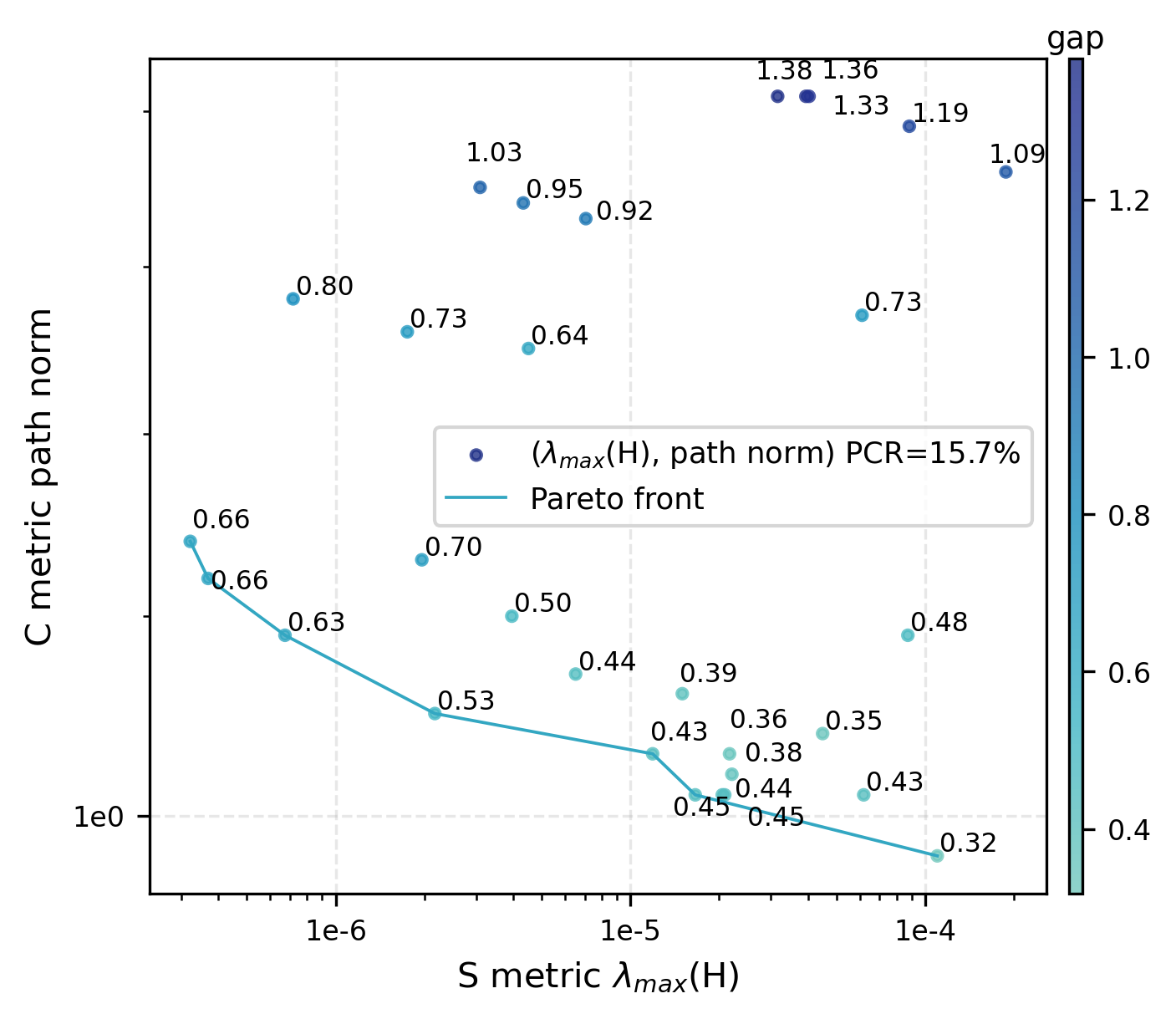}\hfill
    \includegraphics[width=0.248\linewidth]{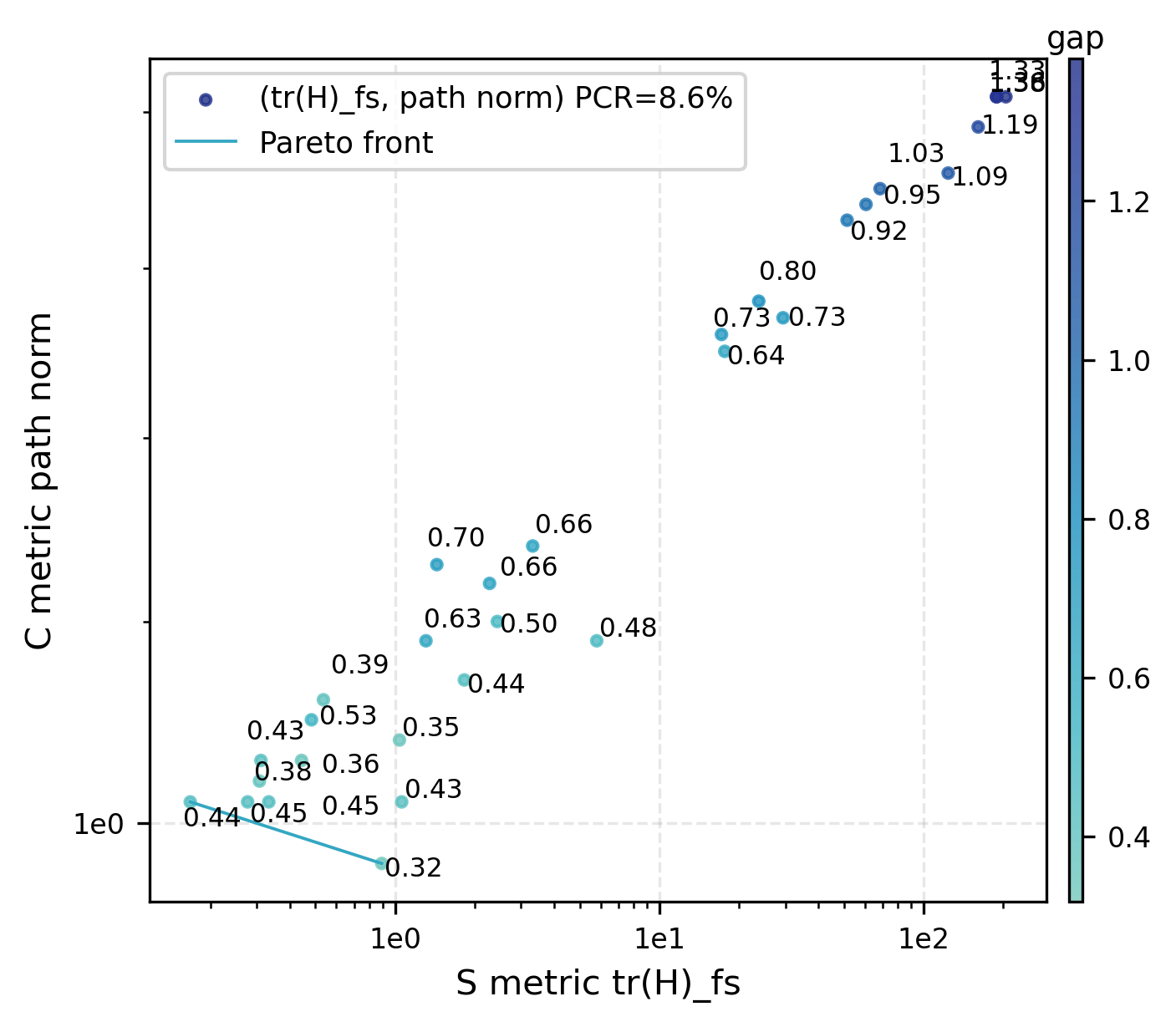}\hfill
    \includegraphics[width=0.248\linewidth]{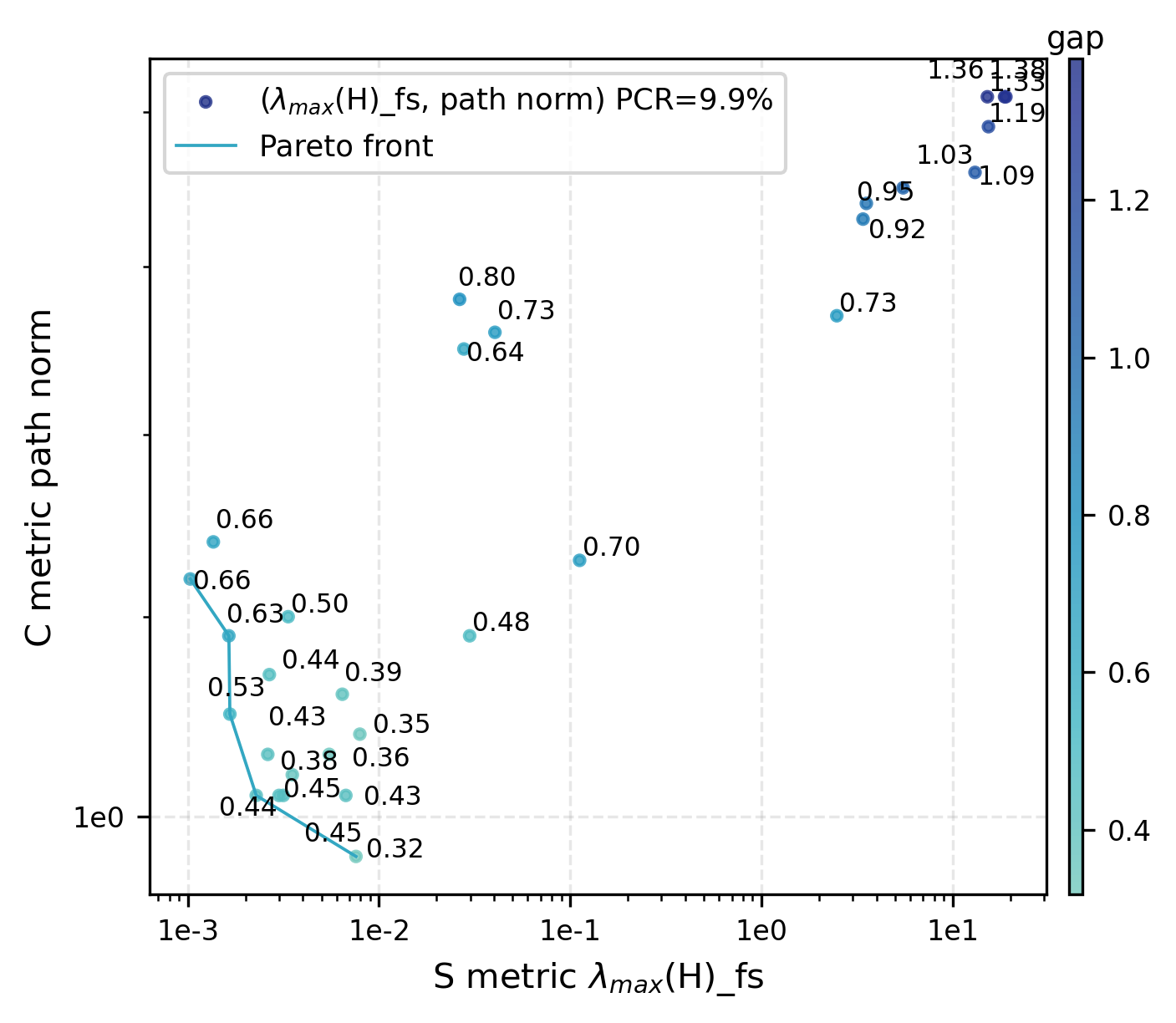}\hfill
    \par\smallskip
    {\scriptsize Path norm paired with four sharpness metrics.\par}
    \medskip

    \includegraphics[width=0.248\linewidth]{figs/resnet18bn_trH__spec_norm.png}\hfill
    \includegraphics[width=0.248\linewidth]{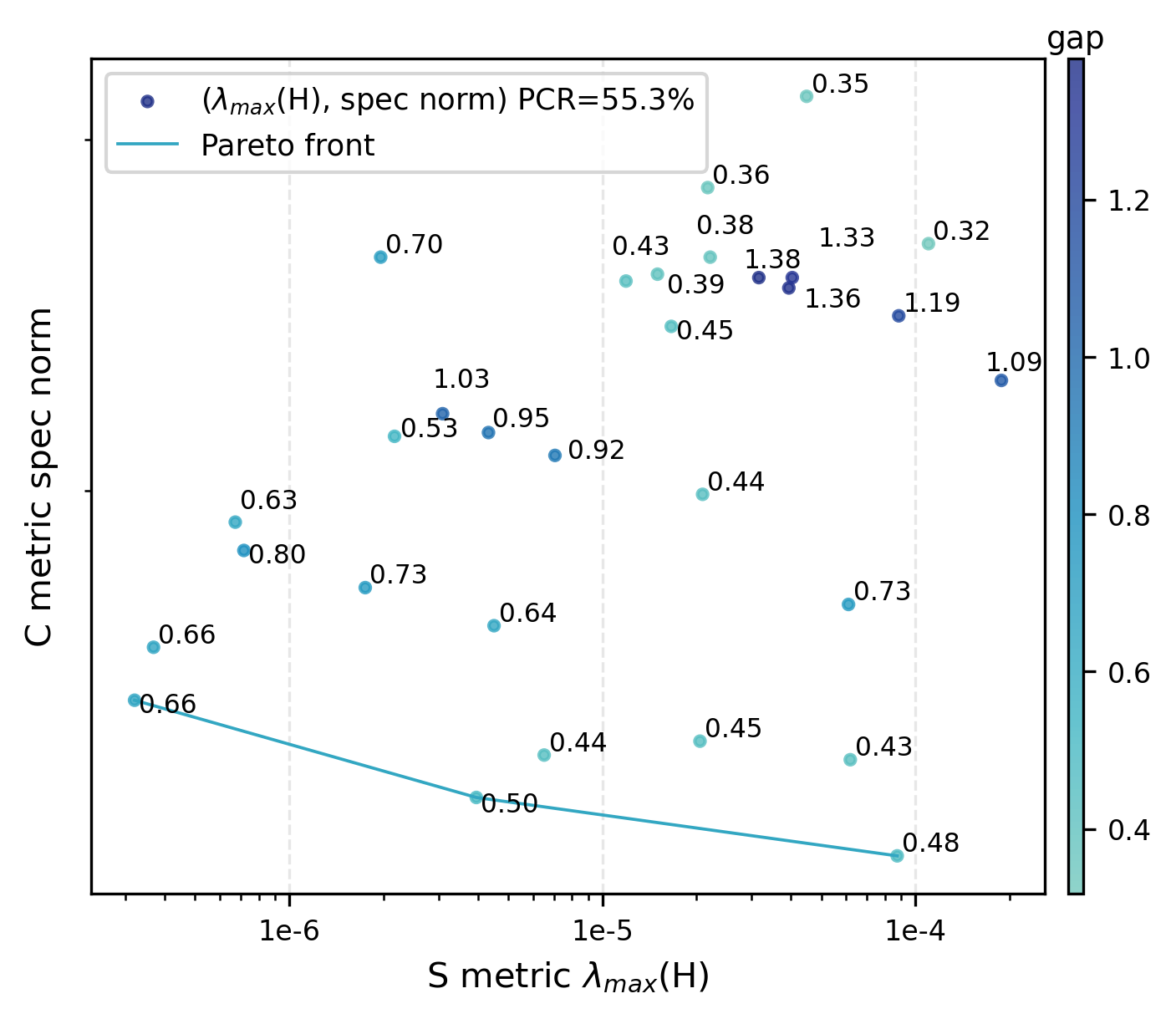}\hfill
    \includegraphics[width=0.248\linewidth]{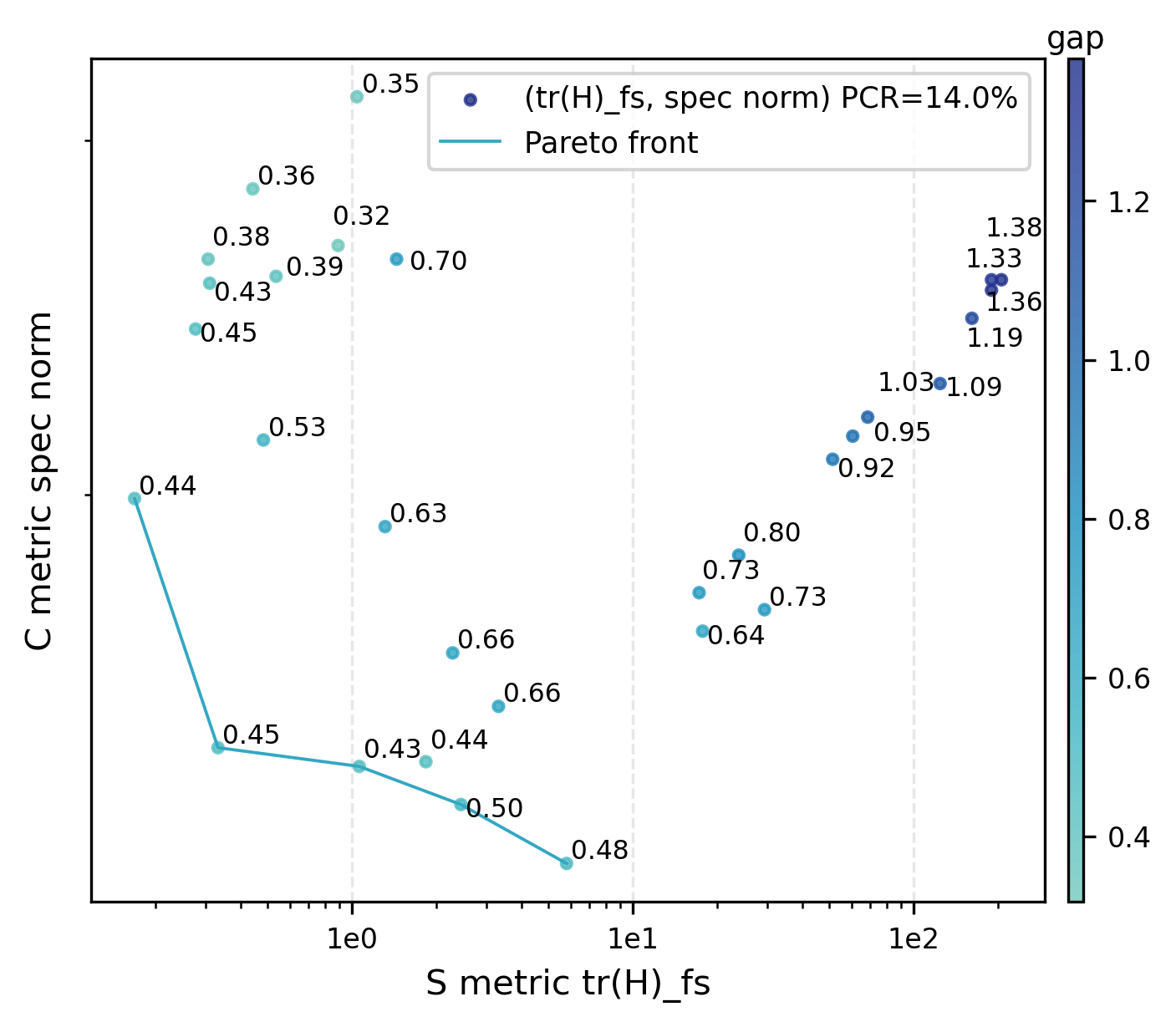}\hfill
    \includegraphics[width=0.248\linewidth]{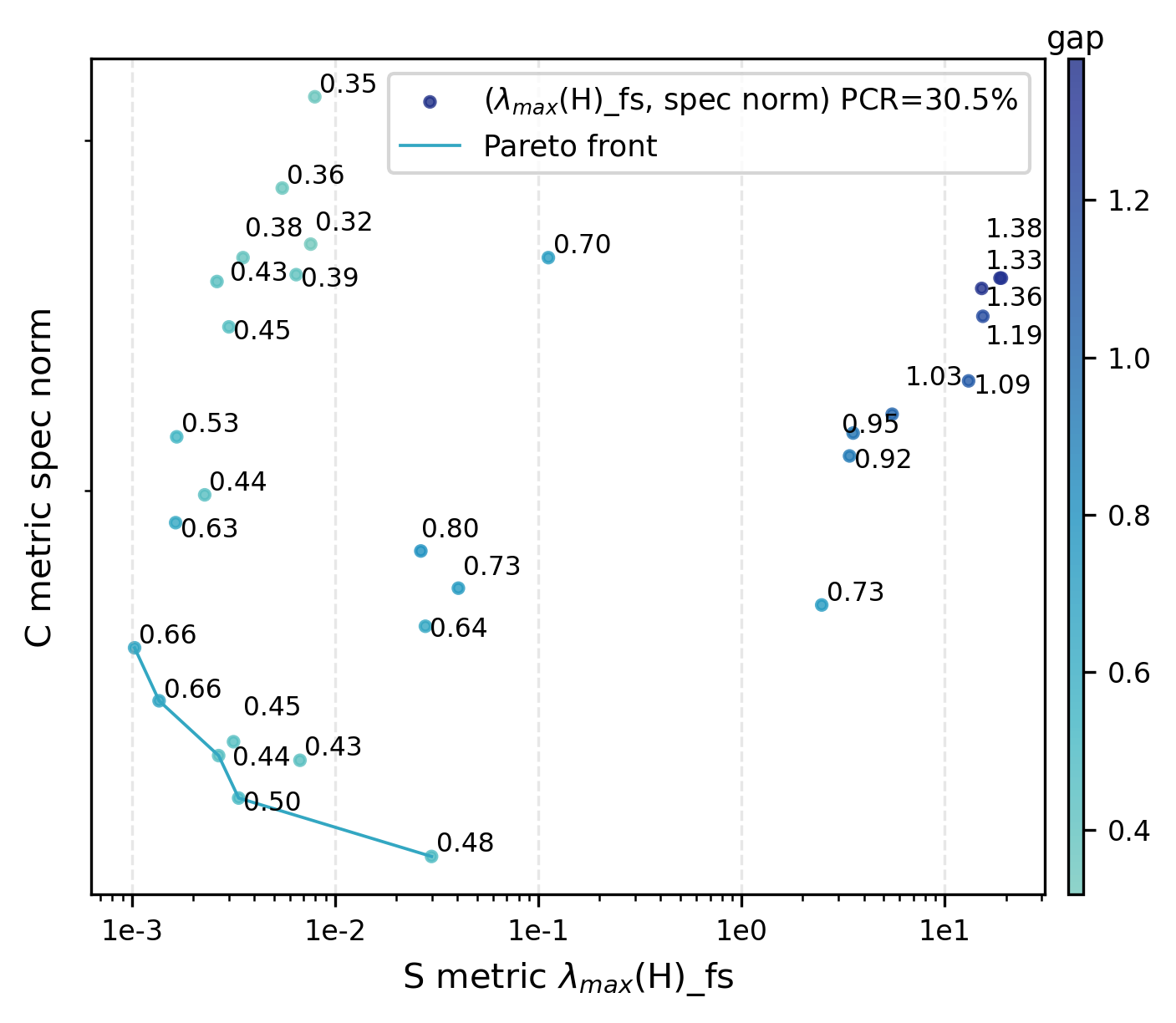}\hfill
    \par\smallskip
    {\scriptsize Spec norm paired with four sharpness metrics.\par}
    \medskip

    \caption{Visualized PCRs for the sgd-trained ResNet18 setting, corresponding to the sixteen PCR values reported in the first vertical block in Table~\ref{tab:main_result_table}.
    Lighter colors indicate smaller generalization gaps.
    Under the expected monotonic relationship, the lower-left point in each candidate pair should have a smaller generalization gap and therefore a lighter color, than the upper-right point.
    If this expected ordering is never violated, the PCR is 0\%;
    if it is frequently violated, the PCR becomes high.}
    \label{fig:PCR_visual_resnet18}
\end{figure}

\newpage

\subsection{Explanatory power of single factors}
\label{subap:single_factor_analysis}




To assess the benefits of the joint view relative to the single-factor view,
we examine how well \textbf{individual factors} explain generalization.

Following the order-based criteria of our two-factor Pareto analysis,
we quantify the disagreement between the ordering induced by a single-factor
metric and that of the generalization gap.
A pair is contradictory if the model with the smaller metric value
has the larger generalization gap.
We define the \emph{\textbf{contradiction ratio}} as
\begin{equation*}
\scalebox{0.90}{$\displaystyle
    \mathrm{CR}
    :=
    \frac{
    \#\{\text{contradictory pairs}\}
    }{
    \#\{\text{pairs in the evaluated collection}\}
    },
$}
\end{equation*}
where the denominator equals {\footnotesize $N(N-1)/2$}
for a collection of {\footnotesize $N$} well-trained models.
We adopt CR as a natural one-dimensional analogue of PCR,
allowing the single-factor and two-factor views to be assessed
from a common order-based perspective.

Tables~\ref{tab:single_factor_analysis_individual},
\ref{tab:single_factor_analysis_mixedsetting},
and~\ref{tab:single_factor_analysis_adamw}
report the single-factor evaluation results,
complementing the two-factor results in
Tables~\ref{tab:main_result_table},
\ref{tab:main_result_table_mixedsetting},
and~\ref{tab:trained_by_adamw}.
We have the following observations.

\begin{enumerate}
[   leftmargin=3.0em, labelsep=0.5em,
    topsep=3pt,
    itemsep=2pt,
    parsep=0pt,
    partopsep=0pt
]
    \item \textbf{Single-factor explanations are limited and inconsistent across settings.}
    None of the eight evaluated single-factor metrics consistently achieves low CR across settings. Even the individual components of the strongly explanatory pair $\SCtrfsFunc$ exhibit substantial ordering violations in some settings; see Tables~\ref{tab:single_factor_analysis_individual},
    \ref{tab:single_factor_analysis_mixedsetting},
    and~\ref{tab:single_factor_analysis_adamw}.
    These results provide evidence that
    \emph{single factors are insufficient} to consistently
    explain variations in generalization, 
    motivating the joint assessment studied in this work.

    \item \textbf{Sharpness and complexity provide complementary information.}
    For sgd-trained VGG13 in Table~\ref{tab:single_factor_analysis_individual}, path norm is the strongest individual metric, with CR 16.6\% and $R^2$ 0.83. Yet the strongest pair in this setting is $\SCtrfsFunc$, achieving PCR 2.8\% and $R^2$ 0.97 (Table~\ref{tab:main_result_table_repeated}). Its components individually yield CRs 21.2\% and 24.1\%, with $R^2$ 0.82 and 0.30, respectively. 
    This illustrates how \emph{two metrics can complement each other} despite neither being the strongest individual metric. Thus, individual performance alone does not determine which metrics form the strongest pair.
    
    \item \textbf{The gains of the joint view depend on the training setting.}
    For sgd-trained ResNet18 no-BN, function norm is the strongest
    single-factor metric, with CR 16.2\% and $R^2$ 0.72
    (Table~\ref{tab:single_factor_analysis_individual}).
    The joint pair $\SCtrfsFunc$ achieves PCR 1.5\% and $R^2$ 0.85
    (Table~\ref{tab:main_result_table_repeated}),
    showing a substantial improvement in explanatory power.
    By comparison, this benefit is less pronounced for the sgd-trained WideResNet no-BN, where the lowest single-factor CR is 29.1\% and the lowest two-factor PCR is 27.7\%. This indicates that the explanatory power remains limited
    even when the two factors are considered jointly in this setting.
    
\end{enumerate}

\newpage
\begin{table}[H]
\captionsetup{size=small}
\caption{
\textbf{Single-factor analysis.} 
As in the two-factor analysis, each vertical block corresponds to an individual setting.
Within each block, results are reported separately for each sharpness or complexity metric introduced
in Section~\ref{sec:metrics_review}.
\emph{Low} CR is prioritized as evidence of
\emph{strong explanatory power}, while high $R^2$ with positive
regression coefficient further supports a linear trend.
Finite-scale sharpness uses the large scale {\footnotesize$\sigma\!=\!0.50$}.
}
\vspace{-10pt}
\centering
\footnotesize
\setlength{\tabcolsep}{8pt}
\setlength{\fboxsep}{0.2pt}
\renewcommand{\arraystretch}{1.0}

\resizebox{0.91\linewidth}{!}{%
\begin{tabular}{
    @{}l
    *{3}{ 
        >{\raggedleft\arraybackslash}p{1.0cm}
        >{\centering\arraybackslash}p{1.45cm}
        !{\SCbodysep}
    }
    >{\raggedleft\arraybackslash}p{1.00cm}
    >{\centering\arraybackslash}p{1.45cm}
    @{}
}
\toprule 
\multicolumn{1}{@{}c}{%
    \multirow{4}{*}{%
        \makecell[c]{%
            \textbf{\textit{Single}-factor}\\
            \textbf{Metric}%
        }%
    }%
}
& \multicolumn{8}{c}{\textbf{CIFAR-10}} \\[-3pt]

& \multicolumn{2}{c}{\textbf{ResNet18} \scriptsize{sgd}}
& \multicolumn{2}{c}{\textbf{VGG13} \scriptsize{sgd}}
& \multicolumn{2}{c}{\textbf{ResNet18 no-BN} \scriptsize{sgd}}
& \multicolumn{2}{c}{\textbf{VGG13 no-BN} \scriptsize{sgd}} \\[-3pt]

\cmidrule(lr){2-3}
\cmidrule(lr){4-5}
\cmidrule(lr){6-7}
\cmidrule(lr){8-9}

& \multicolumn{1}{r}{\textbf{CR}}
& \multicolumn{1}{c}{$\boldsymbol{R^2}$ \scriptsize{$S$ or $C$}}

& \multicolumn{1}{r}{\textbf{CR}}
& \multicolumn{1}{c}{$\boldsymbol{R^2}$ \scriptsize{$S$ or $C$}}

& \multicolumn{1}{r}{\textbf{CR}}
& \multicolumn{1}{c}{$\boldsymbol{R^2}$ \scriptsize{$S$ or $C$}}

& \multicolumn{1}{r}{\textbf{CR}}
& \multicolumn{1}{c}{$\boldsymbol{R^2}$ \scriptsize{$S$ or $C$}}
\\[-3pt]

& \multicolumn{8}{c}{%
    \scriptsize 
    \qquad CR$\downarrow$: lower is better 
    \qquad
    $R^2$$\uparrow$: higher with {\tiny$(+)$} is better %
} \\[-3pt]

\midrule

$\SCsharpTr$
& 55.4\% & \SCrsq{0.05}{(+)}
& 63.3\% & \SCrsq{0.01}{(-)}
& 63.4\% & \SCrsq{0.07}{(-)}
& 56.8\% & \SCrsq{0.05}{(-)}\\

$\SCsharpMax$
& 57.2\% & \SCrsq{0.03}{(+)}
& 63.6\% & \SCrsq{0.00}{(-)}
& 59.5\% & \SCrsq{0.03}{(-)}
& 55.6\% & \SCrsq{0.05}{(-)}\\

$\SCsharpTrFS$
& 13.3\% & \SCrsq{0.86}{(+)}
& 21.2\% & \SCrsq{0.82}{(+)}
& 19.1\% & \SCrsq{0.53}{(+)}
& 17.5\% & \SCrsq{0.70}{(+)}\\

$\SCsharpMaxFS$
& 26.9\% & \SCrsq{0.80}{(+)}
& 32.8\% & \SCrsq{0.58}{(+)}
& 30.2\% & \SCrsq{0.35}{(+)}
& 27.8\% & \SCrsq{0.68}{(+)}\\

\addlinespace[2pt]

func norm
& 25.7\% & \SCrsq{0.20}{(+)}
& 24.1\% & \SCrsq{0.30}{(+)}
& 16.2\% & \SCrsq{0.72}{(+)}
& 13.1\% & \SCrsq{0.68}{(+)}\\

param norm
& 14.0\% & \SCrsq{0.68}{(+)}
& 18.2\% & \SCrsq{0.65}{(+)}
& 36.4\% & \SCrsq{0.20}{(+)}
& 46.7\% & \SCrsq{0.42}{(+)}\\

path norm 
& 8.3\%  & \SCrsq{0.90}{(+)}
& 16.6\% & \SCrsq{0.83}{(+)}
& 39.1\% & \SCrsq{0.18}{(+)}
& 49.2\% & \SCrsq{0.42}{(+)}\\

spec norm
& 48.7\% & \SCrsq{0.03}{(+)}
& 40.3\% & \SCrsq{0.16}{(+)}
& 46.2\% & \SCrsq{0.15}{(+)}
& 51.0\% & \SCrsq{0.44}{(+)}\\

\midrule

\multicolumn{1}{@{}c}{%
    \multirow{3}{*}{%
        \makecell[c]{%
            \textbf{\textit{Single}-factor}\\
            \textbf{Metric}%
        }%
    }%
}
& \multicolumn{8}{c}{\textbf{CIFAR-100}} \\[-3pt]

& \multicolumn{2}{c}{\textbf{WideResNet} \scriptsize{sgd}}
& \multicolumn{2}{c}{\textbf{ViT} \scriptsize{adamw}}
& \multicolumn{2}{c}{\textbf{WideResNet no-BN} \scriptsize{sgd}}
& \multicolumn{2}{c}{\textbf{ViT no-LN} \scriptsize{adamw}} \\[-3pt]

\cmidrule(lr){2-3}
\cmidrule(lr){4-5}
\cmidrule(lr){6-7}
\cmidrule(lr){8-9}

& \multicolumn{1}{r}{\textbf{CR}}
& \multicolumn{1}{c}{$\boldsymbol{R^2}$ \scriptsize{$S$ or $C$}}
& \multicolumn{1}{r}{\textbf{CR}}
& \multicolumn{1}{c}{$\boldsymbol{R^2}$ \scriptsize{$S$ or $C$}}
& \multicolumn{1}{r}{\textbf{CR}}
& \multicolumn{1}{c}{$\boldsymbol{R^2}$ \scriptsize{$S$ or $C$}}
& \multicolumn{1}{r}{\textbf{CR}}
& \multicolumn{1}{c}{$\boldsymbol{R^2}$ \scriptsize{$S$ or $C$}}
\\[-3pt]


\midrule

$\SCsharpTr$
& 46.0\% & \SCrsq{0.42}{(+)}
& 84.8\% & \SCrsq{0.05}{(-)}
& 58.1\% & \SCrsq{0.03}{(+)}
& 39.4\% & \SCrsq{0.03}{(+)}\\

$\SCsharpMax$
& 42.0\% & \SCrsq{0.52}{(+)}
& 77.3\% & \SCrsq{0.05}{(-)}
& 59.5\% & \SCrsq{0.01}{(-)}
& 39.4\% & \SCrsq{0.12}{(+)}\\

$\SCsharpTrFS$
& 5.8\%  & \SCrsq{0.88}{(+)}
& 26.6\% & \SCrsq{0.56}{(+)}
& 46.3\% & \SCrsq{0.00}{(-)}
& 32.0\% & \SCrsq{0.28}{(+)}\\

$\SCsharpMaxFS$
& 12.3\% & \SCrsq{0.81}{(+)}
& 55.3\% & \SCrsq{0.02}{(-)}
& 49.1\% & \SCrsq{0.03}{(-)}
& 26.5\% & \SCrsq{0.35}{(+)}\\

\addlinespace[2pt]

func norm
& 10.9\% & \SCrsq{0.74}{(+)}
& 12.0\% & \SCrsq{0.80}{(+)}
& 29.1\% & \SCrsq{0.28}{(+)}
& 17.2\% & \SCrsq{0.70}{(+)}\\

param norm
& 11.2\% & \SCrsq{0.70}{(+)}
& 13.0\% & \SCrsq{0.77}{(+)}
& 41.9\% & \SCrsq{0.00}{(+)}
& 54.0\% & \SCrsq{0.01}{(-)}\\

path norm 
& 10.5\% & \SCrsq{0.78}{(+)}
& -- & --
& 44.0\% & \SCrsq{0.00}{(-)}
& -- & --\\

spec norm
& 50.0\% & \SCrsq{0.01}{(+)}
& -- & --
& 43.2\% & \SCrsq{0.00}{(-)}
& -- & --\\

\bottomrule
\end{tabular}
}
\label{tab:single_factor_analysis_individual}
\end{table}

\vspace{-0.5em}

\begin{table}[H]
\captionsetup{size=small} 
\caption{
\textbf{Two-factor analysis.} 
Table~\ref{tab:main_result_table} is replicated here for convenient reference.
}
\vspace{-10pt}
\centering
\footnotesize
\setlength{\tabcolsep}{5pt}
\setlength{\fboxsep}{0.2pt}
\renewcommand{\arraystretch}{1.0}

\resizebox{0.91\linewidth}{!}{%
\begin{tabular}{
    @{}l
    *{3}{ 
        >{\raggedleft\arraybackslash}p{1.0cm}
        >{\centering\arraybackslash}p{1.45cm}
        !{\SCbodysep}
    }
    >{\raggedleft\arraybackslash}p{1.00cm}
    >{\centering\arraybackslash}p{1.45cm}
    @{}
}
\toprule 
\multicolumn{1}{@{}c}{%
    \multirow{4}{*}{%
        \makecell[c]{%
            $\boldsymbol{(S,C)}$\\
            \textbf{Metric \textit{Pair}}%
        }%
    }%
}
& \multicolumn{8}{c}{\textbf{CIFAR-10}} \\[-3pt]

& \multicolumn{2}{c}{\textbf{ResNet18} \scriptsize{sgd}}
& \multicolumn{2}{c}{\textbf{VGG13} \scriptsize{sgd}}
& \multicolumn{2}{c}{\textbf{ResNet18 no-BN} \scriptsize{sgd}}
& \multicolumn{2}{c}{\textbf{VGG13 no-BN} \scriptsize{sgd}} \\[-3pt]

\cmidrule(lr){2-3}
\cmidrule(lr){4-5}
\cmidrule(lr){6-7}
\cmidrule(lr){8-9}

& \multicolumn{1}{r}{\textbf{PCR}}
& \multicolumn{1}{c}{$\boldsymbol{R^2}\,(S,C)$}

& \multicolumn{1}{r}{\textbf{PCR}}
& \multicolumn{1}{c}{$\boldsymbol{R^2}\,(S,C)$}

& \multicolumn{1}{r}{\textbf{PCR}}
& \multicolumn{1}{c}{$\boldsymbol{R^2}\,(S,C)$}

& \multicolumn{1}{r}{\textbf{PCR}}
& \multicolumn{1}{c}{$\boldsymbol{R^2}\,(S,C)$}
\\[-3pt]

& \multicolumn{8}{c}{%
    \scriptsize 
    \qquad PCR$\downarrow$: lower is better 
    \qquad
    $R^2$$\uparrow$: higher with {\tiny$(+,+)$} is better %
} \\[-3pt]

\midrule

\SCtrFunc
& 1.2\% & \SCrsq{0.48}{(+,+)}
& 8.2\% & \SCrsq{0.31}{(+,+)}
& 4.1\% & \SCrsq{0.79}{(+,+)}
& 5.4\% & \SCrsq{0.68}{(+,+)}\\

\SClmaxFunc
& 2.6\% & \SCrsq{0.45}{(+,+)}
& 8.4\% & \SCrsq{0.31}{(+,+)}
& 3.9\% & \SCrsq{0.79}{(+,+)} 
& 5.8\% & \SCrsq{0.68}{(+,+)}\\

\iteSCtrfsFunc
& 1.5\% & \SCrsq{0.95}{(+,+)}
& 2.8\% & \SCrsq{0.97}{(+,+)} 
& 1.5\% & \SCrsq{0.85}{(+,+)} 
& 3.4\% & \SCrsq{0.85}{(+,+)}\\

\iteSClmaxfsFunc
& 1.9\% & \SCrsq{0.92}{(+,+)}
& 4.5\% & \SCrsq{0.87}{(+,+)}
& 1.7\% & \SCrsq{0.82}{(+,+)} 
& 4.3\% & \SCrsq{0.85}{(+,+)}\\

\addlinespace[2pt]

\SCtrLtwo
& 3.5\%  & \SCrsq{0.84}{(+,+)}
& 13.4\% & \SCrsq{0.70}{(+,+)} 
& 49.5\% & \SCrsq{0.21}{(-,+)} 
& 61.0\% & \SCrsq{0.42}{(-,+)} \\

\SClmaxLtwo
& 4.4\%  & \SCrsq{0.84}{(+,+)}
& 13.0\% & \SCrsq{0.69}{(+,+)} 
& 33.0\% & \SCrsq{0.20}{(-,+)}  
& 59.0\% & \SCrsq{0.42}{(+,+)} \\

\iteSCtrfsLtwo
& 3.0\% & \SCrsq{0.98}{(+,+)}
& 6.0\% & \SCrsq{0.95}{(+,+)}
& 3.4\% & \SCrsq{0.68}{(+,+)}  
& 3.0\% & \SCrsq{0.72}{(+,-)} \\

\iteSClmaxfsLtwo
& 3.6\% & \SCrsq{0.97}{(+,+)}
& 8.5\% & \SCrsq{0.92}{(+,+)}
& 6.0\% & \SCrsq{0.56}{(+,+)} 
& 5.6\% & \SCrsq{0.70}{(+,-)} \\

\addlinespace[2pt]

\SCtrPath 
& 15.1\% & \SCrsq{0.90}{(+,+)}
& 31.1\% & \SCrsq{0.85}{(-,+)}
& 61.0\% & \SCrsq{0.19}{(-,+)}
& 66.7\% & \SCrsq{0.42}{(-,+)}\\

\SClmaxPath
& 15.7\% & \SCrsq{0.90}{(+,+)}
& 31.0\% & \SCrsq{0.84}{(-,+)}
& 47.2\% & \SCrsq{0.18}{(-,+)}
& 68.3\% & \SCrsq{0.42}{(-,+)}\\ 

\iteSCtrfsPath
& 8.6\% & \SCrsq{0.96}{(+,+)}
& 16.8\% &\SCrsq{0.86}{(+,+)}
& 5.1\% & \SCrsq{0.68}{(+,+)}
& 3.2\% & \SCrsq{0.72}{(+,-)}\\

\iteSClmaxfsPath
& 9.9\% & \SCrsq{0.96}{(+,+)}
& 19.6\% &\SCrsq{0.84}{(+,+)}
& 7.7\% & \SCrsq{0.56}{(+,+)}
& 6.1\% & \SCrsq{0.70}{(+,-)}\\

\addlinespace[2pt]

\SCtrSpec
& 54.0\% & \SCrsq{0.07}{(+,+)}
& 53.6\% & \SCrsq{0.17}{(-,+)} 
& 80.6\% & \SCrsq{0.20}{(-,+)} 
& 69.1\% & \SCrsq{0.44}{(-,+)} \\

\SClmaxSpec
& 55.3\% & \SCrsq{0.05}{(+,+)}
& 54.1\% & \SCrsq{0.17}{(-,+)}
& 74.0\% & \SCrsq{0.17}{(-,+)} 
& 72.2\% & \SCrsq{0.44}{(-,+)} \\

\iteSCtrfsSpec
& 14.0\% & \SCrsq{0.88}{(+,-)}
& 21.0\% & \SCrsq{0.83}{(+,-)}
&  4.4\% & \SCrsq{0.66}{(+,+)} 
&  3.4\% & \SCrsq{0.75}{(+,-)} \\

\iteSClmaxfsSpec
& 30.5\% & \SCrsq{0.82}{(+,-)}
& 31.4\% & \SCrsq{0.58}{(+,-)} 
&  7.8\% & \SCrsq{0.53}{(+,+)} 
&  6.7\% & \SCrsq{0.73}{(+,-)} \\[-2pt]

\midrule

\multicolumn{1}{@{}c}{%
    \multirow{3}{*}{%
        \makecell[c]{%
            $\boldsymbol{(S,C)}$\\
            \textbf{Metric Pair}%
        }%
    }%
}
& \multicolumn{8}{c}{\textbf{CIFAR-100}} \\[-3pt]

& \multicolumn{2}{c}{\textbf{WideResNet} \scriptsize{sgd}}
& \multicolumn{2}{c}{\textbf{ViT} \scriptsize{adamw}}
& \multicolumn{2}{c}{\textbf{WideResNet no-BN} \scriptsize{sgd}}
& \multicolumn{2}{c}{\textbf{ViT no-LN} \scriptsize{adamw}} \\[-3pt]

\cmidrule(lr){2-3}
\cmidrule(lr){4-5}
\cmidrule(lr){6-7}
\cmidrule(lr){8-9}

& \multicolumn{1}{r}{\textbf{PCR}}
& \multicolumn{1}{c}{$\boldsymbol{R^2}\,(S,C)$}
& \multicolumn{1}{r}{\textbf{PCR}}
& \multicolumn{1}{c}{$\boldsymbol{R^2}\,(S,C)$}
& \multicolumn{1}{r}{\textbf{PCR}}
& \multicolumn{1}{c}{$\boldsymbol{R^2}\,(S,C)$}
& \multicolumn{1}{r}{\textbf{PCR}}
& \multicolumn{1}{c}{$\boldsymbol{R^2}\,(S,C)$}
\\[-3pt]


\midrule

\SCtrFunc
&  4.6\% & \SCrsq{0.88}{(+,+)} 
& 40.6\% & \SCrsq{0.80}{(-,+)} 
& 27.7\% & \SCrsq{0.40}{(+,+)} 
& 16.7\% & \SCrsq{0.70}{(+,+)}  \\

\SClmaxFunc
&  2.9\% & \SCrsq{0.88}{(+,+)} 
& 26.2\% & \SCrsq{0.80}{(-,+)} 
& 31.1\% & \SCrsq{0.29}{(+,+)} 
& 16.5\% & \SCrsq{0.72}{(+,+)} \\

\iteSCtrfsFunc
& 1.7\% & \SCrsq{0.95}{(+,+)} 
& 9.4\% & \SCrsq{0.82}{(+,+)} 
& 31.3\% &\SCrsq{0.32}{(-,+)} 
& 12.8\% & \SCrsq{0.74}{(+,+)} \\

\iteSClmaxfsFunc
& 1.8\% & \SCrsq{0.92}{(+,+)} 
& 14.5\% & \SCrsq{0.80}{(-,+)} 
& 31.5\% & \SCrsq{0.36}{(-,+)} 
& 13.2\% & \SCrsq{0.74}{(+,+)}  \\

\addlinespace[2pt]

\SCtrLtwo
& 2.4\% & \SCrsq{0.93}{(+,+)} 
& 39.8\% & \SCrsq{0.78}{(-,+)} 
& 51.0\% & \SCrsq{0.05}{(+,+)} 
& 24.5\% & \SCrsq{0.04}{(+,+)} \\

\SClmaxLtwo
& 1.5\% & \SCrsq{0.91}{(+,+)} 
& 18.3\% & \SCrsq{0.78}{(-,+)} 
& 54.4\% & \SCrsq{0.01}{(-,-)} 
& 17.9\% & \SCrsq{0.13}{(+,+)}\\

\iteSCtrfsLtwo
& 2.1\% & \SCrsq{0.96}{(+,+)} 
& 7.0\% & \SCrsq{0.78}{(+,+)}  
& 40.0\% & \SCrsq{0.01}{(-,+)} 
& 39.5\% & \SCrsq{0.47}{(+,-)}\\

\iteSClmaxfsLtwo
& 2.3\% & \SCrsq{0.95}{(+,+)} 
& 11.0\% & \SCrsq{0.78}{(-,+)} 
& 41.4\% & \SCrsq{0.04}{(-,+)} 
& 20.2\% & \SCrsq{0.35}{(+,-)}\\

\addlinespace[2pt]

\SCtrPath
& 12.8\% & \SCrsq{0.90}{(+,+)} 
& -- & -- 
& 56.9\% & \SCrsq{0.03}{(+,+)} 
& -- & -- \\

\SClmaxPath
& 8.4\% & \SCrsq{0.91}{(+,+)} 
& -- & -- 
& 61.8\% & \SCrsq{0.02}{(-,-)} 
& -- & -- \\

\iteSCtrfsPath
& 4.7\% & \SCrsq{0.96}{(+,+)} 
& -- & -- 
& 41.9\% & \SCrsq{0.01}{(+,-)} 
& -- & -- \\

\iteSClmaxfsPath
& 4.3\% & \SCrsq{0.95}{(+,+)} 
& -- & -- 
& 43.3\% & \SCrsq{0.03}{(-,-)} 
& -- & -- \\

\addlinespace[2pt]

\SCtrSpec
& 47.4\% & \SCrsq{0.43}{(+,-)} 
& -- & -- 
& 53.0\% & \SCrsq{0.03}{(+,+)} 
& -- & -- \\

\SClmaxSpec
& 43.9\% & \SCrsq{0.53}{(+,-)} 
& -- & -- 
& 57.4\% & \SCrsq{0.03}{(-,-)} 
& -- & -- \\

\iteSCtrfsSpec
& 5.8\% & \SCrsq{0.90}{(+,-)} 
& -- & -- 
& 38.5\% & \SCrsq{0.01}{(-,-)} 
& -- & -- \\

\iteSClmaxfsSpec
& 10.8\% & \SCrsq{0.83}{(+,-)} 
& -- & -- 
& 39.6\% & \SCrsq{0.05}{(-,-)} 
& -- & -- \\
\bottomrule
\end{tabular}
}
\label{tab:main_result_table_repeated}
\end{table}

\newpage
\begin{table}[H]
\captionsetup{size=small}
\caption{
\textbf{Single-factor analysis} (mixed settings).
}
\vspace{-10pt}
\centering
\footnotesize
\setlength{\tabcolsep}{6pt}
\setlength{\fboxsep}{0.2pt}
\renewcommand{\arraystretch}{1.1}

\resizebox{0.99\linewidth}{!}{%
\begin{tabular}{
    @{}l
    *{3}{ 
        >{\raggedleft\arraybackslash}p{1.0cm}
        >{\centering\arraybackslash}p{1.45cm}
        !{\SCbodysep}
    }
    >{\raggedleft\arraybackslash}p{1.00cm}
    >{\centering\arraybackslash}p{1.45cm}
    @{}
}
\toprule 
\multicolumn{1}{@{}c}{%
    \multirow{2}{*}{%
        \makecell[c]{%
            \textbf{\textit{Single}-factor}\\
            \textbf{Metric}%
        }%
    }%
}
& \multicolumn{2}{c}{%
    \makecell[c]{%
        \multirow{2}{*}{\textbf{Cross-architecture}}\\
        {}\\
        {\scriptsize mixed four}
    }%
}
& \multicolumn{2}{c}{%
    \makecell[c]{%
        \textbf{Cross-optimizer}\\
        \textbf{\;\;\& Cross-architecture}\\
        {\scriptsize mixed eight}
    }%
}
& \multicolumn{4}{c}{%
    \makecell[c]{%
        \multirow{2}{*}{\textbf{Cross-dataset \& Cross-architecture}}\\
        {}\\
        {\scriptsize mixed three\qquad\qquad\qquad\qquad mixed six}
    }%
}   \\[-1pt]

\cmidrule(lr){2-3}
\cmidrule(lr){4-5}
\cmidrule(lr){6-9}

& \multicolumn{1}{r}{\textbf{CR}}
& \multicolumn{1}{c}{$\boldsymbol{R^2}$ \scriptsize{$S$ or $C$}}

& \multicolumn{1}{r}{\textbf{CR}}
& \multicolumn{1}{c}{$\boldsymbol{R^2}$ \scriptsize{$S$ or $C$}}

& \multicolumn{1}{r}{\textbf{CR}}
& \multicolumn{1}{c}{$\boldsymbol{R^2}$ \scriptsize{$S$ or $C$}}

& \multicolumn{1}{r}{\textbf{CR}}
& \multicolumn{1}{c}{$\boldsymbol{R^2}$ \scriptsize{$S$ or $C$}}
\\[-3pt]

\midrule

$\SCsharpTr$
& 68.9\% & \SCrsq{0.02}{(-)}
& 67.5\% & \SCrsq{0.01}{(-)}
& 54.1\% & \SCrsq{0.01}{(-)}
& 48.5\% & \SCrsq{0.04}{(+)}\\

$\SCsharpMax$
& 67.0\% & \SCrsq{0.01}{(-)}
& 67.3\% & \SCrsq{0.01}{(-)}
& 58.6\% & \SCrsq{0.01}{(-)}
& 48.1\% & \SCrsq{0.02}{(+)}\\

$\SCsharpTrFS$
& 36.1\% & \SCrsq{0.02}{(+)}
& 33.5\% & \SCrsq{0.13}{(+)}
& 20.8\% & \SCrsq{0.69}{(+)}
& 30.3\% & \SCrsq{0.04}{(+)}\\

$\SCsharpMaxFS$
& 38.3\% & \SCrsq{0.04}{(+)}
& 38.3\% & \SCrsq{0.04}{(+)}
& 25.8\% & \SCrsq{0.60}{(+)}
& 30.4\% & \SCrsq{0.10}{(+)}\\

\addlinespace[4pt]

func norm
& 12.0\% & \SCrsq{0.81}{(+)}
& 11.8\% & \SCrsq{0.25}{(+)}
& 16.2\% & \SCrsq{0.81}{(+)}
& 10.2\% & \SCrsq{0.85}{(+)}\\

param norm
& 39.4\% & \SCrsq{0.17}{(+)}
& 37.1\% & \SCrsq{0.00}{(-)}
& 15.9\% & \SCrsq{0.80}{(+)}
& 46.3\% & \SCrsq{0.01}{(+)}\\

path norm 
& 47.2\% & \SCrsq{0.10}{(+)}
& 31.6\% & \SCrsq{0.60}{(+)}
& 21.9\% & \SCrsq{0.34}{(+)}
& 52.5\% & \SCrsq{0.01}{(-)}\\

spec norm
& 43.2\% & \SCrsq{0.10}{(+)}
& 31.1\% & \SCrsq{0.25}{(+)}
& 38.7\% & \SCrsq{0.10}{(+)}
& 53.7\% & \SCrsq{0.00}{(+)}\\[-2pt]

\bottomrule
\end{tabular}
}
\label{tab:single_factor_analysis_mixedsetting}
\end{table}

\begin{table}[H]
\captionsetup{size=small}
\caption{
\textbf{Two-factor analysis} (mixed settings).
Table~\ref{tab:main_result_table_mixedsetting} is replicated here for convenient reference.
}
\vspace{-10pt}
\centering
\footnotesize
\setlength{\tabcolsep}{5pt}
\setlength{\fboxsep}{0.2pt}
\renewcommand{\arraystretch}{1.15}

\resizebox{0.99\linewidth}{!}{%
\begin{tabular}{
    @{}l
    *{3}{
        >{\raggedleft\arraybackslash}p{1.05cm}
        >{\centering\arraybackslash}p{1.50cm}
        !{\SCbodysep}
    }
    >{\raggedleft\arraybackslash}p{1.00cm}
    >{\centering\arraybackslash}p{1.40cm}
    @{}
}
\toprule

\multicolumn{1}{@{}c}{%
    \multirow{2}{*}{%
        \makecell[c]{%
            $\boldsymbol{(S,C)}$\\
            \textbf{Metric \textit{Pair}}%
        }%
    }%
}
& \multicolumn{2}{c}{%
    \makecell[c]{%
        \multirow{2}{*}{\textbf{Cross-architecture}}\\
        {}\\
        {\scriptsize mixed four}
    }%
}
& \multicolumn{2}{c}{%
    \makecell[c]{%
        \textbf{Cross-optimizer}\\
        \textbf{\;\;\& Cross-architecture}\\
        {\scriptsize mixed eight}
    }%
}
& \multicolumn{4}{c}{%
    \makecell[c]{%
        \multirow{2}{*}{\textbf{Cross-dataset \& Cross-architecture}}\\
        {}\\
        {\scriptsize mixed three\qquad\qquad\qquad\qquad mixed six}
    }%
}   \\[-1pt]

\cmidrule(lr){2-3}
\cmidrule(lr){4-5}
\cmidrule(lr){6-9}

& \multicolumn{1}{r}{\textbf{PCR}}
& \multicolumn{1}{c}{$\boldsymbol{R^2}\,(S,C)$}
& \multicolumn{1}{r}{\textbf{PCR}}
& \multicolumn{1}{c}{$\boldsymbol{R^2}\,(S,C)$}
& \multicolumn{1}{r}{\textbf{PCR}}
& \multicolumn{1}{c}{$\boldsymbol{R^2}\,(S,C)$}
& \multicolumn{1}{r}{\textbf{PCR}}
& \multicolumn{1}{c}{$\boldsymbol{R^2}\,(S,C)$}
\\[-3pt]


\midrule

\SCtrFunc 
& 4.1\% & \SCrsq{0.81}{(+,+)} 
& 11.5\%& \SCrsq{0.26}{(-,+)} 
& 2.3\% & \SCrsq{0.83}{(+,+)} 
& 2.2\% & \SCrsq{0.86}{(+,+)} \\
\SClmaxFunc     
& 4.1\% & \SCrsq{0.81}{(+,+)}  
& 11.6\%& \SCrsq{0.25}{(-,+)} 
& 2.2\% & \SCrsq{0.81}{(+,+)} 
& 2.9\% & \SCrsq{0.86}{(+,+)} \\
\SCtrfsFunc    
& 2.1\% & \SCrsq{0.88}{(+,+)}  
& 6.6\% & \SCrsq{0.32}{(+,+)}
& 3.0\% & \SCrsq{0.92}{(+,+)}  
& 2.7\% & \SCrsq{0.87}{(+,+)} \\
\SClmaxfsFunc    
& 1.7\% & \SCrsq{0.89}{(+,+)}  
& 6.1\% & \SCrsq{0.25}{(+,+)}
& 2.4\% & \SCrsq{0.92}{(+,+)}  
& 2.4\% & \SCrsq{0.87}{(+,+)} \\

\addlinespace[2pt]

\SCtrLtwo       
& 65.6\% &\SCrsq{0.18}{(-,+)}  
& 64.6\%& \SCrsq{0.01}{(-,-)} 
& 8.5\% & \SCrsq{0.81}{(+,+)} 
& 41.0\% & \SCrsq{0.05}{(+,+)} \\
\SClmaxLtwo   
& 62.3\% &\SCrsq{0.17}{(-,+)}  
& 63.5\%& \SCrsq{0.01}{(-,-)} 
& 9.8\% & \SCrsq{0.81}{(+,+)}  
& 39.8\% & \SCrsq{0.03}{(+,+)} \\
\SCtrfsLtwo      
& 29.9\% &\SCrsq{0.17}{(+,+)}  
& 27.0\%& \SCrsq{0.14}{(+,-)} 
& 9.2\% & \SCrsq{0.89}{(+,+)}  
& 30.7\% &\SCrsq{0.04}{(+,-)} \\
\SClmaxfsLtwo   
& 27.7\% & \SCrsq{0.19}{(+,+)}  
& 26.3\% & \SCrsq{0.05}{(+,-)} 
& 8.4\% & \SCrsq{0.88}{(+,+)} 
& 27.8\% & \SCrsq{0.11}{(+,-)} \\

\addlinespace[2pt]

\SCtrPath      
& 70.1\% & \SCrsq{0.11}{(-,+)}   
& 49.4\%& \SCrsq{0.61}{(-,+)} 
& 28.2\% & \SCrsq{0.35}{(-,+)} 
& 51.9\% & \SCrsq{0.04}{(+,-)}\\
\SClmaxPath      
& 68.5\% & \SCrsq{0.10}{(-,+)}  
& 49.1\%& \SCrsq{0.61}{(-,+)} 
& 30.8\% & \SCrsq{0.35}{(-,+)} 
& 51.6\% & \SCrsq{0.02}{(+,-)}\\
\SCtrfsPath     
& 36.8\% & \SCrsq{0.10}{(+,+)}  
& 25.8\%& \SCrsq{0.61}{(+,+)} 
& 17.1\% & \SCrsq{0.75}{(+,+)} 
& 36.1\% & \SCrsq{0.05}{(+,-)}\\
\SClmaxfsPath  
& 36.8\% & \SCrsq{0.11}{(+,+)}  
& 25.2\%& \SCrsq{0.61}{(+,+)} 
& 17.9\% & \SCrsq{0.67}{(+,+)}  
& 34.2\% & \SCrsq{0.12}{(+,-)} \\

\addlinespace[2pt]

\SCtrSpec        
& 66.8\% & \SCrsq{0.11}{(-,+)}  
& 47.5\%& \SCrsq{0.26}{(-,+)} 
& 43.4\% & \SCrsq{0.11}{(-,+)} 
& 53.3\% & \SCrsq{0.04}{(+,+)}  \\
\SClmaxSpec       
& 64.5\% & \SCrsq{0.11}{(-,+)}  
& 47.2\%& \SCrsq{0.26}{(-,+)} 
& 47.5\% & \SCrsq{0.11}{(-,+)} 
& 53.0\% & \SCrsq{0.02}{(+,+)} \\
\SCtrfsSpec       
& 28.2\% & \SCrsq{0.12}{(+,+)}  
& 19.6\%& \SCrsq{0.35}{(+,+)} 
& 14.8\% & \SCrsq{0.70}{(+,+)} 
& 32.7\% & \SCrsq{0.04}{(+,+)}  \\
\SClmaxfsSpec     
& 29.7\% & \SCrsq{0.14}{(+,+)}  
& 19.3\%& \SCrsq{0.30}{(+,+)}  
& 20.1\% & \SCrsq{0.61}{(+,+)} 
& 31.4\% & \SCrsq{0.11}{(+,+)}  \\[-2pt]

\midrule

\makecell[tc]{%
    \textbf{Models being}\\
    \textbf{pooled together}\\
    {(the number of}\\ 
    {well-trained models}\\
    {in the collection)}%
}
& \multicolumn{2}{c!{\SCfootersep}}{%
    \scriptsize \makecell[tl]{%
    CIFAR-10, {sgd}\\
        \quad ResNet18 \tiny{$\;(30)$}\\
        \quad VGG13 \tiny{$\;(34)$} \\
        \quad ResNet18 no-BN \tiny{$\;(58)$} \\
        \quad VGG13 no-BN \tiny{$\;(30)$}%
    }%
}
& \multicolumn{2}{c!{\SCfootersep}}{%
    \scriptsize \makecell[tl]{%
    CIFAR-10, {sgd \& adamw}\\
        \quad ResNet18 \tiny{$\;(76)$}\\
        \quad VGG13 \tiny{$\;(73)$}\\
        \quad ResNet18 no-BN \tiny{$\;(93)$}\\
        \quad VGG13 no-BN \tiny{$\;(67)$}%
    }%
}
& \multicolumn{2}{c!{\SCfootersep}}{%
    \scriptsize \makecell[tl]{%
    CIFAR-10, {sgd}\\
        \quad ResNet18 \tiny{$\;(30)$}\\
        \quad VGG13 \tiny{$\;(34)$}\\
    CIFAR-100, {sgd}\\
        \quad WideResNet \tiny{$\;(24)$}%
    }%
}
& \multicolumn{2}{c}{%
    \scriptsize \makecell[tl]{%
    CIFAR-10, {sgd}\\
        \quad ResNet18 \tiny{$\;(30)$}\\
        \quad VGG13 \tiny{$\;(34)$}\\
        \quad ResNet18 no-BN \tiny{$\;(58)$}\\
        \quad VGG13 no-BN \tiny{$\;(30)$}\\
    CIFAR-100, {sgd}\\
        \quad WideResNet \tiny{$\;(24)$}\\
        \quad WideResNet no-BN \tiny{$(47)$}\\%
    }%
} \\[-3pt]

\makecell[c]{%
    \textbf{Collection size}
}
& \multicolumn{2}{c!{\SCbodysep}}{$152$}
& \multicolumn{2}{c!{\SCbodysep}}{$309$}
& \multicolumn{2}{c!{\SCbodysep}}{$88$}
& \multicolumn{2}{c}{$223$}
\\[-2pt]

\bottomrule
\end{tabular}%
}
\label{tab:main_result_table_mixedsetting_fullREPEAT}
\end{table}

\newpage
\begin{table}[h]
\captionsetup{size=small}
\caption{
\textbf{Single-factor analysis} (optimizer AdamW).
}
\vspace{-10pt}
\centering
\footnotesize
\setlength{\tabcolsep}{8pt}
\setlength{\fboxsep}{0.2pt}
\renewcommand{\arraystretch}{1.0}

\resizebox{0.99\linewidth}{!}{%
\begin{tabular}{
    @{}l
    *{3}{ 
        >{\raggedleft\arraybackslash}p{1.0cm}
        >{\centering\arraybackslash}p{1.45cm}
        !{\SCbodysep}
    }
    >{\raggedleft\arraybackslash}p{1.00cm}
    >{\centering\arraybackslash}p{1.45cm}
    @{}
}
\toprule 
\multicolumn{1}{@{}c}{%
    \multirow{4}{*}{%
        \makecell[c]{%
            \textbf{\textit{Single}-factor}\\
            \textbf{Metric}%
        }%
    }%
}
& \multicolumn{8}{c}{\textbf{CIFAR-10}} \\[-3pt]

& \multicolumn{2}{c}{\textbf{ResNet18} \scriptsize{adamw}}
& \multicolumn{2}{c}{\textbf{VGG13} \scriptsize{adamw}}
& \multicolumn{2}{c}{\textbf{ResNet18 no-BN} \scriptsize{adamw}}
& \multicolumn{2}{c}{\textbf{VGG13 no-BN} \scriptsize{adamw}} \\[-3pt]

\cmidrule(lr){2-3}
\cmidrule(lr){4-5}
\cmidrule(lr){6-7}
\cmidrule(lr){8-9}

& \multicolumn{1}{r}{\textbf{CR}}
& \multicolumn{1}{c}{$\boldsymbol{R^2}$ \scriptsize{$S$ or $C$}}

& \multicolumn{1}{r}{\textbf{CR}}
& \multicolumn{1}{c}{$\boldsymbol{R^2}$ \scriptsize{$S$ or $C$}}

& \multicolumn{1}{r}{\textbf{CR}}
& \multicolumn{1}{c}{$\boldsymbol{R^2}$ \scriptsize{$S$ or $C$}}

& \multicolumn{1}{r}{\textbf{CR}}
& \multicolumn{1}{c}{$\boldsymbol{R^2}$ \scriptsize{$S$ or $C$}}
\\[-3pt]


\midrule

$\SCsharpTr$
& 78.4\% & \SCrsq{0.00}{(+)}
& 55.5\% & \SCrsq{0.01}{(+)}
& 78.3\% & \SCrsq{0.29}{(-)}
& 61.7\% & \SCrsq{0.04}{(+)}\\

$\SCsharpMax$
& 78.9\% & \SCrsq{0.00}{(+)}
& 52.6\% & \SCrsq{0.01}{(+)}
& 74.8\% & \SCrsq{0.21}{(-)}
& 63.4\% & \SCrsq{0.04}{(+)}\\

$\SCsharpTrFS$
& 38.3\% & \SCrsq{0.02}{(+)}
& 21.3\% & \SCrsq{0.89}{(+)}
& 14.8\% & \SCrsq{0.27}{(+)}
& 20.1\% & \SCrsq{0.76}{(+)}\\

$\SCsharpMaxFS$
& 39.0\% & \SCrsq{0.01}{(+)}
& 20.2\% & \SCrsq{0.80}{(+)}
& 31.3\% & \SCrsq{0.04}{(+)}
& 36.9\% & \SCrsq{0.58}{(+)}\\

\addlinespace[4pt]

func norm
& 19.7\% & \SCrsq{0.53}{(+)}
& 12.6\% & \SCrsq{0.81}{(+)}
& 13.8\% & \SCrsq{0.66}{(+)}
& 24.6\% & \SCrsq{0.10}{(+)}\\

param norm
& 13.6\% & \SCrsq{0.75}{(+)}
& 20.6\% & \SCrsq{0.53}{(+)}
& 17.6\% & \SCrsq{0.43}{(+)}
& 15.9\% & \SCrsq{0.77}{(+)}\\

path norm 
& 24.5\% & \SCrsq{0.48}{(+)}
& 20.2\% & \SCrsq{0.68}{(+)}
& 17.8\% & \SCrsq{0.55}{(+)}
& 14.9\% & \SCrsq{0.70}{(+)}\\

spec norm
& 18.5\% & \SCrsq{0.57}{(+)}
& 19.7\% & \SCrsq{0.62}{(+)}
& 19.0\% & \SCrsq{0.48}{(+)}
& 16.7\% & \SCrsq{0.61}{(+)}\\

\bottomrule
\end{tabular}
}
\label{tab:single_factor_analysis_adamw}
\end{table}


\begin{table}[h]
\captionsetup{size=small}
\caption{
\textbf{Two-factor analysis} (optimizer AdamW). 
Table~\ref{tab:trained_by_adamw} is replicated here for convenient reference.
}
\vspace{-10pt}

\centering
\footnotesize
\setlength{\tabcolsep}{5pt}
\setlength{\fboxsep}{0.2pt}
\renewcommand{\arraystretch}{1.2}
\resizebox{0.99\linewidth}{!}{%
\begin{tabular}{
    @{}l
    *{3}{
        >{\raggedleft\arraybackslash}p{1.05cm}
        >{\centering\arraybackslash}p{1.50cm}
        !{\SCbodysep}
    }
    >{\raggedleft\arraybackslash}p{1.00cm}
    >{\centering\arraybackslash}p{1.40cm}
    @{}
}
\toprule

\multicolumn{1}{@{}c}{%
    \multirow{3}{*}{%
        \makecell[c]{%
            $\boldsymbol{(S,C)}$\\
            \textbf{Metric \textit{Pair}}%
        }%
    }%
}
& \multicolumn{8}{c}{\textbf{CIFAR-10}} \\

& \multicolumn{2}{c}{\textbf{ResNet18} \scriptsize{adamw}}
& \multicolumn{2}{c}{\textbf{VGG13} \scriptsize{adamw}}
& \multicolumn{2}{c}{\textbf{ResNet18 no-BN} \scriptsize{adamw}}
& \multicolumn{2}{c}{\textbf{VGG13 no-BN} \scriptsize{adamw}} \\[-3pt]

\cmidrule(lr){2-3}
\cmidrule(lr){4-5}
\cmidrule(lr){6-7}
\cmidrule(lr){8-9}

& \multicolumn{1}{r}{\textbf{PCR}}
& \multicolumn{1}{c}{$\boldsymbol{R^2}\,(S,C)$}
& \multicolumn{1}{r}{\textbf{PCR}}
& \multicolumn{1}{c}{$\boldsymbol{R^2}\,(S,C)$}
& \multicolumn{1}{r}{\textbf{PCR}}
& \multicolumn{1}{c}{$\boldsymbol{R^2}\,(S,C)$}
& \multicolumn{1}{r}{\textbf{PCR}}
& \multicolumn{1}{c}{$\boldsymbol{R^2}\,(S,C)$}
\\[-3pt]


\midrule

\SCtrFunc
& 44.1\% & \SCrsq{0.59}{(+,+)}
& 18.0\% & \SCrsq{0.83}{(+,+)}
& 32.5\% & \SCrsq{0.69}{(-,+)}
& 28.4\% & \SCrsq{0.15}{(+,+)}\\

\SClmaxFunc
& 46.1\% & \SCrsq{0.59}{(+,+)}
& 17.1\% & \SCrsq{0.83}{(+,+)}
& 27.5\% & \SCrsq{0.67}{(-,+)}
& 29.9\% & \SCrsq{0.15}{(+,+)}\\

\iteSCtrfsFunc
& 3.6\% & \SCrsq{0.70}{(+,+)}
& 6.7\% & \SCrsq{0.91}{(+,+)}
& 6.0\% & \SCrsq{0.75}{(+,+)}
& 6.0\% & \SCrsq{0.78}{(+,+)}\\

\iteSClmaxfsFunc
& 4.3\% & \SCrsq{0.64}{(+,+)}
& 5.5\% & \SCrsq{0.81}{(+,+)}
& 6.7\% & \SCrsq{0.73}{(+,+)}
& 10.8\% &\SCrsq{0.66}{(+,+)}\\

\addlinespace[4pt]

\SCtrLtwo
& 28.9\% & \SCrsq{0.78}{(+,+)}
& 8.8\% & \SCrsq{0.57}{(+,+)}
& 42.5\% & \SCrsq{0.51}{(-,+)}
& 17.0\% & \SCrsq{0.79}{(+,+)}\\

\SClmaxLtwo
& 30.1\% & \SCrsq{0.78}{(+,+)}
& 8.1\% & \SCrsq{0.57}{(+,+)}
& 36.7\% & \SCrsq{0.49}{(-,+)}
& 17.0\% & \SCrsq{0.79}{(+,+)}\\

\iteSCtrfsLtwo
& 10.1\% & \SCrsq{0.76}{(+,+)}
& 7.3\% & \SCrsq{0.91}{(+,+)}
& 10.5\% & \SCrsq{0.43}{(-,+)}
& 4.5\% & \SCrsq{0.85}{(+,+)}\\

\iteSClmaxfsLtwo
& 9.7\% & \SCrsq{0.78}{(+,+)}
& 7.1\% & \SCrsq{0.81}{(+,+)}
& 9.6\% & \SCrsq{0.51}{(-,+)}
& 5.9\% & \SCrsq{0.93}{(+,+)}\\

\addlinespace[4pt]

\SCtrPath 
& 56.2\% & \SCrsq{0.48}{(-,+)}
& 31.0\% & \SCrsq{0.70}{(-,+)}
& 43.9\% & \SCrsq{0.59}{(-,+)}
& 16.6\% & \SCrsq{0.73}{(+,+)}\\

\SClmaxPath
& 56.7\% & \SCrsq{0.48}{(-,+)}
& 29.5\% & \SCrsq{0.70}{(-,+)}
& 37.9\% & \SCrsq{0.56}{(-,+)}
& 16.5\% & \SCrsq{0.73}{(+,+)}\\

\iteSCtrfsPath
& 29.3\% & \SCrsq{0.60}{(-,+)}
& 19.6\% & \SCrsq{0.89}{(+,-)}
& 10.0\% & \SCrsq{0.55}{(+,+)}
& 4.6\% & \SCrsq{0.84}{(+,+)}\\

\iteSClmaxfsPath
& 28.3\% & \SCrsq{0.48}{(-,+)}
& 17.9\% & \SCrsq{0.89}{(+,+)}
& 9.5\% & \SCrsq{0.56}{(-,+)}
& 6.1\% & \SCrsq{0.91}{(+,+)}\\

\addlinespace[4pt]

\SCtrSpec
& 42.2\% & \SCrsq{0.64}{(+,+)}
& 17.8\% & \SCrsq{0.66}{(+,+)}
& 44.9\% & \SCrsq{0.56}{(-,+)}
& 16.1\% & \SCrsq{0.66}{(+,+)}\\

\SClmaxSpec
& 44.1\% & \SCrsq{0.64}{(+,+)}
& 15.9\% & \SCrsq{0.66}{(+,+)}
& 38.3\% & \SCrsq{0.52}{(-,+)}
& 16.3\% & \SCrsq{0.66}{(+,+)}\\

\iteSCtrfsSpec
& 12.7\% & \SCrsq{0.69}{(+,+)}
& 8.9\% & \SCrsq{0.94}{(+,+)}
& 9.7\% & \SCrsq{0.49}{(+,+)}
& 4.7\% & \SCrsq{0.83}{(+,+)}\\

\iteSClmaxfsSpec
& 12.6\% & \SCrsq{0.69}{(+,+)}
& 8.7\% & \SCrsq{0.87}{(+,+)}
& 9.4\% & \SCrsq{0.49}{(-,+)}
& 6.0\% & \SCrsq{0.89}{(+,+)}\\

\bottomrule
\end{tabular}%

}
\label{tab:trained_by_adamwREPEAT}
\end{table}

\end{document}